\documentclass{article}
\usepackage{microtype}
\usepackage{graphicx}
\usepackage{subfigure}
\usepackage{booktabs} % for professional tables

\usepackage{setspace}

\usepackage[moderate, bibbreaks=normal, bibnotes=normal, mathspacing=normal, mathdisplays=normal, charwidths=normal, tracking=normal, indent=normal, lists=tight]{savetrees}
\usepackage{hyperref}
\usepackage{amsmath,amsfonts,bm, mathtools}
\usepackage{amsthm}

\usepackage[accepted]{icml2022}

\icmltitlerunning{SQ-VAE}
\icmltitlerunning{SQ-VAE: Variational Bayes on Discrete Representation with Self-annealed Stochastic Quantization}

\usepackage{amsmath,amsfonts,bm}
\usepackage{amsthm}

\def\beqref#1{(\ref{#1})}
\newcommand{\pdata}{p_{\rm{data}}}
\newcommand{\E}{\mathbb{E}}
\newcommand{\Ls}{\mathcal{L}}

\newcommand{\softmax}{\mathrm{softmax}}

\newcommand{\KL}{D_{\mathrm{KL}}\hspace{-1pt}}

\DeclareMathOperator*{\argmin}{arg\,min}

\usepackage{hyperref}
\usepackage{url}
\usepackage{multirow}
\usepackage[super]{nth}
\usepackage{comment}
\usepackage{url}

\usepackage{threeparttable}
\usepackage{array}
\theoremstyle{plain}

\newtheorem{proposition}{Proposition}

\newtheorem*{hypothesis*}{Main Hypothesis}
\newcommand{\bhline}[1]{\noalign{\hrule height #1}}

\newcounter{num}
\newcommand{\Rnum}[1]{\setcounter{num}{#1}\Roman{num}}

\newcommand{\diag}{\mathrm{diag}}

\newcommand{\sg}{\mathrm{sg}}
\renewcommand{\softmax}{\text{softmax}}
\newcommand{\bb}{\mathbf{b}}
\newcommand{\bB}{\mathbf{B}}
\newcommand{\bI}{\mathbf{I}}

\newcommand{\bk}{\mathbf{k}}

\newcommand{\bv}{\mathbf{v}}
\newcommand{\bV}{\mathbf{V}}
\newcommand{\bw}{\mathbf{w}}
\newcommand{\bW}{\mathbf{W}}
\newcommand{\bx}{\mathbf{x}}

\newcommand{\bz}{\mathbf{z}}
\newcommand{\bZ}{\mathbf{Z}}
\newcommand{\bZq}{\bZ_\mathrm{q}}
\newcommand{\hbZq}{\hat{\bZ}_\mathrm{q}}

\newcommand{\btheta}{{\bm\theta}}
\newcommand{\bomega}{{\bm\omega}}
\newcommand{\bphi}{{\bm\phi}}
\newcommand{\bvarphi}{{\bm\varphi}}
\newcommand{\realnum}{\mathbb{R}}

\usepackage{colortbl}
\definecolor{tb}{rgb}{0.678,0.839,1.0}
\definecolor{tr}{rgb}{1.0,1.0,1.0}

\begin{document}

\twocolumn[
% \icmltitle{SQ-VAEs: Bridging the Gap between VAE and VQ-VAE for\\ Discrete Representation of Continuous and Categorical Distributions}
% \icmltitle{SQ-VAE: \\Bridging VAE and VQ-VAE with Stochastic dequantization and quantization}
% \icmltitle{SQ-VAE: VQ-VAE That Completely Follows Variational Inference}
% \icmltitle{SQ-VAE: Training VQ-VAE in Variational Framework}

% \icmltitle{SQ-VAE: Learning Discrete Representation via Stochasticity and Variational Inference}
% \icmltitle{SQ-VAE: Learning Discrete Representation via Variational Inference with Stochasticity}
% \icmltitle{From VAE to VQ-VAE:\\ Variational Bayes on Discrete Representation via Stochastic DeQuantization}  

% \icmltitle{SQ-VAE:\\ Bridging Training Schemes of VQ-VAE and VAE with Stochastic Quantization}
% \icmltitle{SQ-VAE: Metamorphosis from VAE to VQ-VAE with\\ Variational Bayes on Discrete Representation}

\icmltitle{SQ-VAE: Variational Bayes on Discrete Representation with\\ Self-annealed Stochastic Quantization}

% Keyword Candidates to be appeared on title
% "From VAE to VQ-VAE", "Stochastic DeQuantization", "Variational Bayesian","Stochastic categorical posterior"

% \icmltitle{SQ-VAEs: Taming VQ-VAE with Variational Inference for\\ Discrete Representation of Continuous and Categorical Distributions}




\begin{icmlauthorlist}
\icmlauthor{Yuhta Takida}{jp}
\icmlauthor{Takashi Shibuya}{jp}
\icmlauthor{WeiHsiang Liao}{jp}
\icmlauthor{Chieh-Hsin Lai}{jp}
\icmlauthor{Junki Ohmura}{jp}
\icmlauthor{Toshimitsu Uesaka}{jp}
\icmlauthor{Naoki Murata}{jp}
\icmlauthor{Shusuke Takahashi}{jp}
\icmlauthor{Toshiyuki Kumakura}{us}
\icmlauthor{Yuki Mitsufuji}{jp}
\end{icmlauthorlist}

\icmlaffiliation{jp}{Sony Group Corporation, Japan}
\icmlaffiliation{us}{Sony Corporation of America, USA}
% \icmlaffiliation{to}{Department of Computation, University of Torontoland, Torontoland, Canada}
% \icmlaffiliation{goo}{Googol ShallowMind, New London, Michigan, USA}
% \icmlaffiliation{ed}{School of Computation, University of Edenborrow, Edenborrow, United Kingdom}

\icmlcorrespondingauthor{Yuhta Takida}{yuta.takida@sony.com}
% \icmlcorrespondingauthor{Eee Pppp}{ep@eden.co.uk}

\icmlkeywords{Machine Learning, ICML}

\vskip 0.3in
]



\printAffiliationsAndNotice{}  % leave blank if no need to mention equal contribution
% \doublespacing
% \onecolumn

\begin{abstract}
One noted issue of vector-quantized variational autoencoder (VQ-VAE) is that the learned discrete representation uses only a fraction of the full capacity of the codebook, also known as codebook collapse. We hypothesize that the training scheme of VQ-VAE, which involves some carefully designed heuristics, underlies this issue. In this paper, we propose a new training scheme that extends the standard VAE via novel stochastic dequantization and quantization, called stochastically quantized variational autoencoder (SQ-VAE). In SQ-VAE, we observe a trend that the quantization is stochastic at the initial stage of the training but gradually converges toward a deterministic quantization, which we call self-annealing. Our experiments show that SQ-VAE improves codebook utilization without using common heuristics. Furthermore, we empirically show that SQ-VAE is superior to VAE and VQ-VAE in vision- and speech-related tasks.
\end{abstract}

%% Section
\section{Introduction}
\label{sec:introduction}
% VAE
The use of a variational autoencoder (VAE)~\citep{kingma2013auto,higgins2017beta,zhao2019infovae} is one of the popular approaches to generative modeling.
% and has achieved remarkable success in various domains such as vision, audio and videos.
VAE consists of a pair of an encoder and a decoder, which are jointly trained by maximizing the evidence lower bound (ELBO) of the observed data~\citep{jordan1999introduction}. The encoder maps the input data to a variable in a latent space, whereas the decoder converts the latent variable back into a sample of data space. A new sample is generated by decoding a latent variable that was sampled from a prior distribution.
%Generative process in VAEs consists of a latent space with a prior distribution and a stochastic decoder.
%The decoder maps the latent variables into a data space for generation.
%The objective function for the VAE, the evidence lower bound (ELBO) of log-likelihood~\citep{jordan1999introduction}, is originally derived in the variational inference where a stochastic encoder is introduced to approximate a reverse process of the decoder.
%In usual VAE, a latent space is set much lower-dimensional continuous one than data space, which enables compression of the target data into much lower dimensional vectors.
%Although VAE is theoretically solid and easy to train, it can cause blurred generation, which is often related to a phenomenon of posterior collapse~\citep{bowman2015generating,sonderby2016ladder,alemi2017fixing,xu2018spherical,he2019lagging,razavi2019preventing,dai2020usual}.

% VQ-VAE
Apart from VAE, a variant called Vector Quantized VAE (VQ-VAE)~\citep{van2017neural}, shows its superiority in several sample generation tasks ~\citep{dhariwal2020jukebox,ramesh2021zero,esser2021taming}. In VQ-VAE, the encoded latent variables are quantized to their nearest neighbors in a learnable codebook, and the data samples are decoded from the quantized latent variables. Samples are generated by first sampling the discrete latent variables from an approximated prior, then the sampled latent variables are decoded into synthetic samples. The approximated prior can be, for example, a PixelCNN~\citep{van2016pixel} that is trained on the latent space of training samples.
%VQ-VAE introduces a codebook to discretely compress the latent variables.
% , and many attractive generative models based on it have been proposed for various tasks~\citep{dhariwal2020jukebox,ramesh2021zero,esser2021taming}.
Although VQ-VAE shares some similarities with VAE, its training does not follow the standard variational Bayes framework~\citep{ghosh2019from}.
Instead, it relies on carefully designed heuristics such as the use of a stop-gradient operator and the straight-through estimation of gradients. Even so, VQ-VAE often suffers from \textit{codebook collapse}, which means that most of the codebook elements are not being used at all. This results in the deterioration of reconstruction accuracy~\citep{kaiser2018fast}.
To address this problem, techniques such as the exponential moving average (EMA)~\citep{polyak1992acceleration} update scheme, codebook reset~\citep{dhariwal2020jukebox,william2020hierarchical}, and hyperparameter tuning are often employed~\citep{roy2018theory}.
We suspect that deterministic quantization is the cause of codebook collapse. Although the original approach with straight-through estimation is intuitive and elegant, some codebook elements can never be selected in cases of bad initialization. Therefore, we propose a framework that combines stochastic quantization and VAE, called stochastically quantized VAE (SQ-VAE)\footnote{Our code is available at \url{https://github.com/sony/sqvae}.}. It can address the low codebook utilization issue of VQ-VAE and can be explained within the scope of the usual variational Bayes framework. Moreover, its training requires no exhaustive hyperparameter tuning and does not rely on heuristic techniques such as stop-gradient, codebook reset, or EMA update. 

SQ-VAE introduces a pair of stochastic dequantization and quantization processes in the latent space. These processes are characterized by probability distributions with trainable parameters. This setup allows us to train the model within the usual variational Bayes framework without the need of conventional heuristics. Training the model requires only one hyperparameter, which can be treated in straightforward ways as in~\citep{jang2017categorical}. Optimizing the ELBO gradually reduces the stochasticity of the quantization process during the training, which we call \textit{self-annealing}.
In general, SQ-VAE does not impose any assumption on the data distribution; hence, we can model the stochastic quantization and dequantization processes via Gaussian distributions for example.
However, we found that when the data distribution is categorical and cross entropy (CE) loss is used, this setup often yields an unsatisfactory performance. As a remedy, we propose the use of the von Mises--Fisher (vMF) distribution (see Appendix~\ref{sec:app_notations}) in place of the Gaussian distribution and show its effectiveness in Section~\ref{sec:sub_vmf_sqvae}. 

% Our main contributions are summarized below.
We summarize our contributions below.
\begin{enumerate}
    %\vspace*{-2pt}
    % \setlength{\parskip}{0cm}
    % \setlength{\itemsep}{0cm}
    \item
    We propose SQ-VAE, which is variational autoencoder equipped with stochastic quantization and trainable posterior categorical distribution. 
    SQ-VAE can be explained within an ordinary variational Bayes framework and may serve as a drop-in replacement of conventional VQ-VAE. 
    \item
    In SQ-VAE, the annealing of the stochasticity of the quantization process leads to a greater codebook utilization. We provide a theoretical insight into this self-annealing and validate it with an empirical study.
    \item
    We design two instances of SQ-VAE: Gaussian SQ-VAE for general cases and vMF SQ-VAE specialized for categorical data distribution. 
    \item
    We evaluate SQ-VAE in vision- and speech-related generation tasks. The evaluation shows that SQ-VAE achieves better reconstruction than VQ-VAE.
    Furthermore, the performance of SQ-VAE can be improved by simply increasing the codebook size, which is not the case for VQ-VAE. 
\end{enumerate}

%This paper is organized as follows. Section~\ref{sec:background} explains the formulation of VAE and VQ-VAE.
%In Section~\ref{sec:sqvae}, we introduce the idea of SQ-VAE and propose the two instances: Gaussian SQ-VAE and vMF SQ-VAE. 
%In Section~\ref{sec:related_works}, we compare SQ-VAE with related works. Section~\ref{sec:experiments} reports the experimental results on vision and speech tasks.
%Finally, Section~\ref{sec:conclusion} concludes the paper.

Throughout this paper, we use $\{b_j\}_{j=1}^{J}$ to denote a set of elements $b_j$; $[N]$ denotes the set of positive integers less than or equal to $N$; the capital letters $P$ and $Q$ denote the probability mass functions, whereas the lower case letters $p$ and $q$ denote the probability density functions.

%----- Section
\section{Background}
\label{sec:background}
%In this section, we briefly introduce VAE and VQ-VAE.
\paragraph{VAE}
Consider an observation $\bx\in\realnum^{D}$ and a target data distribution $\pdata(\bx)$, which models finite samples.
The standard VAE consists of a stochastic encoder–decoder pair: a decoder $p_\btheta(\bx|\bz)$ and an approximated posterior $q_\bphi(\bz|\bx)$, where $\btheta$ and $\bphi$ are trainable parameters. The latent variables $\bz\in\realnum^{d_z}$ are assumed to follow a prior distribution $p(\bz)$.
%, and $q_\bphi(\bz|\bx)$ acts as the approximated posterior. 
Data are generated by first sampling $\bz$ from the prior $p(\bz)$ then obtaining $\bx$ by feeding $\bz$ into the stochastic decoder, $p_\btheta(\bx|\bz)$.
The negative ELBO per sample $\bx$ is expressed as $\Ls_\text{VAE}=$
\begin{align}
    \E_{q_\bphi(\bz|\bx)}\left[-\log{}p_\btheta(\bx|\bz)\right]
    +\KL(q_\bphi(\bz|\bx)\parallel p(\bz)).
    \label{eq:elbo_general}
\end{align}
To compute the ELBO of likelihood of samples $\bx$ analytically, the approximated posterior is usually modeled with conditional Gaussian as $q_\bphi(\bz|\bx)=\mathcal{N}(g_\bphi(\bx),\diag(\bm{\sigma}_\bphi(\bx)))$ with two mappings $g_\bphi:\realnum^{D}\to\realnum^{d_z}$ and $\bm{\sigma}_\bphi:\realnum^{D}\to\realnum^{d_z}$.

If the target data distribution is continuous, the stochastic decoder can be modeled by a Gaussian distribution with a mapping $f_{\btheta}:\realnum^{d_z}\to\realnum^D$ as
\begin{align}
    p_{\btheta}(\bx|\bz)=\mathcal{N}(f_{\btheta}(\bz),\sigma^2\bI),
    \label{eq:decoder_gaussian}
\end{align}
which reduces the first term in \beqref{eq:elbo_general} into the mean squared error (MSE).
In contrast, if the data distribution is discrete and has $C_\mathrm{all}$ categories, the stochastic decoder for the $d$th element of $\bx$, $x_d$, can be modeled as a categorical distribution with $f_{\btheta,d}^c:\realnum^{d_z}\to\realnum$ $(c\in[C_\mathrm{all}])$ as
\begin{align}
    % P_{\btheta}(x_d=c|\bz)=\mathrm{\softmax}_c(f_{\btheta,d}^c(\bz)),
    P_{\btheta}(x_d=c|\bz)=\mathrm{\softmax}_c(\{f_{\btheta,d}^{c^\prime}(\bz)\}_{{c^\prime}=1}^{C_\mathrm{all}}),
    \label{eq:decoder_categorical}
\end{align}
where the softmax is operated among $c^\prime$ (see Appendix~\ref{sec:app_notations}). In this case, the first term in \beqref{eq:elbo_general} becomes the CE loss.

\paragraph{VQ-VAE}
In contrast to VAE, VQ-VAE consists of a deterministic encoder--decoder path and a trainable \textit{codebook}.
The codebook is a set $\mathbf{B}$, which contains $K$ $d_b$-dimensional vectors $\{\bb_k\}_{k=1}^K$.
%% Moved from section 3
% With the codebook elements, we first define a discrete space  as $\mathbf{B}:=\{\bb_k\}_{k=1}^K$.
%We first define a discrete space $\mathbf{B}$ as all the plausible codebook vectors during training.
A $d_z$-dimensional discrete latent space related to the codebook can be interpreted as the $d_z$-ary Cartesian power of $\mathbf{B}$, $\mathbf{B}^{d_z}\subset\realnum^{d_b\times{d_z}}$. 
%is constructed as Cartesian products of $\mathbf{B}$, $\mathbf{B}^{d_z}(\subset\realnum^{d_b\times{}d_z})$. %$\mathbf{B}^{d_z}(\subset\realnum^{d_b\times{}d_z})$, to depict the latent features of the target distribution $\pdata(\bx)$.
We denote a latent variable in $\mathbf{B}^{d_z}$ and its $i$th column vector as $\bZq\in\mathbf{B}^{d_z}$ and $\bz_{\mathrm{q},i}\in\mathbf{B}$, respectively.
The deterministic encoding process from $\bx$ to $\bZq$ includes a mapping $\hat{\bZ}_{\mathrm{q}}=g_{\bphi}(\bx)$ with $g_{\bphi}: \realnum^D\to\realnum^{d_b\times{}d_z}$ and the quantization process of $\hat{\bZ}_\mathrm{q}$ onto $\mathbf{B}^{d_z}$.
The quantization process is modeled as a deterministic categorical posterior distribution, in which $\hat{\bz}_{\mathrm{q},i}$ is always mapped to its nearest neighbor $\bz_{\mathrm{q},i}$, i.e., $\bz_{\mathrm{q},i}=\argmin_{\bb_k}\|\hat{\bz}_{\mathrm{q},i}-\bb_k\|_2$.
% Although this quantization is intuitive and elegant, some codebook elements can never be selected in cases of bad initialization. 
The objective function of VQ-VAE is
\begin{align}
    \Ls_\text{VQ}
    =&-\log{}p_\btheta(\bx|\bZq)+\|\sg[g_\bphi(\bx)]-\bZq\|_F^2\notag\\
    &~+\beta\|g_\bphi(\bx)-\sg[\bZq]\|_F^2,
    \label{eq:objective_vq}
\end{align}
where $\sg[\cdot]$ denotes the stop-gradient operator and $\beta$ is set between 0.1 and 2.0~\citep{van2017neural}.
To improve performance and convergence rate, EMA update is often applied only to the second term, which corresponds to the update of the codebook.
%the second term is often trained by EMA.

Note that the objective functions of VAE and VQ-VAE can both be interpreted as the sum of the reconstruction error and the latent regularization penalty.

\begin{figure*}[t]
\vskip 0.1in
  \centering
   \includegraphics[width=.83\textwidth]{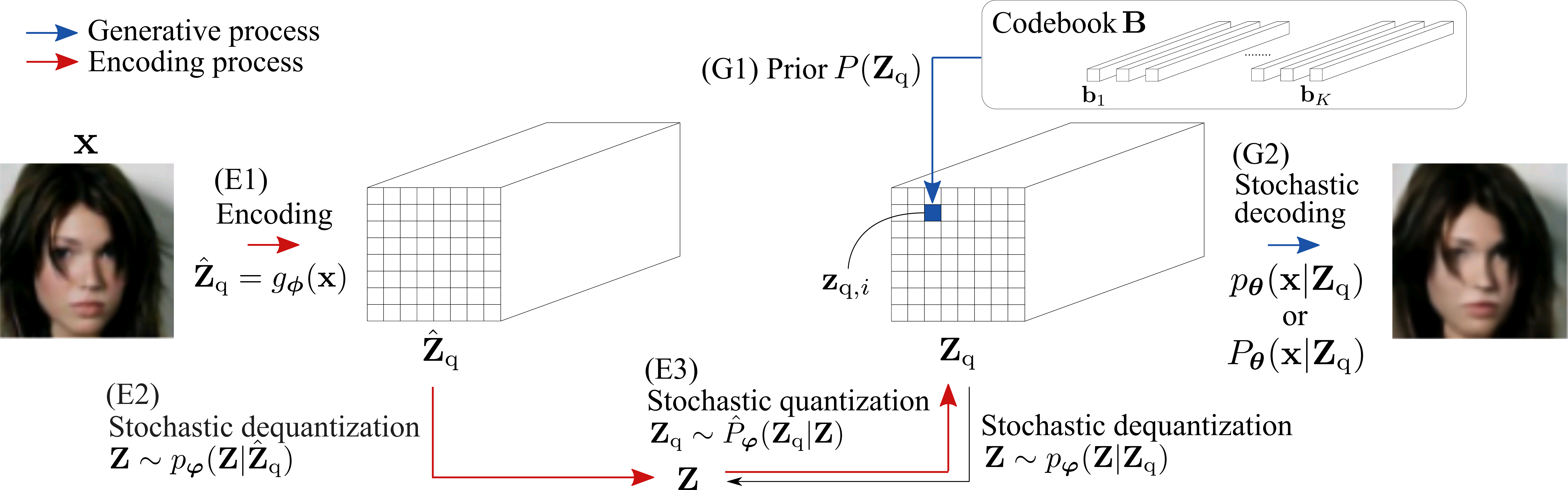}
   \caption{
   Encoding and generative processes of SQ-VAE.
   The encoding path from $\bx$ to $\bZq$ consists of (E1) deterministic encoding, (E2) stochastic dequantization, and (E3) quantization processes.
   For generation, in (G1) we first sample $\bZq\in\mathbf{B}^{d_z}$ from the prior $p(\mathbf{Z}_\mathrm{q})$.
   Then, in (G2) we feed $\bZq$ into the stochastic decoder to generate data samples.
   %We consider a dequantized latent variable $\bZ$, which is not used for generation but generated from $\bZq$.
   }
   \label{fig:sq-vae}
\vskip -0.1in
\end{figure*}

%%----- Section
\section{Stochastically Quantized VAE}
\label{sec:sqvae}
In this section, we propose SQ-VAE and its two instances, which are Gaussian SQ-VAE and vMF SQ-VAE. This framework bridges the training schemes of VAE and VQ-VAE. It relieves VQ-VAE from the heuristic techniques and reduces the difficulty of hyperparameter tuning. Moreover, it incorporates the self-annealing of a trainable categorical posterior distribution, which gradually approaches the deterministic quantization of VQ-VAE during the training.
Furthermore, we provide theoretical and empirical support about the benefit of the self-annealing mechanism.
The pseudo-codes of the two instances can be found in Appendix~\ref{sec:app_psuedo_code}.

%-- subsection
\subsection{Overview of SQ-VAE}
\label{sec:sub_outline_sqvae}

The outline of SQ-VAE is shown in Figure~\ref{fig:sq-vae}.
Identical to VQ-VAE, SQ-VAE also has a trainable codebook $\mathbf{B}:=\{\bb_k\}_{k=1}^K$.
As a generative model, the goal of SQ-VAE is to learn a generative process $\bx\sim{}p_\btheta(\bx|\bZq)$ %(or $\bx\sim{}P_\btheta(\bx|\bZq)$) 
with $\bZq\sim{}P(\bZq)$ to generate samples that belong to the data distribution $\pdata(\bx)$, where $P(\bZq)$ denotes the prior distribution of the discrete latent space $\mathbf{B}^{d_z}$.
The prior $P(\bZq)$ is assumed to be an i.i.d. uniform distribution in the main training stage as in VQ-VAE, i.e., $P(\bz_{\mathrm{q},i}=\bb_k)=1/K$ for $k\in[K]$. A second training will take place to learn $P(\bZq)$ after the main training stage. Since the exact evaluation of $p_\btheta(\bZq|\bx)$ is intractable, the  approximated posterior $q_\bphi(\bZq|\bx)$ is used instead.

% In this setup, although we can establish the generative process in a way identical to VQ-VAE, the construction of the encoding process from $\bx$ to $\bZq$ is not straightforward owing to the discrete property of $\bZq$.
In this setup, although we can establish the generative process following that in VQ-VAE, the construction of the encoding process from $\bx$ to $\bZq$ is not straightforward owing to the discrete property of $\bZq$. Therefore, we introduce two auxiliary variables to ease the explanation: $\bZ$ and $\hbZq$.
$\bZ$ is the continuous variable converted from $\bZq$ via the dequantization process $p_\bvarphi(\bZ|\bZq)$, where $\bvarphi$ indicates its parameters.
%In this way, we may obtain the continuous tensor by  $\bZ|\bZq\sim{}p_\bvarphi(\bZ|\bZq)$.
Furthermore, we may derive the inverse process of $p_\bvarphi(\bZ|\bZq)$, i.e., the stochastic quantization process $\hat{P}_\bvarphi(\bZq|\bZ)$, from Bayes' theorem $\hat{P}_\bvarphi(\bZq|\bZ)\propto{}p_\bvarphi(\bZ|\bZq)P(\bZq)$.
On the other hand, $\hbZq$ is defined as $\hbZq=g_{\bphi}(\bx)$, which is the output of the deterministic encoder $g_{\bphi}:\realnum^{D}\to\mathbb{R}^{d_b\times{}d_z}$ given a sample $\bx$.
Ideally, $\hbZq$ should be close to $\bZq$.
Similarly, the dequantization process of $\hbZq$ can be written as $\bZ|\hbZq\sim{}p_\bvarphi(\bZ|\hbZq)$.
As in Figure~\ref{fig:sq-vae}, stacking the processes $p_\bvarphi(\bZ|\hbZq)$ and $\hat{P}_\bvarphi(\bZq|\bZ)$ connects $\hbZq$ and $\bZq$, and thus establishes the stochastic encoding process from $\bx$ to $\bZq$ as $Q_\bomega(\bZq|\bx):=\E_{q_\bomega(\bZ|\bx)}[\hat{P}_\bvarphi(\bZq|\bZ)]$, where $\bomega:=\{\bphi,\bvarphi\}$ and $q_\bomega(\bZ|\bx):=p_\bvarphi(\bZ|g_{\bphi}(\bx))$.
%
% As in Figure~\ref{fig:sq-vae}, stacking the processes $p_\bvarphi(\bZ|\hbZq)$ and $\hat{P}_\bvarphi(\bZq|\bZ)$ connects $\hbZq$ and $\bZq$, and thus establishes the stochastic encoding process from $\bx$ to $\bZq$ as $q_\bomega(\bZ|\bx)\hat{P}_\bvarphi(\bZq|\bZ)$, where $\bomega:=\{\bphi,\bvarphi\}$ and $q_\bomega(\bZ|\bx):=p_\bvarphi(\bZ|g_{\bphi}(\bx))$.

At this point, we can derive the ELBO for SQ-VAE as $\log{}p_{\btheta}(\bx)\geq-\Ls_\text{SQ}(\bx;\btheta,\bomega,\mathbf{B}):=$
% $\log{}p_{\btheta}(\bx)\geq$
\begin{align}
    % &\E_{Q_{\bomega}(\bZq|\bx)}\left[\log{}p_{\btheta}(\bx|\bZq)
    % \!-\!\KL(q_\bomega(\bZ|\bx)\parallel p_\bvarphi(\bZ|\bZq))\right]\notag\\
    % &~-\E_{q_\bomega(\bZ|\bx)}H(\hat{P}_\bvarphi(\bZq|\bZ))
    % % =:-\Ls_\text{SQ}(\bx),
    % =:-\Ls_\text{SQ}(\bx;\btheta,\bomega,\mathbf{B}),
    &\E_{q_\bomega(\bZ|\bx)\hat{P}_\bvarphi(\bZq|\bZ)}\left[\log{}\frac{p_{\btheta}(\bx|\bZq)p_\bvarphi(\bZ|\bZq)P(\bZq)}{q_\bomega(\bZ|\bx)\hat{P}_\bvarphi(\bZq|\bZ)}\right]\nonumber\\
    &=\E_{q_\bomega(\bZ|\bx)\hat{P}_\bvarphi(\bZq|\bZ)}\left[\log{}\frac{p_{\btheta}(\bx|\bZq)p_\bvarphi(\bZ|\bZq)}{q_\bomega(\bZ|\bx)}\right]\notag\\
    % &\geq\E_{q_\bomega(\bZ|\bx)\hat{P}_\bvarphi(\bZq|\bZ)}\left[\log{}p_{\btheta}(\bx|\bZq)
    % \!-\!\log\frac{q_\bomega(\bZ|\bx)}{p_\bvarphi(\bZ|\bZq)}\right]\notag\\
    &\qquad+\E_{q_\bomega(\bZ|\bx)}H(\hat{P}_\bvarphi(\bZq|\bZ))+\mathrm{const.},
    % =:-\Ls_\text{SQ}(\bx),
    % =:-\Ls_\text{SQ}(\bx;\btheta,\bomega,\mathbf{B}),
    \label{eq:elbo_sqvae}
\end{align}
% \begin{align}
%     &\E_{Q_{\bomega}(\bZq|\bx)}\left[\log{}p_{\btheta}(\bx|\bZq)
%     \!-\!\KL(p_\bvarphi(\bZ|g_{\bphi}(\bx))\parallel p_\bvarphi(\bZ|\bZq))\right]\notag\\
%     &\qquad-\E_{q_\bomega(\bZ|\bx)}H(\hat{P}_\bvarphi(\bZq|\bZ))=:-\Ls_\text{SQ}(\bx),
%     \label{eq:elbo_sqvae}
% \end{align}
where $H(P)$ denotes the entropy of $P$. In \beqref{eq:elbo_sqvae}, since $P(\bZq)$ is assumed to follow a uniform distribution, it results into a constant term and is thus omitted.
We hereafter omit the parameters of $\Ls_\text{SQ}$ for simplicity.
In the end, the main training is carried out by minimizing $\E_{\pdata(\bx)}\Ls_\text{SQ}(\bx)$. %, i.e., the negative ELBO
The encoder, the decoder, and the codebook are all optimized simultaneously during the process. In this way, the codebook optimization no longer requires heuristic techniques such as the stop-gradient, EMA, and codebook reset~\citep{dhariwal2020jukebox,william2020hierarchical}.
The expectation in the first term of \beqref{eq:elbo_sqvae} involves the categorical distribution $\hat{P}_\bvarphi(\bZq|\bZ)$, which can be approximated by the Gumbel--softmax relaxation~\citep{jang2017categorical,maddison2017concrete} to use the reparameterization trick in the backward pass of conventional VAE. 

\begin{table*}[t]
\centering
    \caption{Different parameterizations of the variance $\bm{\Sigma}_\bvarphi$ in Gaussian SQ-VAE.}
    \vskip 0.1in
    \renewcommand{\arraystretch}{1.25}
    \small
    \begin{tabular}{c|c|c|c}
        \bhline{0.8pt}
         & Variance $\bm{\Sigma}_\bvarphi$ & Unnormalized log-probability & Regularization objective $\mathcal{R}_\bvarphi^\mathcal{N}(\bZ, \bZq)$\\
        \hline
        (\Rnum{1}) & $\sigma_\bvarphi^2\mathbf{I}$
        & $\|\bb_k-\mathbf{z}_{i}(\mathbf{x})\|_2^2/2\sigma_\bvarphi^2$
        & $\|\bZ-\bZq\|_F^2/{2\sigma_\bvarphi^2}$\\
        (\Rnum{2}) & $\sigma_\bvarphi^2(\mathbf{x})\mathbf{I}$
        & $\|\bb_k-\mathbf{z}_{i}(\mathbf{x})\|_2^2/2\sigma_\bvarphi^2(\mathbf{x})$
        & $\|\bZ-\bZq\|_F^2/{2\sigma_\bvarphi^2(\bx)}$\\
        (\Rnum{3}) & $\sigma_{\bvarphi,i}^2(\mathbf{x})\mathbf{I}$
        & $\|\bb_k-\mathbf{z}_{i}(\mathbf{x})\|_2^2/2\sigma_{\bvarphi,i}^2(\mathbf{x})$
        & $\sum_{i=1}^{d_z}\|\bz_{i}(\bx)-\bz_{\mathrm{q},i}\|_2^2/{2\sigma_{\bvarphi,i}^2(\bx)}$\\
        (\Rnum{4}) & $\diag(\bm{\sigma}_{\bvarphi,i}^2(\mathbf{x}))$
        & $\sum_{j=1}^{d_b}(b_{k,j}-z_{i,j}(\mathbf{x}))^2/2\sigma_{\bvarphi,i,j}^2(\mathbf{x})$
        & $\sum_{i=1}^{d_z}\sum_{j=1}^{d_b}(z_{i,j}(\bx)-z_{\mathrm{q},i,j})^2/{2\sigma_{\bvarphi,i,j}^2(\bx)}$\\
        \bhline{0.8pt}
    \end{tabular}
    \label{tb:parameterization_variance}
    \vskip -0.1in
\end{table*}

%-- subsection
\subsection{Gaussian SQ-VAE}
\label{sec:sub_gaussian_sqvae}
% Assuming the dequantization process follows Gaussian distribution results in Gaussian SQ-VAE.
We design Gaussian SQ-VAE by assuming that the dequantization process follows a Gaussian distribution.
On the basis of the assumption, the dequantization process is modeled as
\begin{align}
    p_{\bvarphi}(\bz_i|\bZq)=\mathcal{N}(\bz_{\mathrm{q},i},\bm{\Sigma}_\bvarphi),
    \label{eq:dequantization_gausian}
\end{align}
in which $\bm{\Sigma}_\bvarphi$ is trainable.
From Bayes' theorem, we may recover $\bZq$ with the inverse of \beqref{eq:dequantization_gausian}, i.e., the stochastic quantization process, as $\hat{P}_\bvarphi(\bz_{\mathrm{q},i}=\bb_k|\bZ)=$
% \begin{align}
%     \softmax_{k}\left(-\frac{1}{2}(\bb_{k}-\bz_i)^\top\bm{\Sigma}_\bvarphi^{-1}(\bb_{k}-\bz_i)\right),
%     \label{eq:quantization_gausian}
% \end{align}
\begin{align}
    \softmax_{k}\left(\left\{-\frac{(\bb_{j}-\bz_i)^\top\bm{\Sigma}_\bvarphi^{-1}(\bb_{j}-\bz_i)}{2}\right\}_{j=1}^K\right),
    \label{eq:quantization_gausian}
\end{align}
where the unnormalized log-probabilities for $\bb_k$ in \beqref{eq:quantization_gausian} correspond to Mahalanobis' distance from $\bz_i$ with the variance $\bm{\Sigma}_\bvarphi$.
%The covariance of both processes is parameterized as $\bm{\Sigma}_\bvarphi$.
We further consider several parameterizations of $\bm{\Sigma}_\bvarphi$ and summarize them in Table~\ref{tb:parameterization_variance} with the corresponding unnormalized negative log-probabilities\footnote{Although more complicated parameterizations exist, they often lead to an unstable optimization in our experiments.}.
We examine their effectiveness in Section~\ref{sec:experiments}. The decoding and encoding setups of Gaussian SQ-VAE are described as follows.
%The encoding follows the process depicted in Figure~\ref{fig:sq-vae} done by concatenating (E1) the encoder, (E2) stochastic dequantizer and (E3) quantizer, we can encode $\bx$ to the discrete tensors $\bZq$ (see Figure~\ref{fig:sq-vae}).
%For the encoding path, we apply the same dequantization process to the encoded tensors $\hbZq$ as $p_{\bvarphi}(\bz_i|\hbZq)=\mathcal{N}(\bz_i|\hat{\bz}_{\mathrm{q},i},\bm{\Sigma}_\bvarphi)$.
% as $\bZq|\bx\sim{}\E_{p_{\bvarphi}(\bZ|g_\bphi(\bx))}[\hat{P}_\bvarphi(\bZq|\bZ)]$.

\paragraph{Decoding}
The usual Gaussian setup is adopted in the decoding such that  $p_{\btheta}(\bx|\bZq)=\mathcal{N}(f_{\btheta}(\bZq),\sigma^2\bI)$, where $\sigma^2\in\realnum_+$ and $\btheta$ are trainable parameters.

\paragraph{Encoding}
The encoding follows the process depicted in Figure~\ref{fig:sq-vae}, and the dequantization process applied to $\hbZq$ is $p_{\bvarphi}(\bz_i|\hbZq)=\mathcal{N}(\hat{\bz}_{\mathrm{q},i},\bm{\Sigma}_\bvarphi)$.

\paragraph{Objective Function}
The substitution of the encoding and decoding processes above into \beqref{eq:elbo_sqvae} gives
$\Ls_{\mathcal{N}\text{-SQ}}=$
\begin{align}
    &\E_{q_\bomega(\bZ|\bx)\hat{P}_\bvarphi(\bZq|\bZ)}\left[\frac{1}{2\sigma^2}\|\bx-f_\btheta(\bZ)\|_2^2+\mathcal{R}_\bvarphi^\mathcal{N}(\bZ, \bZq)\right]\notag\\
    &\quad-\E_{q_\bomega(\bZ|\bx)}H\left(\hat{P}_\bvarphi(\bZq|\bZ)\right)+\frac{D}{2}\log\sigma^2+\mathrm{const.},
    \label{eq:elbo_gaussian_sqvae}
\end{align}
where $\mathcal{R}_\bvarphi^\mathcal{N}(\bZ, \bZq)$ denotes the regularization objective in Table~\ref{tb:parameterization_variance}, depending on the parameterization of $\bm{\Sigma}_\bvarphi$.
The derivation detail can be found in Appendix~\ref{sec:app_sub_elbo_gaussian_sqvae}.
%. Its variations for each respective parameterization of the variance $\bm{\Sigma}_\bvarphi$ is summarized in Table~\ref{tb:parameterization_variance}.

%-- subsection
\subsection{Self-annealed Quantization}
\label{sec:sub_behavior_quantization}

Before proposing the next SQ-VAE instance, we would like to demonstrate the effectiveness of trainable parameters in (de)quantization processes.
In this subsection, we adopt the parameterization $\bm{\Sigma}_\bvarphi=\sigma_\bvarphi^2\bI$ (type \Rnum{1} in Table~\ref{tb:parameterization_variance}) for simplicity.

According to \beqref{eq:quantization_gausian}, $\bm{\Sigma}_\bvarphi$ controls the degree of stochasticity of the quantization during the training.
% In this setting, the first and second terms in the objective function~\beqref{eq:elbo_gaussian_sqvae} is balanced with trainable scalar parameters $\sigma^2$ and $\sigma_\bvarphi^2$.
%To investigate the convergence point of $\sigma_\bvarphi^2$, we 
We first consider two extreme cases, $\sigma^2\to\infty$ and $\sigma^2\to0$, with the following proposition whose proof is given in Appendix~\ref{sec:app_proof_proposition}.
\begin{proposition}
    Assume that $\pdata(\bx)$ has finite support, whereas $g_{\bphi}$ and $\{\bb_k\}_{k=1}^K$ are bounded.
    Let $\bomega^{*}=\{\bphi^*,\bvarphi^*\}$ be a minimizer of $\E_{\pdata(\bx)}\KL(Q_{\bomega}(\bZq|\bx)\parallel{}P_\btheta(\bZq|\bx))$ with fixed $\btheta$, $\sigma^2$ and $\{\bb_k\}_{k=1}^K$.
    If $\sigma^2\to0$, then $\sigma_{\bvarphi^*}^2\to0$.
    % Assuming that a probability $q_\bphi(\bZq)$ that $\bZq=g_\bphi(\bx)$ follows with $\bx\sim\pdata(\mathbf{x})$ has finite support, $f_{\btheta}$ and $g_{\bphi}$ can be arbitrarily complex, and $K>N_{\mathrm{data}}^{1/d_z}$ is finite.
    % Suppose $\bomega^{*}=\{\bphi^*,\bvarphi^*\}$ to be a minimizer of $\E_{\pdata(x)}\KL(Q_{\bomega}(\bZq|\bx)\parallel{}P_\btheta(\bZq|\bx))$ with $\btheta$ and $\sigma^2$ fixed.
    % If $\sigma^2\to0$, then $\sigma_{\bvarphi^*}^2\to0$.
    \label{th:to_deterministic_quantization_gaussian}
\end{proposition}
% \begin{proof}
%     In Appendix~\ref{sec:app_proof_proposition}.
% \end{proof}
% When $\sigma^2\to\infty$, the first term in \beqref{eq:elbo_gaussian_sqvae} diminishes. It is minimized when $\sigma_\bvarphi^2\to\infty$, where  $P_\bvarphi(\bz_{\mathrm{q},i}=\bb_k|\bZ)$ approaches uniform distribution.
When $\sigma^2\to\infty$, the first term in \beqref{eq:elbo_gaussian_sqvae} diminishes. It is minimized when $\sigma_\bvarphi^2\to\infty$, where $P_\bvarphi(\bz_{\mathrm{q},i}=\bb_k|\bZ)$ approaches a uniform distribution.
On the other hand, according to Proposition~\ref{th:to_deterministic_quantization_gaussian}, when $\sigma^2\to0$, it leads to $\sigma_\bvarphi^2\to0$. This means that $P_\bvarphi(\bz_{\mathrm{q},i}=\bb_k|\bZ)$ converges to the Kronecker delta function $\delta_{k,\hat{k}}$, where $\hat{k}=\argmin_{k}\|\bz_{i}-\bb_k\|_2$. This deterministic quantization is exactly the posterior categorical distribution of VQ-VAE.
According to the two cases above, if $\sigma^2$ decreases gradually during the training, the quantization process will also gradually decrease its stochasticity and approach the deterministic quantization. We refer to this as \textit{self-annealing}. 
%The extreme cases suggest that the assignment of quantization becomes \textit{harder} gradually as $\sigma^2\to0$.
%On the other hand, the quantization in VQ-VAE is always 1-NN assignment, which can make some codebook elements never used in cases of bad initializations.

\begin{figure*}[t]
\vskip 0.1in
  \centering
  \subfigure[Variance parameters]{\includegraphics[width=0.28\textwidth]{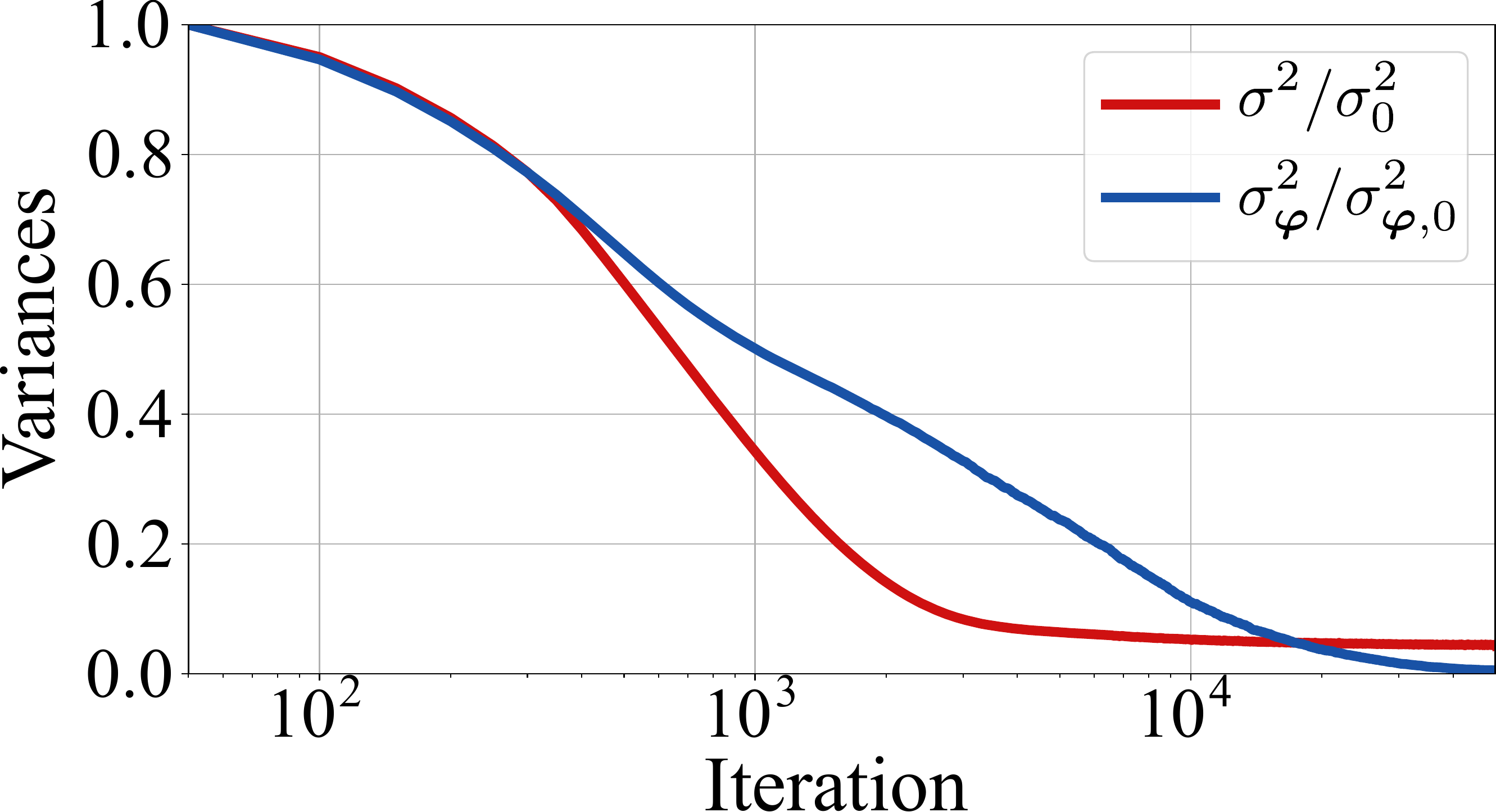}}
\hspace{10pt}%
  \subfigure[$H(\hat{P}_\bvarphi(\bz_{\mathrm{q},i}|\bZ))$
%   Average entropy of $\hat{P}_\bvarphi(\bz_{\mathrm{q},i}|\bZ)$
  ]{\includegraphics[width=0.28\textwidth]{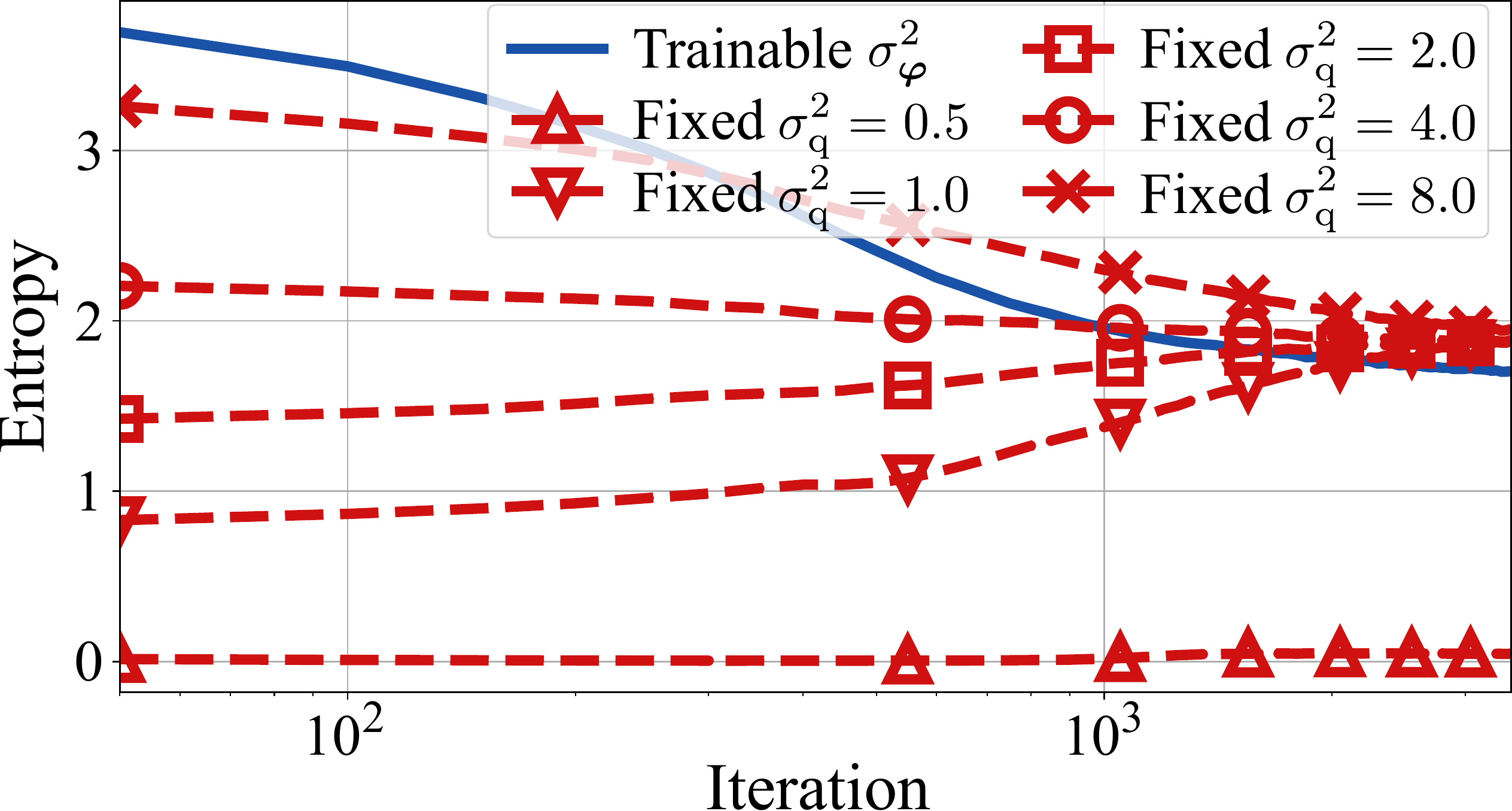}}
\hspace{10pt}%
  \subfigure[MSE]{\includegraphics[width=0.28\textwidth]{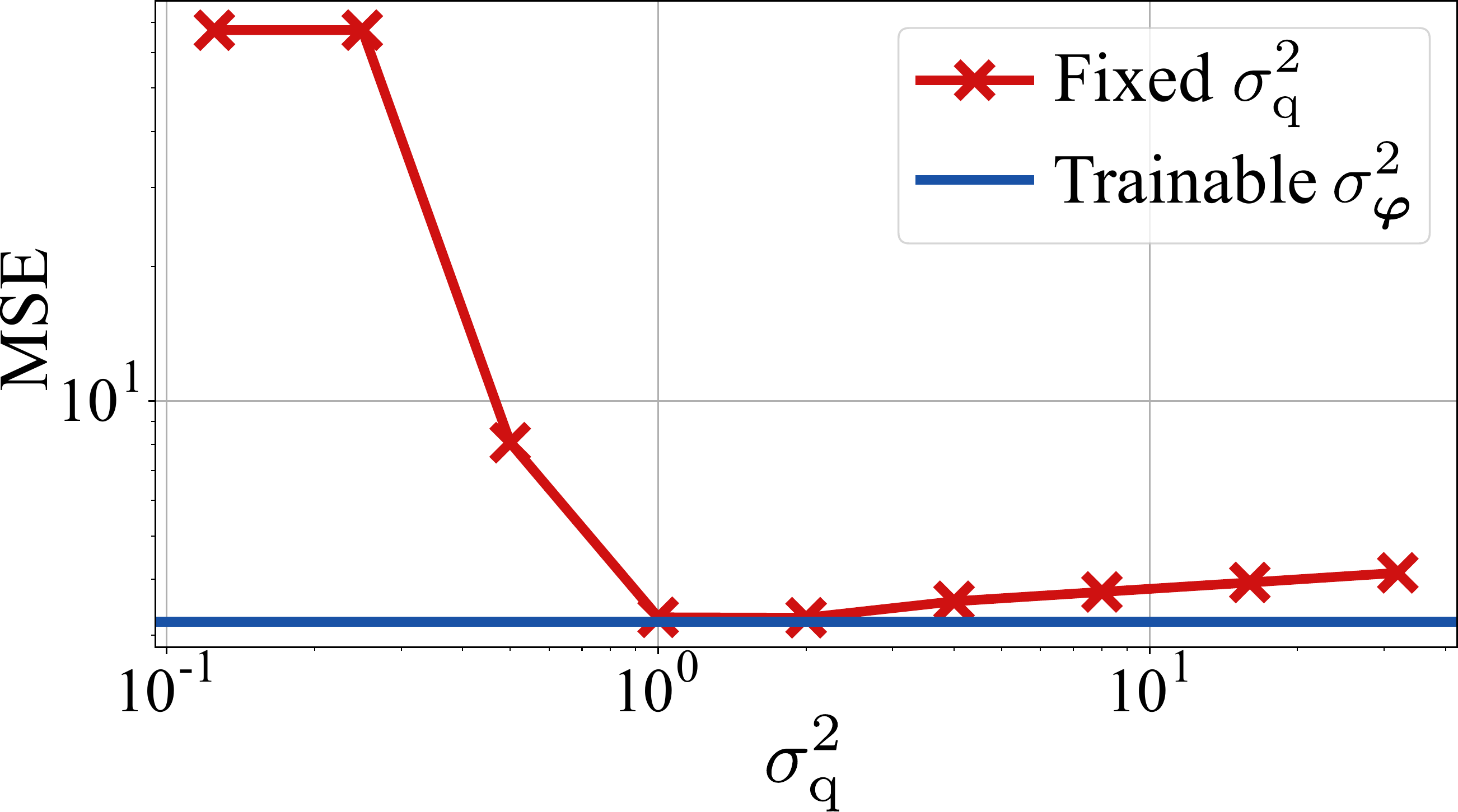}}
%   \subfigure[Trainable $\sigma_\bvarphi^2$]{\includegraphics[width=0.27\textwidth]{figures/results/analysis_mnist_topk.pdf}}
%   \hspace{20pt}%
%   \subfigure[Fixed $\sigma_\mathrm{q}^2=0.5$]{\includegraphics[width=0.27\textwidth]{figures/results/analysis_mnist_topk_fixed.pdf}}
%   \hspace{20pt}%
%   \subfigure[Fixed $\sigma_\mathrm{q}^2=2.0$]{\includegraphics[width=0.27\textwidth]{figures/results/analysis_mnist_topk_fixed2.pdf}}
  \caption{Empirical study on the dynamics related to $\sigma_\bvarphi^2$ in Section~\ref{sec:sub_behavior_quantization}. (a) The variance parameter $\sigma_\bvarphi^2$ (blue) decreased with $\sigma^2$ (red), where $\sigma_0^2$ and $\sigma_{\bvarphi,0}^2$ are their initial values. (b) Average entropy of the quantization process w.r.t. the iteration, which is obtained by Monte Carlo estimation. (c) MSE for trainable $\sigma_\bvarphi^2$ and various values of $\sigma_\mathrm{q}^2$ on the test set.
%   (d)-(f) The evolution of averaged probability mass of the top seven codebook elements. 
  }
  \label{fig:results_adaptive_quantization}
\vskip -0.1in
\end{figure*}

\paragraph{Dynamics of the variance parameter}
To verify whether \textit{self-annealing} happens during the training, we conduct an experiment on MNIST~\citep{lecun1998gradient}.
%To verify the Proposition~\ref{th:to_deterministic_quantization_gaussian}, 
We train Gaussian SQ-VAE with $\bm{\Sigma}_\bvarphi=\sigma_\bvarphi^2\bI$.
As targets for comparison, we also train models with $\sigma_\bvarphi^2$ \textit{fixed} to a designated $\sigma_\mathrm{q}^2$. %, i.e., $P_\bvarphi(\bz_{\mathrm{q},i}=\bb_k)\propto\exp(-\|\bb_k-\bz_i\|_2^2/2\sigma_\mathrm{q}^2)$ throughout the training.
% ($\sigma_\mathrm{q}^2=0.5$ has been adopted by~\citet{roy2018theory} and \citet{william2020hierarchical}, see Section~\ref{sec:related_works}).
The details of the experimental setup can be found in Appendix~\ref{sec:app_details_observation}. 
% The decoder and encoder are implemented using standard two-layer ConvNet architectures.
% For codebook capacity, we set $(d_b,K)=(64,128)$.
The results are summarized in Figure~\ref{fig:results_adaptive_quantization}. 
% The results on Fashion-MNIST and CIFAR10 are deferred to Appendix~\ref{sec:app_details_observation}.
%First, we plot variances $\sigma^2$ and $\sigma_\bvarphi^2$ in

In Figure~\ref{fig:results_adaptive_quantization}(a), as the epoch grows, $\sigma_\bvarphi^2$ decreases along with $\sigma^2$, which agrees with Proposition~\ref{th:to_deterministic_quantization_gaussian} and our expectation.
%In Figure~\ref{fig:results_adaptive_quantization}(b), we show the average entropy of $\hat{P}_\bvarphi(\bz_{\mathrm{q},i}|\bZ)$, i.e., $\E_{\pdata(\bx)p_\bvarphi(\bZ|g_\bphi(\bx))}[\hat{P}_\bvarphi(\bz_{\mathrm{q},i}|\bZ)]$, w.r.t. the iterations.
As shown in Figure~\ref{fig:results_adaptive_quantization}(b), with trainable $\sigma_\bvarphi^2$, the average entropy decreases as the training progresses. These two results suggest that self-annealing occurs in practical situations.
%which gradually enhanced the probability for likely codebook elements.
%The results suggest that our trainable quantization achieved soft-to-hard annealing.
On the other hand, Figure~\ref{fig:results_adaptive_quantization}(b) indicates that with a fixed $\sigma_\mathrm{q}^2$, the average entropy stays relatively constant. Moreover, as shown in Figure~\ref{fig:results_adaptive_quantization}(c), the MSE is greatly affected by the $\sigma_\mathrm{q}^2$ selected. Although there is an optimum for the fixed $\sigma_\mathrm{q}^2$, the trainable $\sigma_\bvarphi^2$, which is indicated by the blue line, achieves the lowest MSE among all cases.
%Therefore, we conclude that the stochastic quantization and self-annealing effect will be less affected by initialization of codebook.
%We will further show in Section~\ref{sec:experiments} that self-annealing also encourages diverse usage of codebook, which reduce the loss of information.
Therefore, we showed that the stochastic quantization and self-annealing together yields a codebook that effectively covers a larger support in the latent space, especially in the beginning of the training stage. This leads to the improvement of reconstruction accuracy, which will be demonstrated further in Section~\ref{sec:experiments}.

%the quantization behaviors are completely different. As for $\sigma_\mathrm{q}^2\leq 0.5$, the average entropy is almost zero throughout the training, which indicates almost deterministic quantization.
% We observe that among all tested stochastic quantization models with fixed $\sigma_\mathrm{q}^2$, the one with $\sigma_\mathrm{q}^2=1.0$  achieves the lowest MSE.
% The blue line indicates the MSE value achieved by trainable $\sigma_\bvarphi^2$.

% \paragraph{Remark}

% while the trainable parameter $\sigma_\bvarphi^2$ ($\kappa_\bvarphi$) does so.

%-- subsection
\subsection{vMF SQ-VAE for Categorical Distributions}
\label{sec:sub_vmf_sqvae}

An intuitive way to adapting SQ-VAE for categorical data distribution is to model the decoder output as a categorical distribution as \beqref{eq:decoder_categorical}. 
Consider a typical classification scenario that the last layer of a decoder is a linear layer followed by a softmax. The decoder can be represented as the combination of the linear layer $\mathbf{w}_{\text{last},c}\in\realnum^F$ and the rest $\tilde{f}_{\btheta^-,d}^\text{rest}:\mathbf{B}^{d_z}\to\realnum^{F}$. It becomes $f_{\btheta,d}^c(\bZq)=\mathbf{w}_{\text{last},c}^\top\tilde{f}_{\btheta^-,d}^\text{rest}(\bZq)$, where $\btheta^-$ denotes the trainable parameters excluding $\mathbf{w}_{\text{last},c}$.
We may represent the ELBO of this model in terms of the decomposition as
% $\Ls_\text{CE-SQ}^\text{na\"ive}=$
\begin{subequations}
    \begin{align}
        % &\E_{Q_{\bomega}(\bZq|\bx)}\left[
        % -\sum_{d=1}^D\log(P_\btheta(x_d=l|\bZq))
        % +\mathcal{R}^\mathcal{N}_\bomega(\bx, \bZq)\right]\notag\\
        &\Ls_\text{CE-SQ}^\text{na\"ive}=\E_{q_\bomega(\bZ|\bx)\hat{P}_\bvarphi(\bZq|\bZ)}\left[
        -\sum_{d=1}^D\log(P_\btheta(x_d=c|\bZq))\right.\nonumber\\
        &\quad\left.\vphantom{\sum_{d=1}^D}+\mathcal{R}^\mathcal{N}_\bvarphi(\bZ, \bZq)\right]
        -\E_{q_\bomega(\bZ|\bx)}H\left(\hat{P}_\bvarphi(\bZq|\bZ)\right)+\mathrm{const.}
        \nonumber\\%\label{eq:elbo_naive_categorical_sqvae}\\
        &\text{with}~\label{eq:elbo_naive_categorical_sqvae}\\
        &
        P_\btheta(x_d=c|\bZq)=\softmax_c\left(\{\mathbf{w}_{\text{last},c^\prime}^\top\tilde{f}_{\btheta^-,d}^\text{rest}(\bZq)\}_{c^\prime=1}^{C_\mathrm{all}}\right).
        % P_\btheta(x_d=c|\bZq)=\softmax_c(\mathbf{w}_{\text{last},c}^\top\tilde{f}_{\btheta^-,d}^\text{rest}(\bZq)).
        \label{eq:decoder_naive_categorical_sqvae}
    \end{align}
\end{subequations}
However, we found that the performance of this \textit{Na\"ive categorical (NC) SQ-VAE} is often unsatisfactory, as shown in Section~\ref{sec:sub_experiments_vmf_sqvae}.
% that the naive CE SQ-VAE often leads to codebook collapse.
% does not work well for categorical distribution.
% as shown in Section~\ref{sec:sub_experiments_vmf_sqvae}, the na\"ive CE SQ-VAE does not work well for categorical distribution.
A possible cause can be found by observing the difference between \beqref{eq:elbo_gaussian_sqvae} and \beqref{eq:elbo_naive_categorical_sqvae}. In \beqref{eq:elbo_naive_categorical_sqvae}, owing to the replacement of Gaussian with categorical distribution, trainable parameters such as $\sigma^2$ no longer exist in the objective function. This means that the model cannot be benefited from the self-annealing effect. 
%It is due to that the first term is no longer scaled with a trainable parameter while $\sigma^2$ in \beqref{eq:elbo_gaussian_sqvae} weights the MSE term.
%Without such a trainable scaling parameter, we cannot expect automatic annealing of quantization from soft to hard, as shown in the end of this subsection.
% (see Appendix~\ref{sec:app_adaptive_quantization_vmf_sqvae}).

\begin{figure}[t]
\vskip 0.1in
   \centering
   \includegraphics[width=.48\textwidth]{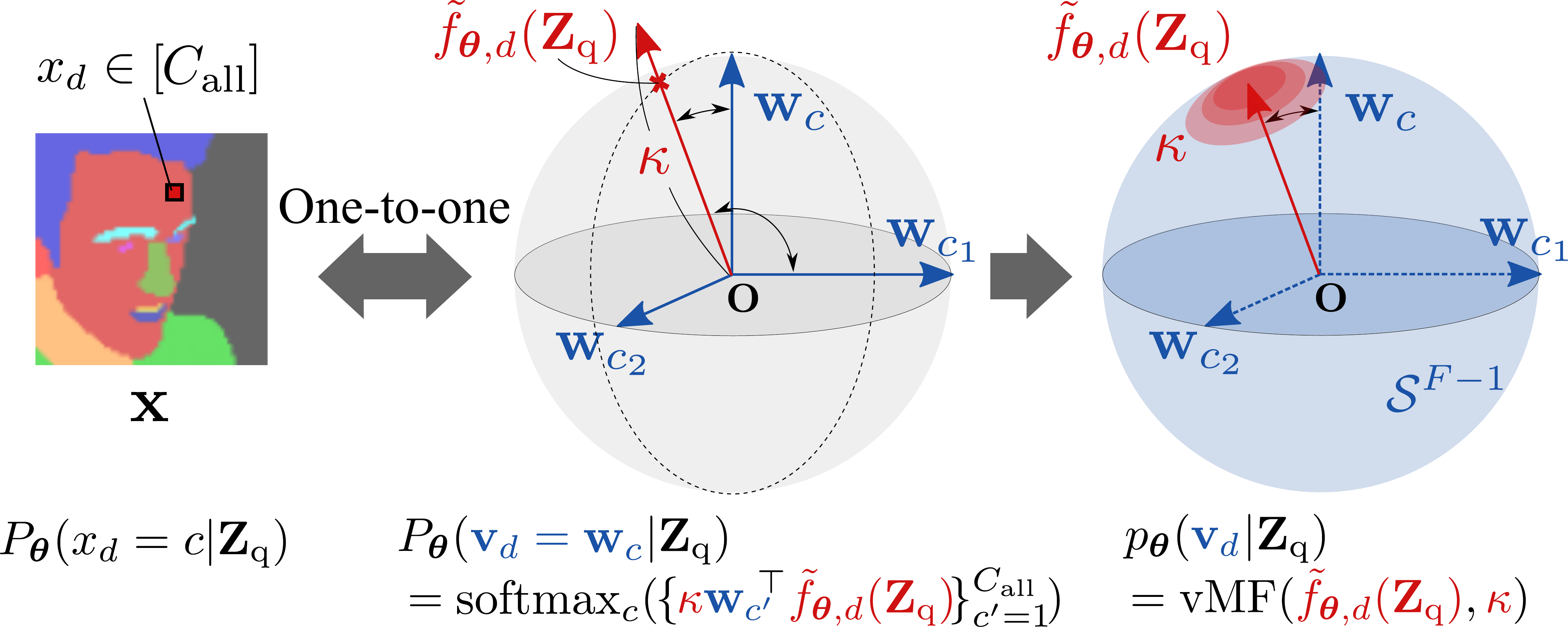}
   \caption{
   vMF decoder.
   }
   \label{fig:vmf_decoder}
\vskip -0.1in
\end{figure}
\begin{figure}[t]
\vskip 0.1in
  \centering
  \subfigure[Concentration parameters]{\includegraphics[height=0.16\textwidth]{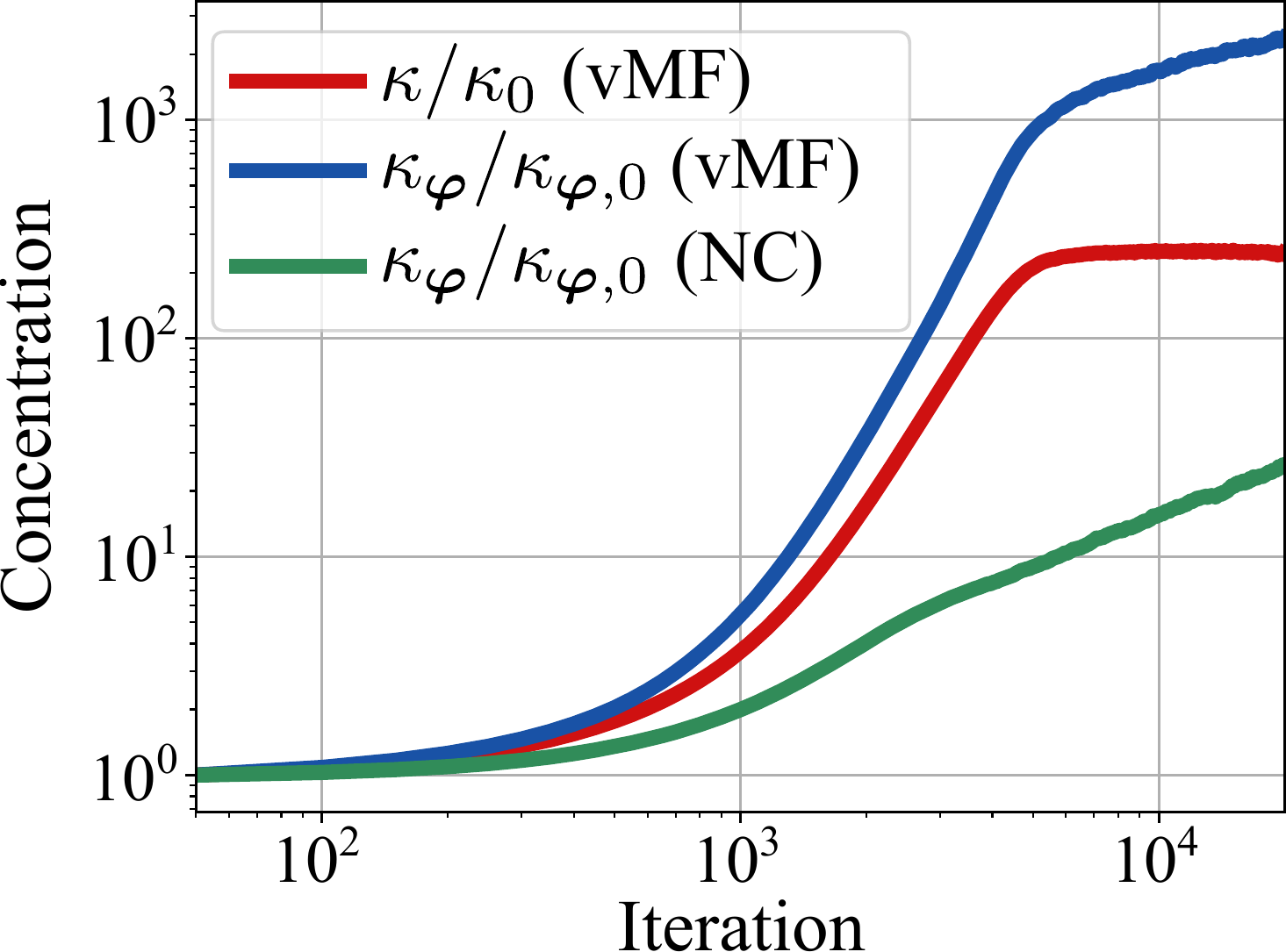}}
\hspace{10pt}%
  \subfigure[$H(\hat{P}_\bvarphi(\bz_{\mathrm{q},i}|\bZ))$]{\includegraphics[height=0.16\textwidth]{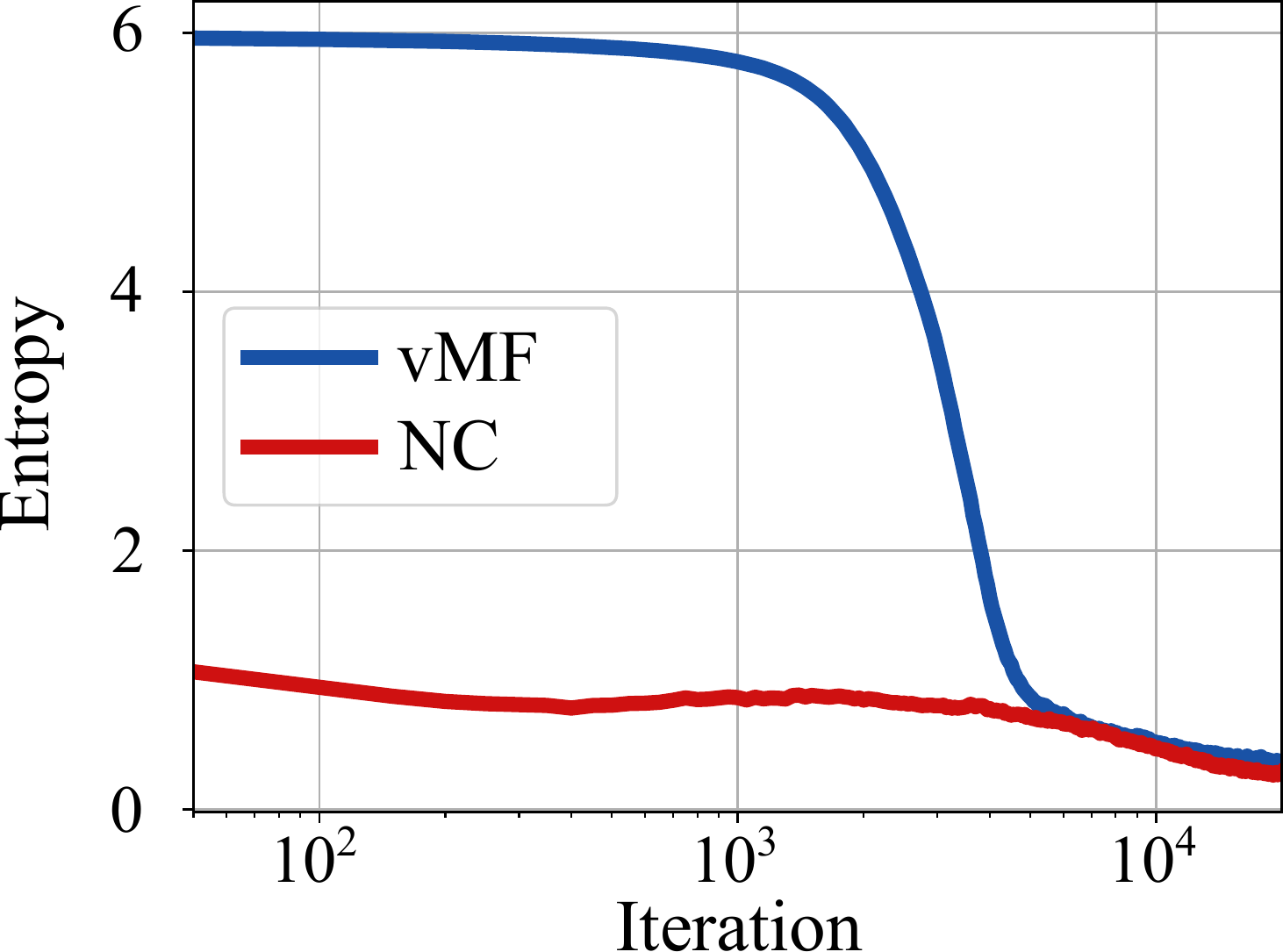}}
  \caption{Comparison between vMF and NC decoders: (a) The concentration parameter of vMF decoder $\kappa_\bvarphi$ increases with $\kappa$, whereas the growth of $\kappa_\bvarphi$ of the NC decoder is relatively small. Here, $\kappa_0$ and $\kappa_{\bvarphi,0}$ indicate initial values. (b) Average entropy of probabilities of quantization processes.
  }
  \label{fig:results_adaptive_quantization_vmf}
\vskip -0.1in
\end{figure}

To gain the advantage from self-annealing, we introduce the vMF distribution to refine the model as in Figure~\ref{fig:vmf_decoder}, and we call it vMF SQ-VAE. Consider a hypersphere $\mathcal{S}^{F-1}$ that is embedded in an $F$-dimensional space. Let $\bw_c$ denote the projection vector\footnote{
% In the current implementation, $\{\bw_c\}_{l=1}^c$ are predefined and nearly-equal distributed on $\mathcal{S}^{F-1}$. We will explore its optimization in future works.
In the current implementation, $\{\bw_c\}_{c=1}^{C_\mathrm{all}}$ are predefined and distributed on $\mathcal{S}^{F-1}$. We will explore its optimization in future work.
} 
of the $c$th data category on the surface of $\mathcal{S}^{F-1}$. Next, we represent the projection of data $x_d$ on the hypersphere as $\bv_d\in\{\bw_c\}_{c=1}^{C_\mathrm{all}}$. If $x_d$ belongs to a category $c$, that is, $x_d=c$, then $\bv_d=\bw_c$ and vice versa. 

%First, we consider a hypersphere $\mathcal{S}^{F-1}$ for representation of $x_d=l$ as in Figure~\ref{fig:vmf_decoder}, where $F$ is the dimension of the enclosed ball.
% and set to no less than $L$.
%Here, we introduce bases, denoted as $\bw_c\in\mathcal{S}^{F-1}$, to represent the target categories on the hypersphere\footnote{
%In our experiments, we set $(\bw_c)_{l=1}^c$ to a set of fixed vectors. Optimization of the bases would be tried in future works.
% For example, if the correlation of the categories are unknown is unknown to fixed one-hot vectors that are orthonormal each other with $F=L$. 
%}.
%Next, we represent $x_d$ with a vector $\bv_d\in\{\bw_c\}_{l=1}^c$ such that $\bv_d=\bw_c$ indicates $x_d=l$ and vice versa.

\paragraph{Decoding}
The first step is to decode $\bZq$ into $\bV:=\{\bv_d\}_{d=1}^D$ with the decoder $\tilde{f}_{\btheta,d}:\mathbf{B}^{d_z}\to\mathcal{S}^{F-1}$. Then, determine the probability of $\bv_d=\bw_c$ with a trainable scalar $\kappa\in\realnum_+$ by using
%To recover $\bx$ from $\bZq$ via $\bV:=\{\bv_d\}_{d=1}^D$ with the one-to-one correspondence, we prepare decoding functions $\tilde{f}_{\btheta,d}:\mathbf{B}^{d_z}\to\mathcal{S}^{F-1}$.
%The probability of $\bv_d=\bw_c$, i.e., $x_d=l$, with $\bZq$ given are modeled with the decoding function and a trainable scalar parameter $\kappa\in\realnum_+$ as
% \begin{align}
%     P_{\bm{\theta}}(\bv_d=\bw_c|\bZq)=\softmax_c\left(\kappa\bw_c^\top{}\tilde{f}_{\btheta,d}(\bZq)\right),
%     \label{eq:decoder_vmf_prepare}
% \end{align}
\begin{align}
    P_{\bm{\theta}}(\bv_d=\bw_c|\bZq)=\softmax_c\left(\left\{\kappa\bw_{c^\prime}^\top{}\tilde{f}_{\btheta,d}(\bZq)\right\}_{c^\prime=1}^{C_\mathrm{all}}\right),
    \label{eq:decoder_vmf_prepare}
\end{align}
which resembles the categorical decoder in \beqref{eq:decoder_naive_categorical_sqvae} except for the normalization onto $\mathcal{S}^{F-1}$ and the scaling factor $\kappa$.
Therefore, we may represent the categorical probabilities for the decoded $\bZq$ as 
%Finally, substituting $w_c$ with $bv_d$ \beqref{eq:decoder_vmf_prepare} with the von Mises-Fisher distribution leads to:
%Finally, we relax the categorical variables $\bv_d\in\{\bw_c\}_{l=1}^c$ into the continuous variables on the unit hypersphere~$\mathcal{S}^{F-1}$.
%The relaxation leads to the following continuous distribution:
\begin{align}
    p_{\bm{\theta}}(\bv_d|\bZq)\propto\exp\left(\kappa\bv_d^\top{}\tilde{f}_{\btheta,d}(\bZq)\right).
    \label{eq:decoder_vmf}
\end{align}
By normalizing \beqref{eq:decoder_vmf} w.r.t. $\bv_d$ over $\mathcal{S}^{F-1}$, we obtain  $p_\btheta(\bv_d|\bZq)=\mathrm{vMF}(\tilde{f}_{\btheta,d}(\bZq),\kappa)$, where $\tilde{f}_{\btheta,d}(\bZq)$ and $\kappa$ correspond to the mean direction and the concentration parameter of the vMF distribution, respectively.
% The connection between the usual categorical decoder in \beqref{eq:elbo_naive_categorical_sqvae} and this vMF decoder is described in Appendix~\ref{sec:app_categorical_vs_vmf_decoder}.
% In addition, the vMF decoder can be applied to VAE (see Appendix~\ref{sec:app_apply_vmf_decoder_vae}).

% by transforming usual CE loss.
% Let us consider a neural decoder with a linear final layer $\bW_\text{final}$, i.e., $f_{\btheta}(\bZq)=\bW_\text{final}\tilde{f}_{\btheta^-}(\bZq)$, where $\btheta^-$ denotes the neural network weight except for the final layer.
% Under this setting, the probability of $\bv_i$ is usually modeled with the softmax operator as
% \begin{align}
%     P(\bv_i=\bw_c|\bZq)=\mathrm{\softmax}_c\left(\bw_c^\top\tilde{f}_{\btheta^-,i}(\bZq)\right),
%     \label{eq:decoder_usual_cel}
% \end{align}
% which is equivalent to \beqref{eq:decoder_categorical}.
% Inspired from \beqref{eq:decoder_usual_cel}, 

\paragraph{Encoding}
Accordingly, we model the stochastic dequantization process of the encoder with the vMF distribution:
\begin{align}
    p_{\bvarphi}(\bz_i|\bZq)=\mathrm{vMF}(\bz_{\mathrm{q},i},\kappa_\bvarphi),
\end{align}
where $\kappa_\bvarphi$ is the trainable concentration parameter\footnote{$\kappa_\bvarphi$ can be dependent on either $\bx$ or $i$. However, we choose $\kappa_\bvarphi$ to be the independent variable.}.
Similarly to Gaussian SQ-VAE in Section~\ref{sec:sub_gaussian_sqvae}, the discrete $\bZq$ is recovered using Bayes' theorem as %$P_\bvarphi(\bz_{\mathrm{q},i}=\bb_k|\bZ)=$
\begin{align}
    \hat{P}_\bvarphi(\bz_{\mathrm{q},i}=\bb_k|\bZ)=
    \softmax_{k}\left(\{\kappa_{\bvarphi}\bb_j^\top\bz_i\}_{j=1}^K\right),
    \label{eq:quantization_vmf}
\end{align}
where the unnormalized log-probabilities of $\bb_k$ in \beqref{eq:quantization_vmf} correspond to the $\kappa_{\bvarphi}$-scaled cosine similarity between $\bb_k$ and $\bz_i$.

\paragraph{Objective Function}
Substituting the encoding and decoding processes into \beqref{eq:elbo_sqvae} leads to
$\Ls_\text{vMF-SQ}=$
\begin{align}
    &\E_{q_\bomega(\bZ|\bx)\hat{P}_\bvarphi(\bZq|\bZ)}\left[-\kappa\sum_{d=1}^D\bv_d^\top{}\tilde{f}_{\btheta,d}(\bZq)
    +\mathcal{R}_\bvarphi^\mathrm{vMF}(\bZ,\bZq)\right]\notag\\
    &\quad-\E_{q_\bomega(\bZ|\bV)}H\left(\hat{P}_\bvarphi(\bZq|\bZ)\right)-\log{}C_F(\kappa)+\mathrm{const.},
    \label{eq:elbo_vmf_sqvae}
\end{align}
where $\mathcal{R}_\bvarphi^\mathrm{vMF}(\bx,\bZq)$ is a regularization objective defined by $\mathcal{R}_\bvarphi^\mathrm{vMF}(\bZ,\bZq)=\sum_{i=1}^{d_z}\kappa_{\bvarphi,i}(1-\bz_{\mathrm{q},i}^\top{}\bz_i)$ (see Appendix~\ref{sec:app_sub_elbo_vmf_sqvae} for details).
Here, $C_F(\kappa)$ denotes the normalizing constant of the vMF distribution (see Appendix~\ref{sec:app_notations}).

\paragraph{Comparing vMF SQ-VAE with \textit{Na\"ive} Categorical SQ-VAE}
In \beqref{eq:elbo_vmf_sqvae}, the first two terms are scaled with $\kappa$ and $\kappa_\bvarphi$. Furthermore, vMF SQ-VAE has a property
that, if $\kappa\to\infty$, then $\kappa_{\bvarphi^*}\to\infty$. Its proof can be done similarly to Proposition~\ref{th:to_deterministic_quantization_gaussian} via setting $\kappa=1/\sigma^2$ and $\kappa_{\bvarphi^*}=1/\sigma_{\bvarphi^*}^2$. %\begin{proposition}
%    Assuming that $\pdata(\bx)$ has finite support, while $g_{\bphi}$ and $\{\bb_k\}_{k=1}^K$ are bounded.
%    Suppose $\bomega^{*}=\{\bphi^*,\bvarphi^*\}$ to be a minimizer of $\E_{\pdata(\bx)}\KL(Q_{\bomega}(\bZq|\bx)\parallel{}P_\btheta(\bZq|\bx))$ with $\btheta$, $\sigma^2$ and $\{\bb_k\}_{k=1}^K$ fixed.
%    If $\kappa\to\infty$, then $\kappa_{\bvarphi^*}\to\infty$.
%    \label{th:to_deterministic_quantization_vmf}
%\end{proposition}
%According to Proposition~\ref{th:to_deterministic_quantization_vmf}, 
As a result, vMF SQ-VAE can also achieve self-annealing as described in Section~\ref{sec:sub_behavior_quantization} if $\kappa\to\infty$. In the experiment on CelebAHQ-Mask~\citep{lee2020maskgan} in Section~\ref{sec:sub_experiments_vmf_sqvae}, $\kappa_\bvarphi$ increases together with $\kappa$ as training progresses, as shown in Figure~\ref{fig:results_adaptive_quantization_vmf}. On the other hand, self-annealing is impossible for NC SQ-VAE owing to the lack of scaling parameters.

\begin{figure*}[t]
\vskip 0.1in
  \centering
  \subfigure[MNIST ($d_b=64$)]{\includegraphics[width=0.27\textwidth]{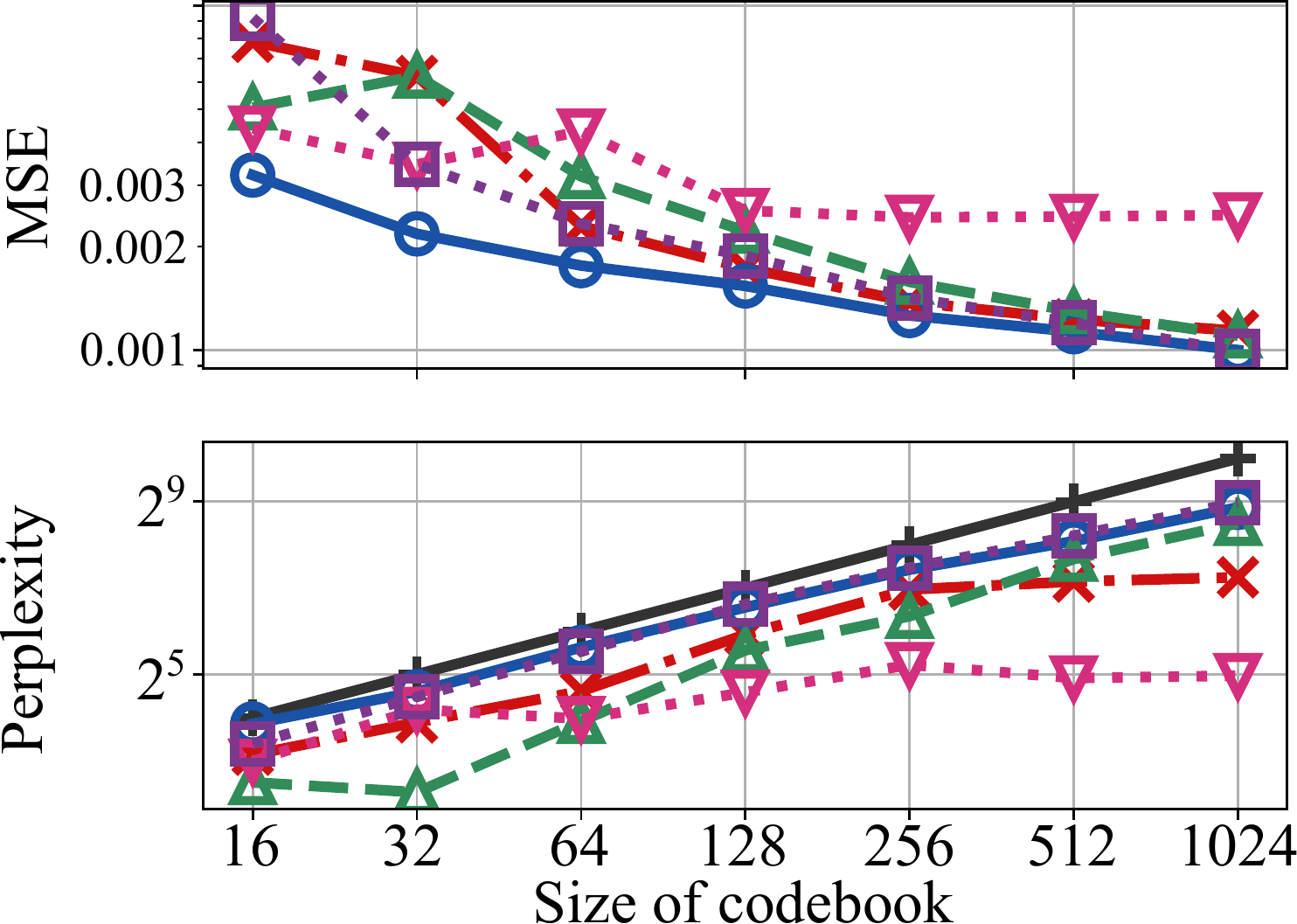}}
  \hspace{5pt}%
  \subfigure[Fashion-MNIST ($d_b=64$)]{\includegraphics[width=0.27\textwidth]{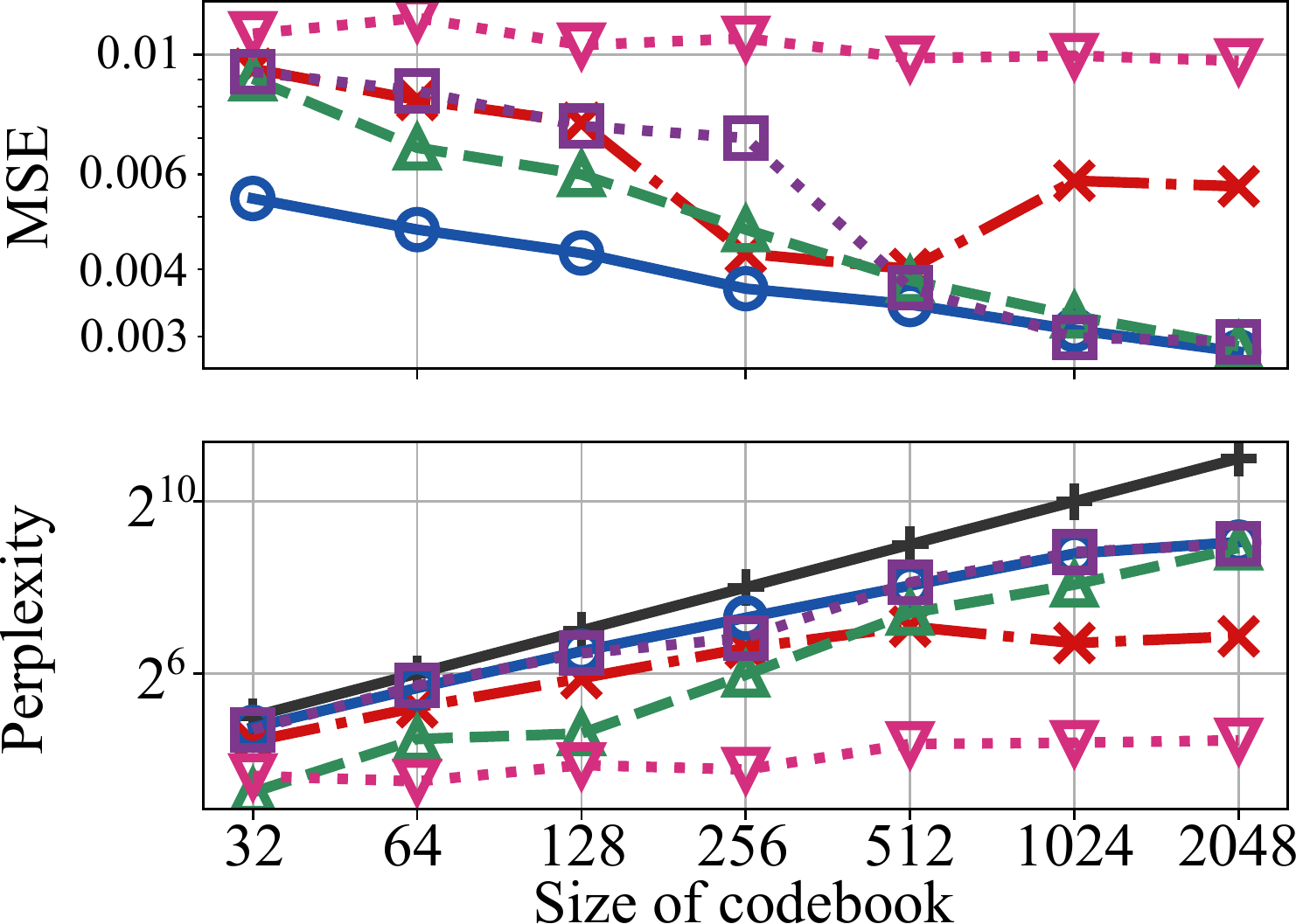}}
  \hspace{5pt}%
  \subfigure[CIFAR10 ($d_b=64$)]{\includegraphics[width=0.27\textwidth]{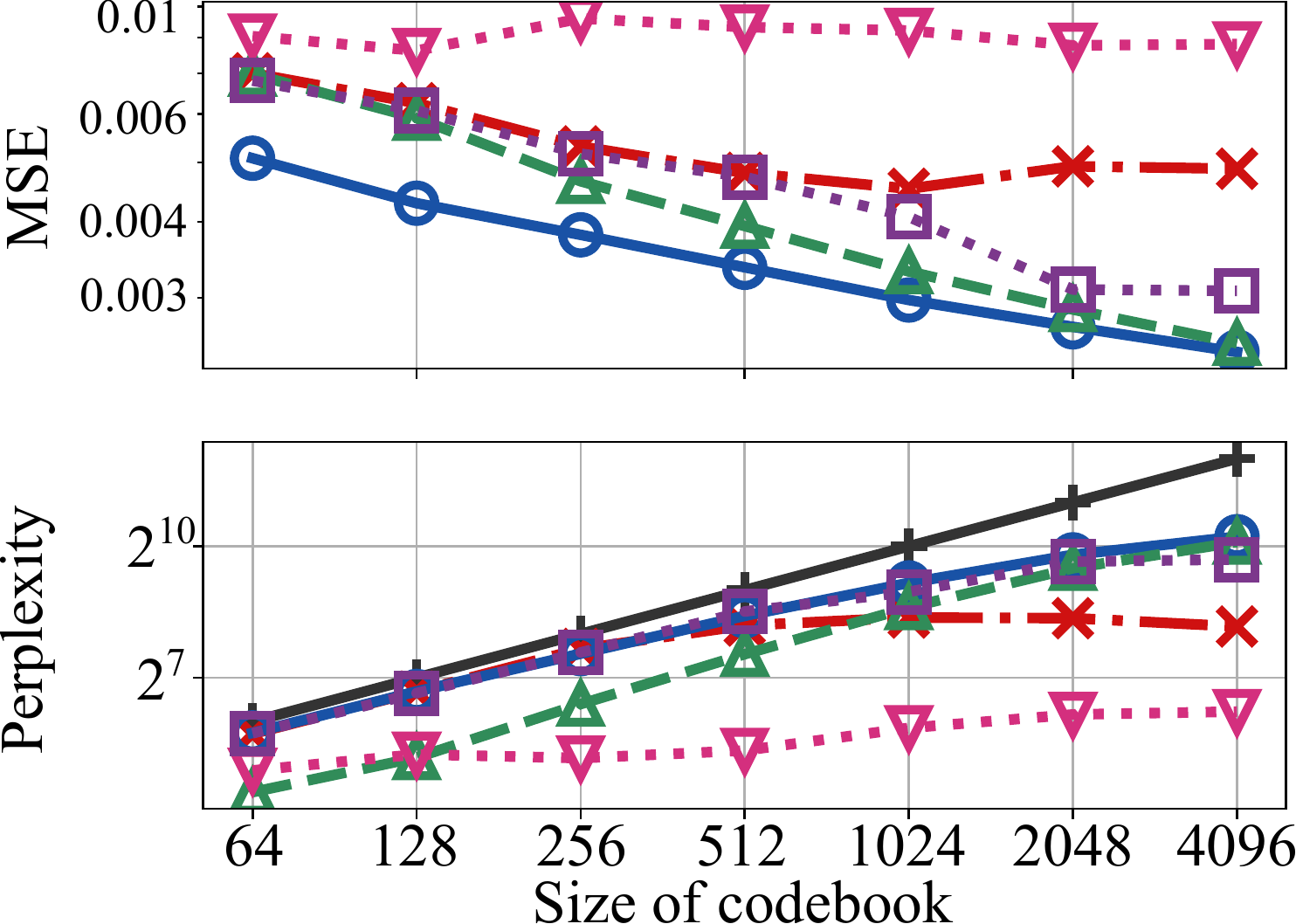}}
  \hspace{1pt}%
  \subfigure{\includegraphics[width=0.15\textwidth]{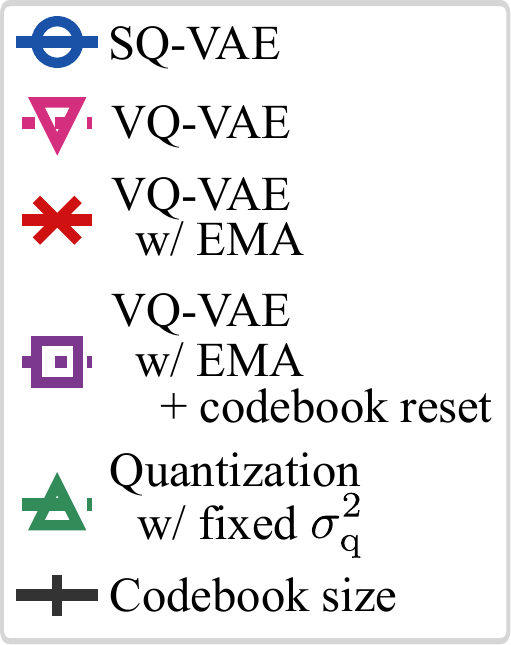}}
\\
%   \stepcounter{figure}
%   \setcounter{caption@flags}{4}
%   \setcounter{subfigure}{3}
  \addtocounter{subfigure}{-1}
  \subfigure[MNIST ($K=128$)]{\includegraphics[width=0.27\textwidth]{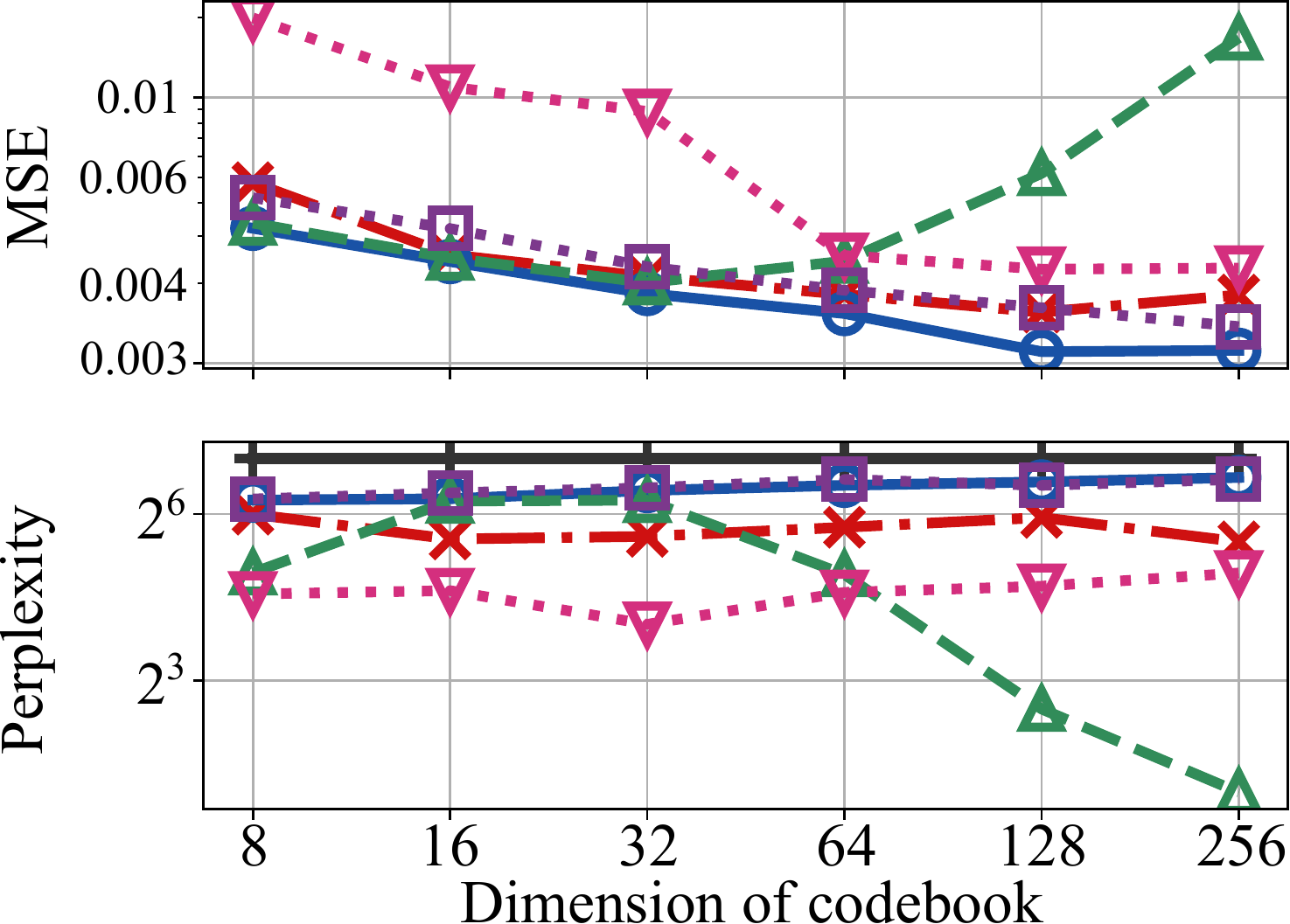}}
  \hspace{5pt}%
  \subfigure[Fashion-MNIST ($K=256$)]{\includegraphics[width=0.27\textwidth]{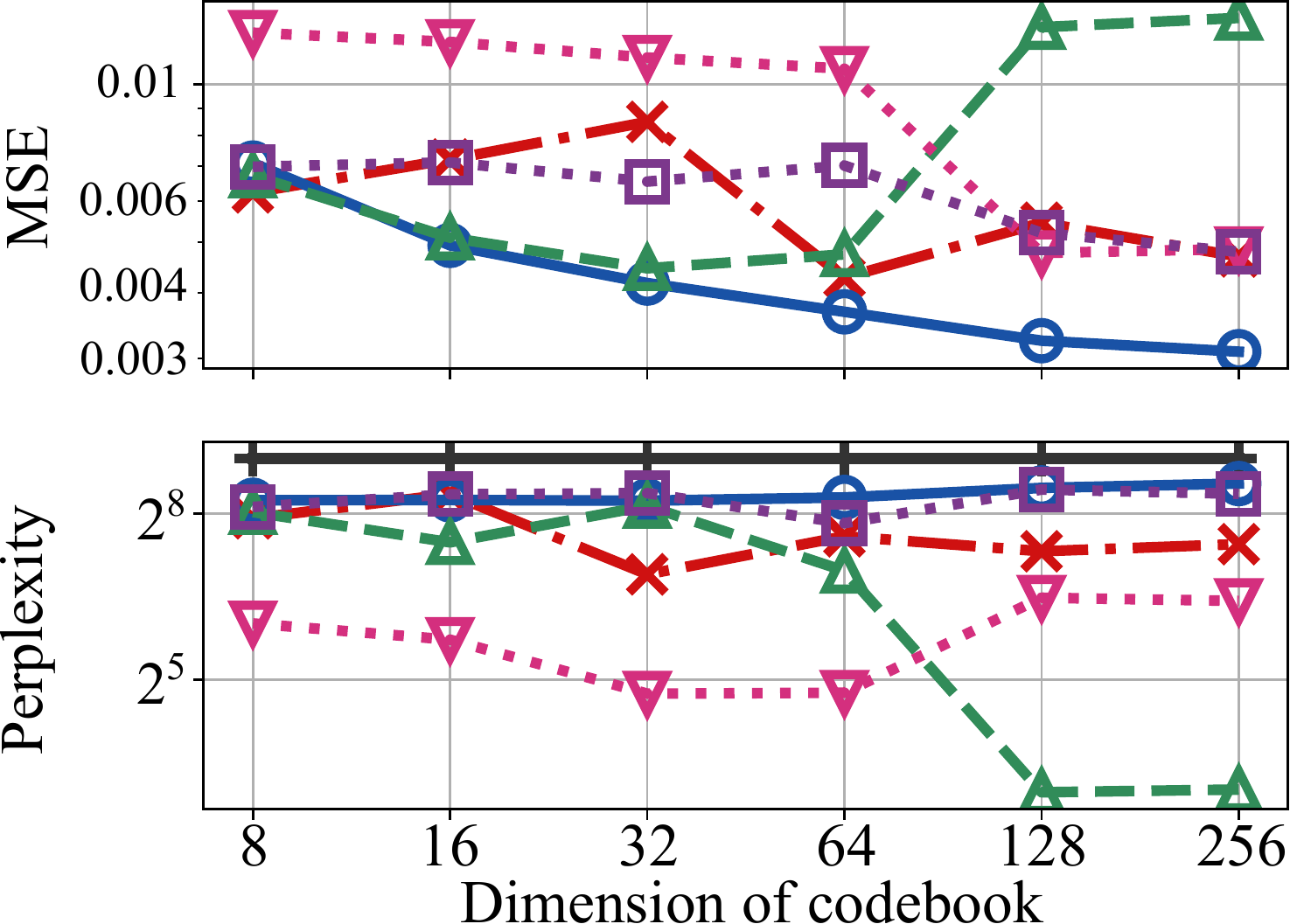}}
  \hspace{5pt}%
  \subfigure[CIFAR10 ($K=512$)]{\includegraphics[width=0.27\textwidth]{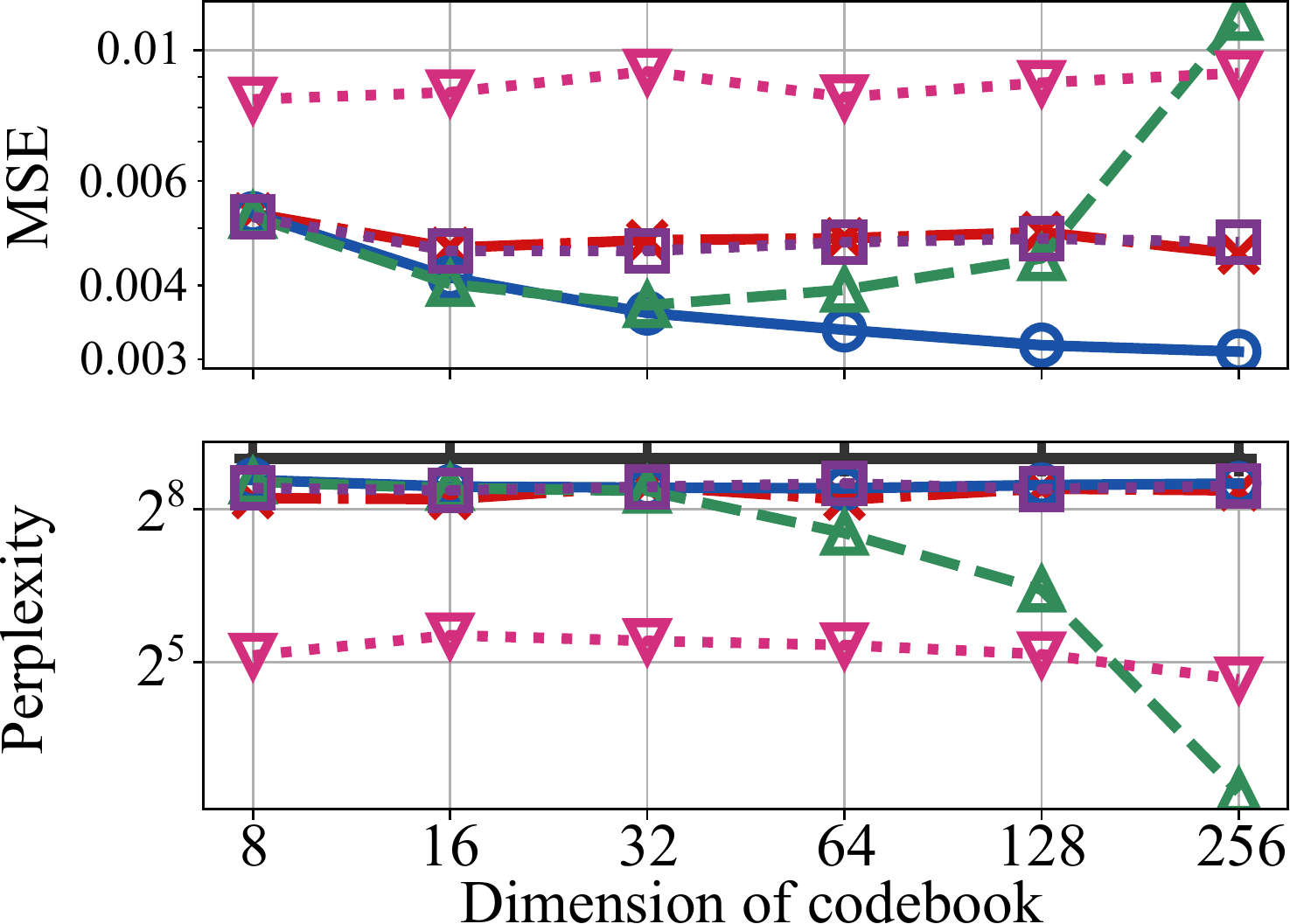}}
  \hspace{1pt}%
  \subfigure{\includegraphics[width=0.15\textwidth]{figures/results/legends.pdf}}
  \caption{
%   Empirical studies on the impact of codebook capacity. (a)-(c) The size $K$ is swept with the dimension $d_b$ fixed to $64$. (d)-(f) Various $d_b$ is tried with the size $K$ fixed to $128$, $256$ and $512$ for MNIST, Fashion-MNIST and CIFAR10, respectively. The black lines with ``$+$'' marks indicate upper bounds of the perplexities, i.e., $K$.
  Empirical studies on the impact of codebook capacity examined on  MNIST Fashion-MNIST and CIFAR10. (a)--(c) The size $K$ is swept with the dimension $d_b$ fixed to $64$. (d)--(f) Various $d_b$ values are tested with the size $K$ fixed as $128$, $256$, and $512$, respectively. The black lines with ``$+$'' marks indicate the upper bounds of the perplexities, i.e., $K$. All the y-axes are in log-scale.
  }
  \label{fig:results_codebook_size}
\vskip -0.1in
\end{figure*}

%%----- Section
\section{Related Work}
\label{sec:related_works}

\paragraph{Latent Vector Quantization}
This is a common approach in VQ-VAE, end-to-end image compression~\citep{toderici2016variable,theis2017lossy} and many other studies.
\citet{van2017neural} trained VQ-VAE with deterministic quantization.
They approximated the gradient of the quantization process with the straight-through estimator~\citep{bengio2013estimating} and utilized the stop-gradient operator. 
Some VQ-VAE-based models also adopted deterministic quantization~\citep{razavi2019generating, dhariwal2020jukebox}.

The common form of stochastic quantization $Q(\bz_{\mathrm{q},i}=\bb_k)\propto\exp(-\|\bb_k-\bz_i\|_2^2)$ has been applied in several VQ-VAE-based models~\citep{roy2018theory, william2020hierarchical}.
\citet{roy2018theory} connected the EMA update of the codebook with an expectation maximization (EM) algorithm and softened the EM algorithm with a stochastic posterior.
\citet{william2020hierarchical} adopted a hierarchical VQ-VAE with a stochastic posterior to compress images at extremely low bit rates.
% They reformulated VQ-VAE as VAE that has a continuous latent space with its prior Gaussian mixture model.
% The stochastic quantization schemes in these studies do not involve trainable parameters in their categorical posterior, which correspond to the fixed schemes in Section~\ref{sec:sub_behavior_quantization}, where we showed our categorical posterior that includes trainable parameters achieved better reconstruction accuracy by inducing self-annealing.
The stochastic quantization schemes in previous literature did not involve trainable parameters in their categorical posterior, which correspond to the schemes with fixed $\sigma_\mathrm{q}^2$ in Section~\ref{sec:sub_behavior_quantization}. We emphasize the fact that our categorical posterior including trainable parameters achieves a higher reconstruction accuracy with the help of self-annealing.

\citet{agustsson2017soft} proposed a controllable quantization scheme for image compression. In their work, annealing is achieved with a predefined hyperparameter scheduling.
% The hyperparameter is manually annealed according to the training iteration. 
However, they mentioned that the annealing rate must be controlled carefully, otherwise, either the annealing cannot progress or the gradient will vanish in the early stage. In contrast, our proposed method does not rely on such tuning.

\citet{wu2020vector} proposed a bottleneck regularizer of the latent space in VAE to extract meaningful representations for downstream tasks. Although the regularizer is \textit{inspired from} the stochastic quantization of latent variables, how the method relates continuous latent features with a discrete codebook is beyond the scope of this work.

\paragraph{Learning Categorical Distributions with VAE}
% Although there are many decoders designed for categorical distributions, their variations are limited.
\citet{kingma2013auto} first proposed a VAE with the Bernoulli decoder for binary data.
% , which leads to binary cross entropy loss for reconstruction.
The Bernoulli decoder can be easily generalized into a multi-class categorical decoder as in \beqref{eq:decoder_categorical}. This categorical decoder has been adopted to the VQ-VAE by~\citet{chorowski2019unsupervised}.

\citet{polykovskiy2020deterministic} proposed a deterministic decoding scheme for discrete data. The categorical probability of its decoder has been limited to be one-hot by applying an argmax operator. The non-differentiability of argmax is solved by the smooth relaxation of the argmax operator with a temperature parameter. In practice, the hyperparameter is manually annealed to $0$ in the training phase. 
By comparing vMF SQ-VAE with this work, we found that the scaling parameter $\kappa$ is self-annealed in the training without a predefined scheduling scheme.

%Similarly to the temperature, in vMF decoder, an extreme case of $\kappa\to\infty$ indicates deterministic decoding with the argmax.
%An advantage of vMF decoder is that $\kappa$ can be optimized with variational inference instead of manual scheduling.

%%----- Section
\section{Experiments}
\label{sec:experiments}
We apply SQ-VAE in several vision- and speech-related tasks to demonstrate its improvement over the conventional VQ-VAE and VAE.
% with EMA, simply denoted as VQ-VAE hereafter, and VAE.
% All the experiments are repeated with at least 3 different random seeds.
All the experiments are repeated with three different random seeds, unless otherwise stated.
Moreover, all the models including VAE are adapted to ensure that they all use the same amount of bits to represent an encoded input. 
In our experiments, we approximate the categorical distributions $\hat{P}_\bvarphi(\bZq|\bZ)$ included in the expectation operator of \beqref{eq:elbo_general} with Gumbel--softmax distributions.
% for reparameterization trick. 
The gap between the two distributions is gradually reduced by annealing the temperature parameter of Gumbel--softmax using the predefined  schedule in~\citet{jang2017categorical}.
%It should be noted that 
This temperature parameter does not affect the categorical probabilities of the original distribution. The effect of $\sigma_\bvarphi^2$ (or $\kappa_\bvarphi$) is definitely different from that of this temperature parameter (see Section~\ref{sec:sub_behavior_quantization}).
We refer to Appendix~\ref{sec:app_details_experiments} for the detailed setup of evaluation and supplemental experiments.

%- subsubsection
\subsection{Continuous Data Distribution}
\label{sec:sub_experiments_gaussian_sqvae}

\begin{table*}
\centering
    \caption{Evaluation on CelebA. The MSE ($\times10^3$) and reconstructed FID (rFID) are evaluated using the test set. The codebook capacity for the discrete latent space is set to $(n_b,k)=(64,512)$. The Roman numerals for Gaussian SQ-VAEs correspond to those in Table~\ref{tb:parameterization_variance}. We also show the FID of samples generated with a prior learned with PixelCNN.}
    \vskip 0.1in
    \renewcommand{\arraystretch}{1.1}
    % \resizebox{\columnwidth}{!}{
    \small
    \begin{tabular}{l|c|c|c|c|c|c|c}
        \bhline{0.8pt}
        \multirow{2}{*}{Model} & \multicolumn{2}{c|}{Reconstruction} & Generation & \multicolumn{4}{c}{Latent manipulation (FID)} \\
        \cline{2-3}
        \cline{5-8}
         & MSE & rFID & (FID) & Neighbor-3 & Neighbor-5 & Neighbor-10 & Interp. \\
        \bhline{0.8pt}
        VAE & 4.79 $\pm$ 0.01 & 40.3 $\pm$ 0.3 & -- & -- & -- & -- & -- \\
        VQ-VAE (EMA) & 1.33 $\pm$ 0.41 & 18.5 $\pm$ 5.1 & 42.0 $\pm$ 11.5 & 31.9 $\pm$ 14.8  & 42.8 $\pm$ 20.7 & 70.7 $\pm$ 35.4 & 28.2 $\pm$ 6.4 \\
        VQ-VAE (EMA+code reset) & 1.62 $\pm$ 0.36 & 22.0 $\pm$ 5.9 & 51.8 $\pm$ 10.8 & 39.7 $\pm$ 12.0 & 52.7 $\pm$ 14.7 & 83.2 $\pm$ 20.4 & 32.6 $\pm$ 7.1 \\
        Quantization w/ fixed $\sigma_\mathrm{q}^2$ & 1.09 $\pm$ 0.01 & 15.9 $\pm$ 0.1 & 38.2 $\pm$ 0.9 & 20.0 $\pm$ 0.4 & 26.4 $\pm$ 0.8 & 41.5 $\pm$ 2.1 & 18.6 $\pm$ 0.3 \\
        \hline
        Gaussian SQ-VAE (\Rnum{1}) & \textbf{0.96} $\pm$ 0.01 & 14.8 $\pm$ 0.3 & 28.2 $\pm$ 0.9 & 17.8 $\pm$ 0.1 & 21.9 $\pm$ 0.1 & \textbf{33.1} $\pm$ 0.3 & \textbf{17.6} $\pm$ 0.6 \\
        Gaussian SQ-VAE (\Rnum{2}) & 0.98 $\pm$ 0.01 & 14.3 $\pm$ 0.2 & \textbf{27.7} $\pm$ 1.1 & 17.8 $\pm$ 0.2 & 22.2 $\pm$ 0.4 & 34.0 $\pm$ 0.9 & \textbf{17.6} $\pm$ 0.1 \\
        Gaussian SQ-VAE (\Rnum{3}) & \textbf{0.96} $\pm$ 0.00 & \textbf{13.9} $\pm$ 0.1 & 28.1 $\pm$ 0.3 & \textbf{17.3} $\pm$ 0.2 & \textbf{21.6} $\pm$ 0.3 & 33.5 $\pm$ 0.6 & 18.5 $\pm$ 0.4 \\
        \bhline{0.8pt}
    \end{tabular}
    % }
    \label{tb:vision_continuous}
    \vskip -0.1in
\end{table*}

\paragraph{Vision}
First, we compare the reconstruction accuracy and codebook utilization of SQ-VAE (\Rnum{1}) with VQ-VAEs~\citep{van2017neural} under different codebook capacity settings, i.e., the codebook dimension $d_b$ and the codebook size $K$. The reconstruction accuracy is measured by MSE, and the codebook utilization is measured by the perplexity of latent variables. This experiment is performed on MNIST, Fashion-MNIST~\citep{xiao2017fashion} and CIFAR10~\citep{krizhevsky2009learning}.
As a target for comparison, we also train models with fixed stochastic quantization $P_\bvarphi(\bz_{\mathrm{q},i}=\bb_k)\propto\exp(-\|\bb_k-\bz_i\|_2^2/2\sigma_\mathrm{q}^2)$~\citep{roy2018theory,william2020hierarchical}.
We choose $\sigma_\mathrm{q}^2=1.0$, which achieves the best MSE with $K=64$ in our preliminary results (see Section~\ref{sec:sub_behavior_quantization}).
In Figure~\ref{fig:results_codebook_size}, SQ-VAE achieved the lowest MSE with the highest perplexity in most of settings. Moreover, its performance is proportional to the size and dimension of the codebook.
On the other hand, the performance of VQ-VAE is less correlated with the codebook settings and therefore suggests the need for careful tuning. Moreover, even when all the heuristic techniques are applied to VQ-VAE, SQ-VAE still outperforms VQ-VAE in terms of MSE, especially when the codebook size is small.

% We also test SQ-VAE with various parameterizations on CelebA 64$\times$64~\citep{liu2015deep} with the codebook capacity set to $(n_b,K)=(64, 512)$.
We also test SQ-VAE with various parameterizations on CelebA 64$\times$64~\citep{liu2015deep} with the codebook capacity set to $(n_b,K)=(64, 512)$.\footnote{We test SQ-VAE (\Rnum{1}--\Rnum{3}) models in this experiment because the training of SQ-VAE (\Rnum{4}) model is unstable on the CelebA dataset.}
A separately trained PixelCNN~\citep{van2016pixel} is used for each model. To keep a fair comparison, these PixelCNNs are trained such that they have the same test log-likelihood. MSE and Fr\'{e}chet Inception (FID)~\citep{heusel2017gans} are used as the metrics for quality assessment. 
To examine the feasibility of the latent space, in addition to the reconstruction and the generation, we apply latent manipulations to latent variables and evaluate the FID of the images reconstructed with modified latent variables. The first manipulation is replacing latent vector elements by the $k$th nearest codebook elements, recorded as Neighbor-$k$.
The second manipulation is that we apply linear interpolation of pairs of encoded points then project the interpolated vectors to their nearest codes. Their mixing ratios are randomly distributed within $[0,1]$.
If the codebook elements are evenly distributed within the support of the dataset in the latent space, the resulting FID would be less affected by these manipulations.

From Table~\ref{tb:vision_continuous}, SQ-VAE achieves the best performance against all the baselines. It also shows that the training of SQ-VAE is stable even for complicated parameterization such as Gaussian SQ-VAE (\Rnum{3}).
On the other hand, the simplest Gaussian SQ-VAE (\Rnum{1}) still yields performance improvement over the fixed stochastic quantization scheme.
% Unfortunately, the training of SQ-VAE (\Rnum{4}) model is unstable on the CelebA dataset.
% However, such complicated parameterization is rarely used in practice.
% However, from the best of our knowledge, such complicated parameterization has never been used in practice.
%the results in the left four metrics suggest that Gaussian SQ-VAE (\Rnum{3}), which is the most complicated parameterization of $\bm{\Sigma}_\bvarphi$, can learn the vicinity of encoded points well by adapting the variance for the respective points. On the other hand, SQ-VAE (\Rnum{1}) can learn wider range of latent space well with the simplest parameterization.
In addition, the result of applying Gaussian SQ-VAE to the CelebA HQ 256$\times$256 dataset~\citep{karras2018progressive} is shown in Appendix~\ref{sec:app_experiment_celebahq}.

\begin{table}
\centering
    \caption{Evaluation on VCTK and ZeroSpeech 2019. The MSE (dB$^2$) of sample reconstruction is evaluated using the test set. We do not apply SQ-VAE (\Rnum{2}) in this evaluation because of the variable length property of speech data and the different manipulations of speech signals between training and inference (see Appendix~\ref{sec:app_sub_details_gaussian_speech}).} 
    \vskip 0.1in
    \small
    \begin{tabular}{l|c|c}
        \bhline{0.8pt}
        \multirow{2}{*}{Model} & \multicolumn{2}{c}{MSE (dB$^2$)} \\
        \cline{2-3}
         & VCTK & ZeroSpeech 2019 \\
        \bhline{0.8pt}
        VQ-VAE w/ EMA & 29.59 $\pm$ 0.25 & 34.33 $\pm$ 1.57 \\
        \hline
        Gaussian SQ-VAE (\Rnum{1}) & 25.52 $\pm$ 0.08 & 33.17 $\pm$ 1.11 \\
        Gaussian SQ-VAE (\Rnum{3}) & 25.94 $\pm$ 0.22 & 34.35 $\pm$ 1.07 \\
        Gaussian SQ-VAE (\Rnum{4}) & \textbf{24.68} $\pm$ 0.21 & \textbf{32.32} $\pm$ 0.88 \\
        \bhline{0.8pt}
    \end{tabular}
    \label{tb:speech_mse_result}
    \vskip -0.1in
\end{table}

\paragraph{Speech}
In this experiment, we test SQ-VAE and VQ-VAE by the reconstruction of the normalized log-Mel spectrogram using two speech datasets: VCTK version 0.80~\citep{veaux2017cstr} and ZeroSpeech 2019 English~\citep{dunbar19zero}.
%(Please refer to Appendix~\ref{sec:app_sub_details_gaussian_speech} for the detail).
% We adopted \citet{niekerk2020vector}'s VQ-VAE model\footnote{\url{https://github.com/bshall/ZeroSpeech}} as the baseline and replaced its RNN-based vocoder with a projection layer.
We adopt the VQ-VAE model of \citet{niekerk2020vector} as the baseline and replace its RNN-based vocoder with a projection layer.
The codebook dimension $d_b$ and the codebook size $K$ are set to $64$ and $512$, respectively.
For VQ-VAE, the first term of \beqref{eq:objective_vq}, $-\log{}p_\btheta(\bx|\bZq)$, is set as $\|\bx-f_\btheta(\bZq)\|_2^2/2\sigma^2$, following \citet{eloff2019unsupervised}.
%In this case, $\sigma^2$ is a hyperparameter that weights reconstruction error.
The hyperparameter $\sigma^2$ is determined by a grid search with $\sigma^2=\{10^{-2}, 10^{-1}, 10^{0}, 10^{1}\}$.
In this experiment, we run each experiment with five different random seeds and report the average and standard deviation of MSE values.
%The MSE results are shown in Table~\ref{tb:speech_mse_result}.
As shown in Table~\ref{tb:speech_mse_result}, SQ-VAE achieves better average MSE than VQ-VAE even though VQ-VAE has been tuned with the hyperparameter $\sigma^2$. %It means that the SQ-VAE can model speech spectrograms better than VQ-VAE.
% Also, standard deviations of SQ-VAEs are smaller than those of VQ-VAEs.
% This demonstrates that the performance of SQ-VAE is more stable across different random seeds. 
%Among the different parameterizations, SQ-VAE (\Rnum{4}) performs the best, which agrees with the result on image reconstruction. However, we aware that the performance of each parameterization depends on dataset properties and model architectures.
%Exploring the best parameterization for different dataset and model architecture is a possible direction for the future work.
%Regardless, these experimental results show that SQ-VAE performs much better than VQ-VAE also on speech data.

% Although SQ-VAE yields improvement on log-Mel spectrogram reconstruction, however, when we applied SQ-VAE to a downstream task called acoustic unit discovery, we cannot observe significant improvement in this case.
Although SQ-VAE improves log-Mel spectrogram reconstruction, we observed no significant improvement when we applied SQ-VAE to a downstream task called acoustic unit discovery.
Further exploration would be conducted to investigate the cause.
The details about the experiment and the samples of reconstructed log-Mel spectrograms are described in Appendix~\ref{sec:app_sub_details_gaussian_speech}.
%The details are described in Appendix~\ref{sec:app_sub_details_gaussian_speech}.

%- subsubsection
\subsection{Categorical Distributions}
\label{sec:sub_experiments_vmf_sqvae}

\begin{table}
\centering
    \caption{Evaluation on CelebA-Mask. The pixel error (\%), mIoU, and perplexity are evaluated using the test set.} 
    \vskip 0.1in
    \renewcommand{\arraystretch}{1.1}
    % \resizebox{\columnwidth}{!}{
    \small
    \begin{tabular}{l|c|c|c}
        \bhline{0.8pt}
        Model & Pixel error & mIoU & Perplexity \\
        \hline
        VAE & 8.79 $\pm$ 0.01 & 55.8 $\pm$ 0.3 & -- \\
        VQ-VAE w/ EMA & 6.95 $\pm$ 0.14 & 59.7 $\pm$ 0.7 & 46.2 $\pm$ 2.0  \\
        % Na\"ive SQ as in \beqref{eq:decoder_naive_categorical_sqvae}
        NC SQ-VAE
        & 6.63 $\pm$ 1.38 & 64.1 $\pm$ 5.4 & 12.6 $\pm$ 5.2 \\
        \hline
        % (vMF decoder) & & & \\
        % VAE & TBU & TBU & -- \\
        vMF SQ-VAE & \textbf{3.51} $\pm$ 0.17 & \textbf{74.6} $\pm$ 0.0 & \textbf{52.4} $\pm$ 0.8 \\
        % vMF SQ-VAE & $\bm{3.53 \pm x.xx}$ & $\bm{74.6 \pm x.x}$ & $\bm{52.7 \pm x.x}$ \\
        \bhline{0.8pt}
    \end{tabular}
    % }
    \label{tb:vision_categorical}
    \vskip -0.2in
\end{table}
\begin{table}
\centering
    \caption{Evaluation on MNIST and gray-scaled CelebA. The MSE ($\times{}10^3$) is evaluated using the test set.} 
    \vskip 0.1in
    \renewcommand{\arraystretch}{1.1}
    % \resizebox{\columnwidth}{!}{
    \small
    \begin{tabular}{l|c|c}
        \bhline{0.8pt}
        Model & MNIST & Gray-CelebA
        \\
        \hline
        VAE & 22.80 $\pm$ 0.32 & 17.73 $\pm$ 0.23 \\
        VQ-VAE w/ EMA & 6.24 $\pm$ 0.18 & 5.19 $\pm$ 0.06 \\
        NC SQ-VAE & 10.89 $\pm$ 0.47 & 3.88 $\pm$ 0.02 \\
        \hline
        % (vMF decoder) & & \\
        % VAE & TBU & TBU \\
        vMF SQ-VAE & \textbf{1.63} $\pm$ 0.21 & \textbf{2.37} $\pm$ 0.01 \\
        \bhline{0.8pt}
    \end{tabular}
    % }
    \label{tb:vision_discrete_ordinal}
    \vskip -0.2in
\end{table}
% \paragraph{Vision}
We test vMF SQ-VAE using CelebAHQ-Mask, which is a categorical image dataset with $L=19$ categories.
In this experiment, the segmentation maps are rescaled to 64$\times$64 with the nearest neighbor interpolation.
We compare vMF SQ-VAE with VAE, VQ-VAE, and NC SQ-VAE ~\beqref{eq:elbo_naive_categorical_sqvae}.
For VAE and VQ-VAE, softmax is applied to the output of the decoders, and CE loss is adopted as the reconstruction objective. 
The reconstruction quality is measured using the pixel error and the mean
of the class-wise intersection over union (mIoU)~\citep{cordts2016cityscapes}.

The result is shown in Table~\ref{tb:vision_categorical}.
As expected, the codebook perplexity in NC SQ-VAE significantly deteriorates, whereas it still achieves better pixel error and mIoU than VQ-VAE.
On the other hand, vMF SQ-VAE outperforms all the baselines with the highest codebook perplexity.

Next, we apply these methods to gray-scaled (256 categories) image datasets: MNIST and gray-scaled CelebA. The reconstruction accuracy is shown in Table~\ref{tb:vision_discrete_ordinal}.
Again, vMF SQ-VAE outperforms the baselines, whereas NC SQ-VAE performs worse than VQ-VAE on MNIST.
%vMF SQ-VAE outperforms the baselines while the MSE of NC SQ-VAE is larger than that of VQ-VAE on MNIST.
The result shows that vMF SQ-VAE is applicable even when the dataset has a large number of categories.

% \textbf{Semantic segmentation}
% We solved semantic segmentation task with autoencoder equipped with codebook on CelebAHQ-Mask and Cityscapes.

%%----- Section
\section{Conclusion}
\label{sec:conclusion}
We proposed SQ-VAE, a VAE-based framework with discretized latent space and combined with a pair of stochastic dequantization and quantization processes that includes a trainable categorical posterior.
This framework bridges the training schemes of VAE and VQ-VAE, and the trainable categorical posterior yields the self-annealing effect, which improves the reconstruction quality. 
Both Gaussian SQ-VAE and vMF SQ-VAE outperform the conventional approaches through experiments using vision and speech tasks. 
Moreover, SQ-VAE requires only a predefined scheduling of the temperature parameter of Gumbel--softmax. 
This hints a potential application of SQ-VAE on data compression.
%Moreover, SQ-VAE outperforms the VQ-VAE equipped with all the common heuristics and requires only a pre-defined scheduling of the temperature parameter of Gumbel--softmax.
%We propose SQ-VAEs, VAEs equipped with a trainable codebook.
%SQ-VAE was derived completely in variational inference with stochastic dequantization and quantization processes.
%The variational quantization scheme enables automatic annealing of quantization manner. In addition, SQ-VAEs reduce ad-hoc techniques and hyperparameters.
%We concretize the model into two variants; Gaussian SQ-VAE and von Mises-Fisher (vMF) SQ-VAE for continuous and categorical distributions, respectively.
%Empirical results on vision and speech dataset show that SQ-VAEs outperform VQ-VAE with the adaptive quantization scheme.
%Among the different parameterizations, SQ-VAE (\Rnum{4}) performs the best in most of the situations. 

% However, we are aware that the performance of each parameterization in the quantization process depends on dataset properties and model architectures.
Exploring the best parameterization of the trainable categorical posterior for different datasets and model architectures would be a direction of our future work.

% \begin{comment}
%%----- Section
\section*{Acknowledgements}
We would like to thank Masato Ishii, Lukas Mauch and Bac NguyenCong for many helpful comments during the preparation of this manuscript.
Besides, we thank anonymous reviewers for their valuable suggestions and comments.
% \end{comment}


% \bibliography{example_paper}
\bibliography{str_def_abrv,refs_dgm,refs_ml,refs_added}
\bibliographystyle{icml2022}

\appendix
% \twocolumn[
% \icmltitle{Supplementary Material for SQ-VAE}
% ]
\onecolumn
% \doublespacing
\icmltitle{Supplementary Material}% for SQ-VAE}

%%----- Section
\section{Notations and Definitions}
\label{sec:app_notations}
Here, we introduce several notations and definitions we used in the article.
Notations used in formulation of SQ-VAE is listed in Table~\ref{tb:notations}.

% \paragraph{Softmax function}
\textbf{Softmax function}~
We use the following notation to represent the $k$-th component of a softmax function that operates among the elements indexed by $k$. It can also be regarded as a function that maps $\mathbb{R}^K$ to $(0,1)^K$.
%We use the following notation to represent the $k$-th component of its output vector:
\begin{align}
    \softmax_k(\{z_j\}_{j=1}^K)&=\frac{\exp(z_k)}{\sum_{j=1}^{K}\exp(z_j)}.
\end{align}

% \paragraph{von Mises-Fisher distribution}
\textbf{von Mises-Fisher distribution}~
The von Mises-Fischer distribution is parameterized with  mean direction and concentration $(\mathbf{m},\kappa)$, which is represented as
\begin{align}
    \mathrm{vMF}(\mathbf{m},\kappa)=C_F(\kappa)\exp(\kappa\mathbf{m}^\top\bv),
\end{align}
where $C_F(\kappa)$ is the normalizing constant obtained by using
\begin{align}
    C_F(\kappa)=\int_{\bv\in\mathcal{S}^{F-1}}\exp(\kappa\mathbf{m}^\top\bv)d\bv.
\end{align}
The normalizing constant can also be expressed with the modified Bessel function of the first kind $I_\nu(\cdot)$\footnote{In our experiments, we use the following implementation provided by \citet{davidson2018hyperspherical} to evaluate the modified Bessel function:
\url{https://github.com/nicola-decao/s-vae-pytorch/tree/master/hyperspherical_vae}}
% \footnote{This is a built-in function in TensorFlow. For Pytorch, the following repository can be used~\citep{davidson2018hyperspherical}. \url{https://github.com/nicola-decao/s-vae-pytorch}}
as
\begin{align}
    \log{}C_F(\kappa)=
    \left(\frac{F}{2}-1\right)\log\kappa
    - \log{}I_{F/2-1}(\kappa)
    - \frac{F}{2}\log(2\pi).
    \label{eq:normalizing_constant_vmf}
\end{align}

%%----- Section
\section{Derivation Details}
\label{sec:app_derivation_elbos}
The ELBO of SQ-VAE can be calculated using the Bayes' theorem $P_\btheta(\bZq|\bx)p_\btheta(\bx)=p_\btheta(\bx|\bZq)P(\bZq)$ as
\begin{align}
    \log{p_\btheta(\bx)}
    &\geq\log{p_\btheta(\bx)}-\KL(q_\bomega(\bZ|\bx)\hat{P}_\bvarphi(\bZq|\bZ)\parallel{}P_\btheta(\bZq|\bx)p_\bvarphi(\bZ|\bZq))\nonumber\\
    &=\E_{q_\bomega(\bZ|\bx)\hat{P}_\bvarphi(\bZq|\bZ)}\left[\log\frac{p_\btheta(\bx)P_\btheta(\bZq|\bx)p_\bvarphi(\bZ|\bZq)}{q_\bomega(\bZ|\bx)\hat{P}_\bvarphi(\bZq|\bZ)}\right]\nonumber\\
    % &=\E_{q_\bomega(\bZ|\bx)\hat{P}_\bvarphi(\bZq|\bZ)}\left[\log{p_\btheta(\bx|\bZq)}-\log\frac{q_\bomega(\bZ|\bx)}{p_\bvarphi(\bZ|\bZq)}-\log\frac{\hat{P}_\bvarphi(\bZq|\bZ)}{P(\bZq)}\right]\nonumber\\
    % &=\E_{Q_{\bomega}(\bZq|\bx)}\left[\log{}p_{\btheta}(\bx|\bZq)
    % -\KL(q_\bomega(\bZ|\bx)\parallel p_\bvarphi(\bZ|\bZq))\right]
    % -\E_{q_\bomega(\bZ|\bx)}H(\hat{P}_\bvarphi(\bZq|\bZ))
    % -d_z\log{K}.
    &=\E_{q_\bomega(\bZ|\bx)\hat{P}_\bvarphi(\bZq|\bZ)}\left[\log\frac{p_\btheta(\bx|\bZq)p_\bvarphi(\bZ|\bZq)}{q_\bomega(\bZ|\bx)}-\log\frac{\hat{P}_\bvarphi(\bZq|\bZ)}{P(\bZq)}\right]\nonumber\\
    &=\E_{q_\bomega(\bZ|\bx)\hat{P}_\bvarphi(\bZq|\bZ)}\left[\log\frac{p_\btheta(\bx|\bZq)p_\bvarphi(\bZ|\bZq)}{q_\bomega(\bZ|\bx)}\right]
    +\E_{q_\bomega(\bZ|\bx)}H(\hat{P}_\bvarphi(\bZq|\bZ))
    -d_z\log{K}.
    \label{eq:elbo_general_sqvae_derive}
    % =:-\Ls_\text{SQ}(\bx)
\end{align}

%-- subsection
\subsection{Gaussian SQ-VAE}
\label{sec:app_sub_elbo_gaussian_sqvae}
By applying the Gaussian assumption as mentioned in Section~\ref{sec:sub_gaussian_sqvae} into~\beqref{eq:elbo_general_sqvae_derive}, we obtain the first two terms as
\begin{align}
    \log{p_\btheta(\bx|\bZq)}
    &=\log\mathcal{N}(f_\btheta(\bZq),\sigma^2\bI)\nonumber\\
    &=-\frac{1}{2\sigma^2}\|\bx-f_\btheta(\bZq)\|_2^2+\frac{D}{2}\log(2\pi\sigma^2),\\
    % \KL(q_\bomega(\bZ|\bx)\parallel{}p_\bvarphi(\bZ|\bZq))
    % &=\KL\left(\prod_{i=1}^{d_z}p_\bvarphi(\bz_i|g_{\bphi,i}(\bx),\bm{\Sigma}_\bvarphi)\parallel{}\prod_{i=1}^{d_z}p_\bvarphi(\bz_i|\bz_{\mathrm{q},i},\bm{\Sigma}_\bvarphi)\right)\nonumber\\
    % &=\KL\left(\prod_{i=1}^{d_z}\mathcal{N}(g_{\bphi,i}(\bx),\bm{\Sigma}_\bvarphi)\parallel{}\prod_{i=1}^{d_z}\mathcal{N}(\bz_{\mathrm{q},i},\bm{\Sigma}_\bvarphi)\right)\nonumber\\
    % &=\sum_{i=1}^{d_z}\KL(\mathcal{N}(g_{\bphi,i}(\bx),\bm{\Sigma}_\bvarphi)\parallel{}\mathcal{N}(\bz_{\mathrm{q},i},\bm{\Sigma}_\bvarphi))\nonumber\\
    % &=\frac{1}{2}\sum_{i=1}^{d_z}\left(\log\frac{|\bm{\Sigma}_\bvarphi|}{|\bm{\Sigma}_\bvarphi|}-d_b+\trace(\bm{\Sigma}_\bvarphi^{-1}\bm{\Sigma}_\bvarphi)+(g_{\bphi,i}(\bx)-\bz_{\mathrm{q},i})^\top\bm{\Sigma}_\bvarphi^{-1}(g_{\bphi,i}(\bx)-\bz_{\mathrm{q},i})\right)\nonumber\\
    % &=\frac{1}{2}\sum_{i=1}^{d_z}(g_{\bphi,i}(\bx)-\bz_{\mathrm{q},i})^\top\bm{\Sigma}_\bvarphi^{-1}(g_{\bphi,i}(\bx)-\bz_{\mathrm{q},i}).
    \E_{q_\bomega(\bZ|\bx)\hat{P}_\bvarphi(\bZq|\bZ)}\left[\log\frac{p_\bvarphi(\bZ|\bZq)}{q_\bomega(\bZ|\bx)}\right]
    &=\E_{q_\bomega(\bZ|\bx)\hat{P}_\bvarphi(\bZq|\bZ)}\left[\log\frac{\prod_{i=1}^{d_z}p_\bvarphi(\bz_i|\bZq)}{\prod_{i=1}^{d_z}q_\bomega(\bz_i|\bx)}\right]\nonumber\\
    &=\sum_{i=1}^{d_z}\E_{q_\bomega(\bZ|\bx)\hat{P}_\bvarphi(\bZq|\bZ)}\left[\log\frac{p_\bvarphi(\bz_i|\bZq)}{p_\bvarphi(\bz_i|g_\bphi(\bx))}\right]\\
    &=\frac{1}{2}\sum_{i=1}^{d_z}\left(-\E_{q_\bomega(\bZ|\bx)\hat{P}_\bvarphi(\bZq|\bZ)}\left[(\bz_{i}-\bz_{\mathrm{q},i})^\top\bm{\Sigma}_\bvarphi^{-1}(\bz_{i}-\bz_{\mathrm{q},i})\right]\right.\nonumber\\
    &\qquad\left.\vphantom{-\E_{q_\bomega(\bZ|\bx)\hat{P}_\bvarphi(\bZq|\bZ)}}+\E_{p_\bvarphi(\bZ|g_\bphi(\bx))}\left[(\bz_{i}-g_{\bphi,i}(\bx))^\top\bm{\Sigma}_\bvarphi^{-1}(\bz_{i}-g_{\bphi,i}(\bx))\right]\right)
    \nonumber\\
    &=-\E_{q_\bomega(\bZ|\bx)\hat{P}_\bvarphi(\bZq|\bZ)}\sum_{i=1}^{d_z}\left[\frac{1}{2}(\bz_{i}-\bz_{\mathrm{q},i})^\top\bm{\Sigma}_\bvarphi^{-1}(\bz_{i}-\bz_{\mathrm{q},i})\right]+\frac{d_bd_z}{2}.
\end{align}
This leads to the objective function of Gaussian SQ-VAE $\Ls_{\mathcal{N}\text{-SQ}}$ as \beqref{eq:elbo_gaussian_sqvae}. 

%-- subsection
\subsection{vMF SQ-VAE}
\label{sec:app_sub_elbo_vmf_sqvae}
The ELBO for vMF SQ-VAE is obtained by replacing $\bx$ with $\bV$ as
\begin{align}
    \log{p_\btheta(\bV)}
    % &\geq\E_{Q_{\bomega}(\bZq|\bV)}\left[\log{}p_{\btheta}(\bV|\bZq)
    % -\KL(q_\bomega(\bZ|\bV)\parallel p_\bvarphi(\bZ|\bZq))\right]
    % -\E_{q_\bomega(\bZ|\bV)}H(\hat{P}_\bvarphi(\bZq|\bZ))
    % -d_z\log{K}.
    &\geq\E_{q_\bomega(\bZ|\bV)\hat{P}_\bvarphi(\bZq|\bZ)}\left[\log\frac{p_\btheta(\bV|\bZq)p_\bvarphi(\bZ|\bZq)}{q_\bomega(\bZ|\bV)}\right]
    -\E_{q_\bomega(\bZ|\bV)}H(\hat{P}_\bvarphi(\bZq|\bZ))
    -d_z\log{K}.
    \label{eq:elbo_vmf_sqvae_derive}
    % =:-\Ls_\text{SQ}(\bx)
\end{align}
% The von Mises-Fischer distribution parameterized with its mean direction and concentration $(\mathbf{m},\kappa)$ is represented as
% \begin{align}
%     \mathrm{vMF}(\bv|\mathbf{m},\kappa)=C_F(\kappa)\exp(\kappa\mathbf{m}^\top\bv).
% \end{align}
By applying the von Mises-Fischer distribution as mentioned in \ref{sec:sub_vmf_sqvae} into \beqref{eq:elbo_general_sqvae_derive}, we obtain the first two terms as
\begin{align}
    \log{p_\btheta(\bx|\bZq)}
    &=\sum_{d=1}^D\log\mathrm{vMF}(\tilde{f}_{\btheta,d}(\bZq),\kappa)\nonumber\\
    &=\kappa\sum_{d=1}^D\bv_d^\top\tilde{f}_{\btheta,d}(\bZq)+\log{C_F(\kappa)},\\
    % \KL(q_\bomega(\bZ|\bx)\parallel{}p_\bvarphi(\bZ|\bZq))
    % &=\KL\left(\prod_{i=1}^{d_z}p_\bvarphi(\bz_i|g_{\bphi,i}(\bx),\bm{\Sigma}_\bvarphi)\parallel{}\prod_{i=1}^{d_z}p_\bvarphi(\bz_i|\bz_{\mathrm{q},i},\bm{\Sigma}_\bvarphi)\right)\nonumber\\
    % &=\KL\left(\prod_{i=1}^{d_z}\mathrm{vMF}(g_{\bphi,i}(\bx),\kappa_\bvarphi)\parallel{}\prod_{i=1}^{d_z}\mathrm{vMF}(\bz_{\mathrm{q},i},\kappa_\bvarphi)\right)\nonumber\\
    % &=\sum_{i=1}^{d_z}\KL\left(\mathrm{vMF}(g_{\bphi,i}(\bx),\kappa_\bvarphi)\parallel{}\mathrm{vMF}(\bz_{\mathrm{q},i},\kappa_\bvarphi)\right)\nonumber\\
    % &=\sum_{i=1}^{d_z}\E_{\mathrm{vMF}(g_{\bphi,i}(\bx),\kappa_\bvarphi)}\left[\kappa_\bvarphi(g_{\bphi,i}(\bx)-\bz_{\mathrm{q},i})^\top\bz_i+\log\frac{C_F(\kappa_\bvarphi)}{C_F(\kappa_\bvarphi)}\right]\nonumber\\
    % &=\sum_{i=1}^{d_z}\kappa_\bvarphi(g_{\bphi,i}(\bx)-\bz_{\mathrm{q},i})^\top g_{\bphi,i}(\bx)\nonumber\\
    % &=\sum_{i=1}^{d_z}\kappa_{\bvarphi,i}(1-\bz_{\mathrm{q},i}^\top g_{\bphi,i}(\bx))
    \E_{q_\bomega(\bZ|\bx)\hat{P}_\bvarphi(\bZq|\bZ)}\left[\log\frac{p_\bvarphi(\bZ|\bZq)}{q_\bomega(\bZ|\bx)}\right]
    &=\E_{q_\bomega(\bZ|\bx)\hat{P}_\bvarphi(\bZq|\bZ)}\left[\log\frac{\prod_{i=1}^{d_z}p_\bvarphi(\bz_i|\bZq)}{\prod_{i=1}^{d_z}q_\bomega(\bz_i|\bx)}\right]\nonumber\\
    &=\sum_{i=1}^{d_z}\E_{q_\bomega(\bZ|\bx)\hat{P}_\bvarphi(\bZq|\bZ)}\left[\log\frac{p_\bvarphi(\bz_i|\bZq)}{p_\bvarphi(\bz_i|g_{\bphi,i}(\bx))}\right]\nonumber\\
    &=\sum_{i=1}^{d_z}\left(\E_{q_\bomega(\bZ|\bx)\hat{P}_\bvarphi(\bZq|\bZ)}\left[\kappa_\bvarphi\bz_{\mathrm{q},i}^\top\bz_{i}\right]-\E_{p_\bvarphi(\bZ|g_\bphi(\bx))}\left[\kappa_\bvarphi{}g_{\bphi,i}(\bx)^\top\bz_{i}\right]\right)
    \nonumber\\
    &=\sum_{i=1}^{d_z}\left(\E_{q_\bomega(\bZ|\bx)\hat{P}_\bvarphi(\bZq|\bZ)}\left[\kappa_\bvarphi\bz_{\mathrm{q},i}^\top\bz_{i}\right]-\kappa_\bvarphi\|g_{\bphi,i}(\bx))\|_2^2\right)\nonumber\\
    &=\E_{q_\bomega(\bZ|\bx)\hat{P}_\bvarphi(\bZq|\bZ)}\left[\sum_{i=1}^{d_z}\kappa_\bvarphi(\bz_{\mathrm{q},i}^\top\bz_{i}-1)\right].
\end{align}

This leads to the objective function of vMF SQ-VAE $\Ls_{\text{vMF-SQ}}$ as \beqref{eq:elbo_vmf_sqvae}. 

\begin{table*}[t]
\centering
    \caption{Notations used in our formulation of SQ-VAE.}
    \vskip 0.1in
    \renewcommand{\arraystretch}{1.25}
    \small
    \begin{tabular}{c|l}
        \bhline{0.8pt}
        \multicolumn{2}{l}{\textbf{Trainable parameters}} \\
        \hline
        $\mathbf{B}$ & A trainable codebook that consists of $K$ codebook elements, $\{\mathbf{b}_k\}_{k=1}^K$.\\
        $\btheta$ & A trainable parameter for the decoder. The decoding function is denoted as $f_\btheta: \realnum^{d_b\times{}d_z}\to\realnum^D$. \\
        $\bphi$ & A trainable parameter for the encoder. The encoding function is denoted as $g_\bphi: \realnum^D\to\realnum^{d_b\times{}d_z}$. \\ 
        $\bvarphi$ & A trainable parameter for probabilistic dequantization and quantization processes in the latent space. \\
        $\bomega$ & A tuple of $\bphi$ and $\bvarphi$, which is used in the probabilistic encoding process. \\
        \hline
        \multicolumn{2}{l}{\textbf{Latent variables}} \\
        \hline
        $\mathbf{Z}_\mathrm{q}$ & A latent variable that consists of $d_z$ codebook elements, i.e., $\mathbf{Z}_\mathrm{q}\in\mathbf{B}^{d_z}$, whose $i$th vector is denoted as $\mathbf{z}_i\in\mathbf{B}$. \\
        $\hat{\mathbf{Z}}_\mathrm{q}$ & A latent variable encoded directly from the encoder, i.e., $\hat{\mathbf{Z}}_\mathrm{q}=g_{\bphi}(\bx)$, which approximates $\mathbf{Z}_\mathrm{q}$. \\
        $\mathbf{Z}$ & A continuous latent variable obtained by dequantization of $\mathbf{Z}_\mathrm{q}$ or $\hat{\mathbf{Z}}_\mathrm{q}$. \\
        \hline
        \multicolumn{2}{l}{\textbf{Probabilistic processes}} \\
        \hline
        $p_\btheta(\mathbf{x}|\mathbf{Z}_{\mathrm{q}})$ & A decoder distribution whose mean is given by $f_\btheta(\mathbf{Z}_{\mathrm{q}})$ (e.g., Gaussian and vMF distribution).\\
        $p_\bvarphi(\mathbf{Z}|\mathbf{Z}_\mathrm{q})$ & A probability distribution for the dequantization of $\mathbf{Z}_\mathrm{q}$ (e.g., Gaussian and vMF distribution).\\
        $q_\bomega(\mathbf{Z}|\mathbf{x})$ &
        A distribution for the encoding from $\mathbf{x}$ to $\mathbf{Z}$, which consists of deterministic encoding and stochastic dequantization. \\
        $\hat{P}_\bvarphi(\mathbf{Z}_\mathrm{q}|\mathbf{Z})$ & A categorical distribution for the quantization of $\mathbf{Z}$, which is inverse process of $p_\bvarphi(\mathbf{Z}|\mathbf{Z}_\mathrm{q})$. \\
        \bhline{0.8pt}
    \end{tabular}
    \label{tb:notations}
    \vskip -0.1in
\end{table*}

%%----- Section
\section{Training Procedures of SQ-VAEs}
\label{sec:app_psuedo_code}

The training procedures of Gaussian SQ-VAE and vMF SQ-VAE are described here in Algorithms~\ref{alg:gaussian} and \ref{alg:vmf}, respectively. $[t]$ indicates the index of training steps.

% \begin{center}
% \begin{minipage}{.45\textwidth}
% \begin{algorithm}[H]
%   \caption{General SQ-VAE}
%   \label{alg:general}
% \begin{algorithmic}
%   \STATE {\bfseries Input:} Dataset $(\bx^{(n)})_{n=1}^{N_\mathrm{data}}$
%   \STATE Initialize parameters $\btheta^{[0]}$, $\bomega^{[0]}$
%   \STATE Initialize a codebook $\{\bb_k^{[0]}\}_{k=1}^K$
%   \FOR{$\tau=1,2,\ldots,T$}
%   \STATE Get $\bx=\mathrm{GetData}()$
%   \STATE Get $\hat{\bZ}_\mathrm{q}=g_{\bphi^{(\tau-1)}}(\bx)$
%   \STATE Sample $\bZq\sim\hat{P}_{\bvarphi^{(\tau-1)}}(\bZq|\hat{\bZ}_\mathrm{q})$
%   \STATE Calculate $\Ls_\text{SQ}^{(\tau-1)}(\bx)$ in \beqref{eq:elbo_sqvae} with $\bx$ and $\bZq$
%   \STATE $\btheta^{(\tau)},\bomega^{(\tau)}, (\bb_k^{(\tau)})_{k=1}^K =\mathrm{Update}(\Ls_\text{SQ}^{(\tau-1)})$
%   \ENDFOR
% \end{algorithmic}
% \end{algorithm}
% \end{minipage}
% \end{center}

\newcommand{\squeezedmathenv}[1]{
\begin{flalign}
{#1\nonumber}
\end{flalign}
}

\begin{center}
\begin{minipage}{.45\textwidth}
\begin{algorithm}[H]
   \caption{Gaussian SQ-VAE}
   \label{alg:gaussian}
\begin{algorithmic}
   \STATE {\bfseries Input:} Dataset $\bx_{data}$% = \{\bx^{(n)}\}_{n=1}^{N_\mathrm{data}}$
   \STATE Initialize the codebook and parameters: 
   \STATE $\bB^{[0]}$, $\btheta^{[0]}$ and $\bomega^{[0]}:=\{\bphi^{[0]}, \bvarphi^{[0]}\}$ %$\{\bb_k^{[0]}\}_{k=1}^K$
   \STATE 
   \FOR{$t=1,2,\ldots,T$}
   \STATE $\bx \leftarrow$ Random minibatch from $\bx_{data}$
   \STATE  
   \STATE $\hat{\bZ}_\mathrm{q} \leftarrow g_{\bphi^{[t-1]}}(\bx)$
   \STATE $\bZq\sim\hat{P}_{\bvarphi^{[t-1]}}(\bZq|\hat{\bZ}_\mathrm{q})$
%   \STATE $\textbf{g} \leftarrow  \nabla_{\btheta,\bomega,\bB}\Ls_{\mathcal{N}\text{-SQ}}^{[t-1]}(\btheta^{[t-1]}, \bomega^{[t-1]}, \bB^{[t-1]}; \bx, \bZq)$
   \STATE $\textbf{g} \leftarrow  \nabla_{\btheta,\bomega,\bB}\Ls_{\mathcal{N}\text{-SQ}}(\btheta^{[t-1]}, \bomega^{[t-1]}, \bB^{[t-1]})$
   \STATE \hspace{80pt}with sampled $\bx$ and $\bZq$
   \STATE $\btheta^{[t]},\bomega^{[t]}, \bB^{[t]} \leftarrow$ Update parameters using $\textbf{g}$
   \ENDFOR
\end{algorithmic}
\end{algorithm}
\end{minipage}
\hspace{5pt}%
\begin{minipage}{.53\textwidth}
\begin{algorithm}[H]
   \caption{vMF SQ-VAE}
   \label{alg:vmf}
\begin{algorithmic}
   \STATE {\bfseries Input:} Dataset $\bx_{data}$ 
% = \{\bx^{(n)}\}_{n=1}^{N_\mathrm{data}}$
   \STATE Initialize the codebook and parameters: 
   \STATE $\bB^{[0]}$, $\btheta^{[0]}$ and $\bomega^{[0]}:=\{\bphi^{[0]}, \bvarphi^{[0]}\}$
   \STATE Determine a vector set  $\bW:=\{\bw_l|\bw_l\in\mathcal{S}^{F-1}\}_{l=1}^L$
   \FOR{$t=1,2,\ldots,T$}
%   \IF{$x_i > x_{i+1}$}
   \STATE $\bx \leftarrow$ Random minibatch from $\bx_{data}$
   \STATE $\bV \leftarrow$ Project $\bx$ onto $\mathcal{S}^{F-1}$
   \STATE $\hat{\bZ}_\mathrm{q} \leftarrow g_{\bphi^{[t-1]}}(\bV)$
   \STATE $\bZq\sim\hat{P}_{\bvarphi^{[t-1]}}(\bZq|\hat{\bZ}_\mathrm{q})$
%   \STATE $\textbf{g} \leftarrow \nabla_{\btheta,\bomega,\bB}\Ls_{\text{vMF-SQ}}^{[t-1]}(\btheta^{[t-1]}, \bomega^{[t-1]}, \bB^{[t-1]}; \bV, \bZq, \bW)$
   \STATE $\textbf{g} \leftarrow  \nabla_{\btheta,\bomega,\bB}\Ls_{\text{vMF-SQ}}(\btheta^{[t-1]}, \bomega^{[t-1]}, \bB^{[t-1]})$
   \STATE \hspace{100pt}with sampled $\bx$ and $\bZq$
   \STATE $\btheta^{[t]},\bomega^{[t]}, \bB^{[t]} \leftarrow$ Update parameters using $\textbf{g}$
   \ENDFOR
%   \UNTIL{$noChange$ is $true$}
\end{algorithmic}
\end{algorithm}
\end{minipage}
\end{center}

% %%----- Section
% \section{vMF Decoder for Categorical Distributions}
% \label{sec:app_vmf_decoder}

% %-- subsection
% \subsection{Comparison with Usual Categorical Decoder}
% \label{sec:app_categorical_vs_vmf_decoder}

% %-- subsection
% \subsection{Application of vMF Decoder to VAE}
% \label{sec:app_apply_vmf_decoder_vae}

%%----- Section
% \section{Propositions of Section~\ref{sec:sub_behavior_quantization}}
\section{Self-Annealed Quantization}
\label{sec:app_details_adaptive_quantization}

%-- subsection
\subsection{Similarity between SQ-VAE and conventional VAE}
\label{sec:app_adaptive_quantization_vmf_sqvae}

A property that is similar to Proposition~\ref{th:to_deterministic_quantization_gaussian} can be observed in Gaussian VAE with the posterior $q_\bphi(\bz|\bx)=\mathcal{N}(g_\bphi(\bx),s^2\bI)$.
In this case, the latent variables are perturbed by adding Gaussian noises with their variance $s^2\bI$ to the encoded points. When $s^2$ is trained, it approaches zero~\citep{takida2021preventing} as the training progresses, which means that the stochastic encoding becomes almost deterministic, i.e., no perturbation.

\subsection{Proof of Proposition~\ref{th:to_deterministic_quantization_gaussian}}
\label{sec:app_proof_proposition}
\setcounter{proposition}{0}
\begin{proposition}
    Assume that $\pdata(\bx)$ has finite support, whereas $g_{\bphi}$ and $\{\bb_k\}_{k=1}^K$ are bounded.
    Let $\bomega^{*}=\{\bphi^*,\bvarphi^*\}$ be a minimizer of $\E_{\pdata(\bx)}\KL(Q_{\bomega}(\bZq|\bx)\parallel{}P_\btheta(\bZq|\bx))$ with fixed $\btheta$, $\sigma^2$, and $\{\bb_k\}_{k=1}^K$.
    %If $\sigma^2\to\infty$, then $\sigma_{\bvarphi^*}^2\to\infty$.
    If $\sigma^2\to0$, then $\sigma_{\bvarphi^*}^2\to0$.
%    Assuming that $\pdata(\bx)$ has finite support, $g_{\bphi}$ can be arbitrarily complex but bounded, and $\{\bb_k\}_{k=1}^K$ is bounded.
%    Suppose $\bomega^{*}=\{\bphi^*,\bvarphi^*\}$ to be a minimizer of $\E_{\pdata(\bx)}\KL(Q_{\bomega}(\bZq|\bx)\parallel{}P_\btheta(\bZq|\bx))$ with $\btheta$, $\sigma^2$ and $\{\bb_k\}_{k=1}^K$ fixed.
%    If $\sigma^2\to0$, then $\sigma_{\bvarphi^*}^2\to0$.
\end{proposition}
\setcounter{proposition}{2}
\begin{proof}
    We denote the number of samples included in the support of $\pdata(\bx)$ as $N_\mathrm{data}$.
    Let $\bZq^{\bk}\in\mathbf{B}^{d_z}$ with a set $\bk\in[K]^{d_z}$ be a discrete tensor with its $i$th elements $\bb_{k_i}$, where $k_i$ indicates the $i$th coordinate of $\bk$ for $i\in[d_z]$.
    Among $f_{\btheta}(\bZq^{\bk})$ with all the possible $\bk\in[K]^{d_z}$, the set of indices corresponding to the nearest neighbor of $\bx^{(n)}$ is denoted as $\hat{\bk}^{(n)}$ for $n\in[N_\mathrm{data}]$, i.e., $\hat{\bk}^{(n)}=\argmin_{\bk\in[K]^{d_z}}\left\|\bx^{(n)}-f_{\btheta}(\bZq^{\bk})\right\|_2$.
    Now, $P_\btheta(\bZq|\bx^{(n)})$ can be evaluated by
    \begin{align}
        P_\btheta(\bZq|\bx^{(n)})
        &\propto P_\btheta(\bx^{(n)}|\bZq)P(\bZq)\notag\\
        % &\propto P_\btheta(\bx|\bZq)\notag\\
        &\propto \exp\left(-\frac{1}{2\sigma^2}\left\|\bx^{(n)}-f_\btheta(\bZq)\right\|_2^2\right).
    \end{align}
    Here, if $\sigma^2\to0$, then $P_\btheta(\bZq|\bx^{(n)})$ becomes the Kronecker delta function $\delta_{\bk,\hat{\bk}^{(n)}}$.
    
    Next, define $\mathcal{B}_k\subset\realnum^{d_b}$ to be a region such that $k=\argmin_{k^\prime}\|\bz-\bb_{k^\prime}\|_2$ for $\bz\in\mathcal{B}_k$.
    % From the assumption $K>N_{\mathrm{data}}^{1/d_z}~(\Leftrightarrow{}K^{d_z}>N$, all the samples $\bx^{(n)}~(n\in[N_{\mathrm{data}}])$ can have different sets of codebook indices each other.
    With the arbitrarily complex $g_\bphi$, $g_{\bphi,i}(\bx^{(n)})\in\mathcal{B}_{k}$ can be achieved for all $n\in[N_\mathrm{data}]$, $i\in[d_z]$ and $k\in[K]$, where $g_{\bphi,i}(\bx^{(n)})$ indicates the $i$th coordinates of $g_{\bphi}(\bx^{(n)})$.
    The divergence $\E_{\pdata(\bx)}\KL(Q_{\bomega}(\bZq|\bx)\parallel{}P_\btheta(\bZq|\bx))$ can be minimized to $0$ with
    \begin{subequations}
        \begin{align}
            &\sigma_{\bvarphi^*}^2\to0\qquad\text{and}
            \label{eq:cond_minimization_1}\\
            &g_{\bphi^*,i}(\bx^{(n)})\in\mathcal{B}_{\hat{k}^{(n)}_i}
            \qquad \text{for all } i\in[d_z]\text{ and }n\in[N_\mathrm{data}]
            \label{eq:cond_minimization_2}
        \end{align}
    \end{subequations}
    since \beqref{eq:cond_minimization_1} and \beqref{eq:cond_minimization_2} lead to $Q_{\bomega}(\bZq|\bx)=\delta_{\bk,\hat{\bk}^{(n)}}$.
    Here, $\hat{k}^{(n)}_i$ denotes the $i$th coordinates of $\hat{\bk}^{(n)}$.
    
    Finally, we prove that if $\sigma^2\to0$ and $\E_{\pdata(\bx)}\KL(Q_{\bomega^*}(\bZq|\bx)\parallel{}P_\btheta(\bZq|\bx))=0$, then $\sigma_{\bvarphi^*}^2\to0$.
    %\beqref{eq:cond_minimization_1} is the necessary condition for  .
    Define $p_1^{(n)}$ and $q_1^{(n)}$ to be $p_1=P_\btheta(\bZq^{\hat{\bk}^{(n)}}|\bx^{(n)})$ and $q_1=Q_\bomega(\bZq^{\hat{\bk}^{(n)}}|\bx^{(n)})$, respectively.
    Now, consider $\sigma_{\bvarphi^*}^2\not\to0$ as $\sigma^2\to0$.
    It immediately follows that $p_1\to{1}$ as $\sigma^2\to0$.
    Moreover, according to the assumption that $\pdata(\bx)$ has finite support and $g_\bphi$ is bounded, the distribution that $\bZ=g_\bphi(\bx)$ follows with $\bx\sim\pdata(\bx)$ has finite support as well, which leads to $q_1\neq1$.
    %%%%%%%%
    Here, the KL divergence for $\bx^{(n)}$ is bounded as
    \begin{align}
        \KL(Q_{\bomega^*}(\bZq|\bx^{(n)})\parallel{}P_\btheta(\bZq|\bx^{(n)}))
        &=\sum_{\bk\in[K]^{d_z}}Q_{\bomega^*}(\bZq^\bk|\bx^{(n)})\left[\log{}Q_{\bomega^*}(\bZq^\bk|\bx^{(n)})-\log{}P_\btheta(\bZq^\bk|\bx^{(n)})\right]\\
        &= -H[Q_{\bomega^*}(\bZq|\bx^{(n)})]-\sum_{\bk\in[K]^{d_z}}Q_{\bomega^*}(\bZq^\bk|\bx)\log{}P_\btheta(\bZq^\bk|\bx^{(n)})\notag\\
        &\geq -\log{K} - (1-q_1)\log(1-p_1)
        \label{eq:ineq_kl_divergence}
    \end{align}
    from
    \begin{align}
        H[Q_{\bomega^*}(\bZq|\bx^{(n)})]
        &\leq\log{K}\qquad\text{and}\\
        \sum_{\bk\in[K]^{d_z}}Q_{\bomega^*}(\bZq^\bk|\bx^{(n)})\log{}P_\btheta(\bZq^\bk|\bx)
        &= Q_{\bomega^*}(\bZq^{\hat{\bk}^{(n)}}|\bx^{(n)})\log{}P_\btheta(\bZq^{\hat{\bk}^{(n)}}|\bx^{(n)})
        +\sum_{\substack{\bk\in[K]^{d_z}\\\bk\neq\hat{\bk}^{(n)}}}Q_{\bomega^*}(\bZq^\bk|\bx^{(n)})\log{}P_\btheta(\bZq^\bk|\bx^{(n)})\\
        &\leq q_1\log{p_1} + (1-q_1)\log(1-p_1)\\
        &\leq (1-q_1)\log(1-p_1)
    \end{align}
    The first term in \beqref{eq:ineq_kl_divergence} is finite since $K$ is assumed to be finite.
    On the other hand, from $1-p_1\to0~(\sigma^2\to0)$ and $q_1\neq1$, it follows that the second term in \beqref{eq:ineq_kl_divergence} diverges to infinity as $\sigma^2\to0$. This leads to the infinite KL divergence and contradicts to the fact that $\sigma_{\bvarphi^*}$ is a minimizer of $\E_{\pdata(\bx)}\KL(Q_{\bomega}(\bZq|\bx)\parallel{}P_\btheta(\bZq|\bx))$. 
    Thus, we must have $\sigma_{\bvarphi^*}\to0$ as $\sigma^2\to0$.
\end{proof}
 
%Proposition~\ref{th:to_deterministic_quantization_vmf} can be proved with the similar procedure to that of Proposition~\ref{th:to_deterministic_quantization_gaussian} while noting that
%\begin{align}
%    P_\btheta(\bZq|\bx^{(n)})
%    &\propto P_\btheta(\bV^{(n)}|\bZq)P(\bZq)\notag\\
%    % &\propto P_\btheta(\bx|\bZq)\notag\\
%    &\propto \exp\left(\kappa\sum_{d=1}^D\bv_d^{(n)\top}f%_{\btheta,d}(\bZq)\right),
%\end{align}
%where $-1\leq\bv_d^{(n)\top}f_{\btheta,d}(\bZq)\leq{}1$ since the vectors live on $\mathcal{S}^{F-1}$.
%\begin{align}
%    P_\btheta(\bZq|\bx^{(n)})
%    &\propto P_\btheta(\bV^{(n)}|\bZq)P(\bZq)\notag\\
%    % &\propto P_\btheta(\bx|\bZq)\notag\\
%    &\propto \exp\left(\kappa\sum_{d=1}^D\bv_d^{(n)\top}f%_{\btheta,d}(\bZq)\right),
%\end{align}
%where $-1\leq\bv_d^{(n)\top}f_{\btheta,d}(\bZq)\leq{}1$ since the vectors live on $\mathcal{S}^{F-1}$.

%-- subsection
\subsection{Details of Section~\ref{sec:sub_behavior_quantization}}
\label{sec:app_details_observation}

We set the codebook capacity to $(d_b,K)=(64, 128)$ for all the models.
We implement the decoder and encoder using the following standard two-layer ConvNet architectures:
\begin{flalign*}
  \bx\in\mathbb{R}^{28\times28\times1}
  &\to\mathrm{Conv}_{32}^{(4\times4)}\to\mathrm{BatchNorm}\to\mathrm{ReLU}&\text{size: }(32,14,14)\\
  &\to\mathrm{Conv}_{64}^{(4\times4)}&\text{size: }(64,7,7)\\
  \bZq\in\mathbf{B}^{7\times7}\subset\mathbb{R}^{64\times7\times7}
  &\to\mathrm{ConvT}_{32}^{(4\times4)}\to\mathrm{BatchNorm}\to\mathrm{ReLU}&\text{size: }(32,14,14)\\
  &\to\mathrm{ConvT}_{1}^{(4\times4)}\to\mathrm{Sigmoid}&\text{size: }(1,28,28),
\end{flalign*}
where the notations for the architecture parts are listed in Table~\ref{tb:block_list}.

The Adam optimizer is used with the initial learning rate of $0.001$.
The temperature parameter in Gumbel--softmax is annealed with the same preset schedule as that in Appendix~\ref{sec:app_details_experiments}.

% %- subsubsection
% \subsubsection{Gaussian SQ-VAE}

% %- subsubsection
% \subsubsection{vMF SQ-VAE}

%-- subsection
\subsection{Case of vMF SQ-VAE}
\label{sec:app_proposition_vmf}
As for vMF SQ-VAE, we have the following proposition:
\setcounter{proposition}{1}
\begin{proposition}
    Assuming that $\pdata(\bx)$ has finite support, whereas $g_{\bphi}$ and $\{\bb_k\}_{k=1}^K$ are bounded.
    Let $\bomega^{*}=\{\bphi^*,\bvarphi^*\}$ be a minimizer of $\E_{\pdata(\bx)}\KL(Q_{\bomega}(\bZq|\bx)\parallel{}P_\btheta(\bZq|\bx))$ with fixed $\btheta$, $\kappa$, and $\{\bb_k\}_{k=1}^K$.
    If $\kappa\to\infty$, then $\kappa_{\bvarphi^*}\to\infty$.
\end{proposition}
This proposition is similar to Proposition~\ref{th:to_deterministic_quantization_gaussian}.
Thus, it can be proved similarly to the proof of Proposition~\ref{th:to_deterministic_quantization_gaussian} while noting the following:
\begin{align}
    P_\btheta(\bZq|\bx^{(n)})
    &\propto P_\btheta(\bV^{(n)}|\bZq)P(\bZq)\notag\\
    % &\propto P_\btheta(\bx|\bZq)\notag\\
    &\propto \exp\left(\kappa\sum_{d=1}^D\bv_d^{(n)\top}\tilde{f}_{\btheta,d}(\bZq)\right),
\end{align}
where $-1\leq\bv_d^{(n)\top}\tilde{f}_{\btheta,d}(\bZq)\leq{}1$ since the vectors live on $\mathcal{S}^{F-1}$.
Here, if the concentration $\kappa$ approaches to $\infty$, then $P_\btheta(\bZq|\bx^{(n)})$ becomes the Kronecker delta function $\delta_{\bk,\hat{\bk}^{(n)}}$.

%%----- Section
\section{Experimental Details}
\label{sec:app_details_experiments}
Throughout all the experiments, we apply the same annealing schedule as that used by~\citet{jang2017categorical} for the temperature parameter of Gumbel--softmax, which is $\tau = \exp(10^{-5}\cdot t)$, where $t$ denotes the global training step.
%Throughout all the experiments, we annealed the temperature parameter of Gumbel-softmax, $\tau$, using the common schedule $\tau = \exp(10^{-5}\cdot t)$ of the global training step $t$ to make categorical distributions reparameterizable, which was used in~\citet{jang2017categorical}.
We set the VQ-VAE hyperparameter $\beta$ in \beqref{eq:objective_vq} and weight decay $\gamma$ in EMA to 0.25 and 0.99, respectively, as suggested in \citet{van2017neural}.
For SQ-VAEs, we use $\bZ=\bZq$ instead of sampling $\bZ\sim{}p_\bvarphi(\bZ|\hat{\bZ}_\mathrm{q})$ in the Monte Carlo estimate of the expectations in \beqref{eq:elbo_sqvae} as in Algorithms~\ref{alg:gaussian} and \ref{alg:vmf}, which stabilizes the estimation of Monte Carlo.

%-- subsection
\subsection{Gaussian SQ-VAE on Image Datasets}
\label{sec:app_sub_details_gaussian_vision}

%- subsubsection
\subsubsection{Datasets and Preprocessing}

\paragraph{MNIST and Fashion-MNIST}
They contain 28$\times$28 grayscale images, which are categorized into 10 classes.
We use the default train/test split (60,000/10,000 samples) and further split 10,000 samples from the training set as the validation set.
% We use the default train/test split (60,000/10,000 samples).
% For training, we further divide the train set into two sets: training set (50,000 samples) and validation set (10,000 samples). 

\paragraph{CIFAR10}
CIFAR10 contains 10 classes of 32$\times$32 RGB images.
%We use the default train/test split (50,000/10,000 samples).
We use the default train/test split (50,000/10,000 samples) and further split 10,000 samples from  the training set as the validation set.
% For training, we further divide the train set into two sets: training set (40,000 samples) and validation set (10,000 samples).

\paragraph{CelebA}
CelebA consists of 202,599 colored face images.
We use the default train/valid/test split.
We preprocess the images with a center cropping of 140$\times$140 and resize them to 64$\times$64 using the bilinear interpolation following these previous studies~\citep{tolstikhin2018wasserstein,ghosh2019from}.

%- subsubsection
\subsubsection{Model Description and Training}

We adopt the ConvResNets from the GitHub repository of DeepMind \texttt{sonnet/examples/vqvae\_example.ipynb}\footnote{\url{https://github.com/deepmind/sonnet/blob/v2/examples/vqvae_example.ipynb}}.
These networks include convolutional layers, transpose convolutional layers, and ResBlocks. Their notations are listed in Table~\ref{tb:block_list}.
\begin{table}
\centering
    \caption{Notations of network layers used on image datasets.} 
    \vskip 0.1in
    \renewcommand{\arraystretch}{1.1}
    % \resizebox{\columnwidth}{!}{
    \small
    \begin{tabular}{l|l}
        \bhline{0.8pt}
        Notation & Description\\
        \bhline{0.8pt}
        $\mathrm{Conv}_{n}^{(4\times4)}$
        & 2D Convolutional layer (channel$=n$, kernel$=4\times4$, stride$=2$) \\
        $\mathrm{Conv}_{n}^{(3\times3)}$
        & 2D Convolutional layer (channel$=n$, kernel$=3\times3$, stride$=1$) \\
        $\mathrm{Conv}_{n}^{(1\times1)}$
        & 2D Convolutional layer (channel$=n$, kernel$=1\times1$, stride$=1$) \\
        $\mathrm{ConvT}_{n}^{(4\times4)}$
        & 2D Transpose convolutional layer (channel$=n$, kernel$=4\times4$, stride$=2$) \\
        $\mathrm{ConvT}_{n}^{(3\times3)}$
        & 2D Transpose convolutional layer (channel$=n$, kernel$=3\times3$, stride$=1$) \\
        $\mathrm{ResBlock}_{n}$
        & Resblock ($\mathrm{ReLU}\to\mathrm{Conv}_{n}^{(3\times3)}\to\mathrm{BatchBorm}\to\mathrm{ReLU}\to\mathrm{Conv}_{n}^{(1\times1)}\to\mathrm{BatchBorm}$ $+$ identity mapping) \\
        \bhline{0.8pt}
    \end{tabular}
    % }
    \label{tb:block_list}
    \vskip -0.1in
\end{table}
We use the following network architectures for the encoders and decoders on MNIST and Fashion-MNIST:
\begin{flalign*}
  \bx\in\mathbb{R}^{28\times28}
  &\to\mathrm{Conv}_{d_b/2}^{(4\times4)}\to\mathrm{BatchNorm}\to\mathrm{ReLU}&\text{size: }(d_b/2,14,14)\\
  &\to\mathrm{Conv}_{d_b}^{(4\times4)}&\text{size of }(d_b,7,7)\\
  &\to[\mathrm{ResBlock}_{d_b}]_{\times{}N_\mathrm{resblock}}&\text{size of }(d_b,7,7)\\
  \bZq\in\mathbf{B}^{7\times7}\subset\mathbb{R}^{d_b\times7\times7}
  &\to[\mathrm{ResBlock}_{d_b}]_{\times{}N_\mathrm{resblock}}&\text{size of }(d_b,7,7)\\
  &\to\mathrm{ConvT}_{d_b/2}^{(4\times4)}\to\mathrm{BatchNorm}\to\mathrm{ReLU}&\text{size: }(d_b/2,14,14)\\
  &\to\mathrm{ConvT}_{1}^{(4\times4)}\to\mathrm{Sigmoid}&\text{size: }(1,28,28),
\end{flalign*}
where $N_\mathrm{resblock}$ is set to 2 and 6 for MNIST and Fashion-MNIST, respectively.
We use the following network architectures for CIFAR10 and CelebA:
\begin{flalign*}
  \bx\in\mathbb{R}^{32\times32\times3}
  &\to\mathrm{Conv}_{d_b/2}^{(4\times4)}\to\mathrm{BatchNorm}\to\mathrm{ReLU}&\text{size: }(d_b/2,w/2,h/2)\\
  &\to\mathrm{Conv}_{d_b}^{(4\times4)}\to\mathrm{BatchNorm}\to\mathrm{ReLU}&\text{size: }(d_b,w/4,h/4)\\
  &\to\mathrm{Conv}_{d_b}^{(3\times3)}&\text{size of }(d_b,w/4,h/4)\\
  &\to[\mathrm{ResBlock}_{d_b}]_{\times{}6}&\text{size of }(d_b,w/4,h/4)\\
  \bZq\in\mathbf{B}^{w/4\times{}h/4}\subset\mathbb{R}^{d_b\times{}w/4\times{}h/4}
  &\to[\mathrm{ResBlock}_{d_b}]_{\times{}6}&\text{size of }(d_b,w/4,h/4)\\
  &\to\mathrm{ConvT}_{d_b/2}^{(3\times3)}\to\mathrm{BatchNorm}\to\mathrm{ReLU}&\text{size: }(d_b,w/4,h/4)\\
  &\to\mathrm{ConvT}_{d_b/2}^{(4\times4)}\to\mathrm{BatchNorm}\to\mathrm{ReLU}&\text{size: }(d_b/2,w/2,h/2)\\
  &\to\mathrm{ConvT}_{3}^{(4\times4)}\to\mathrm{Sigmoid}&\text{size: }(3,w,h),
\end{flalign*}
where $w$ and $h$ denote the width and height of the target images, respectively.

We use the Adam optimizer~\citep{kingma2014adam} with initial learning rates of 0.0003 and 0.001 for VQ-VAE and the other models, respectively.
The learning rate will be halved every 3 epochs if the validation loss is not improving.
We train 100 epochs with the minibatch size of 32 for MNIST, Fashion-MNIST, and CIFAR10 and 70 epochs for CelebA.

Regarding the codebook reset applied in Figure~\ref{fig:results_codebook_size} with EMA, we adopt the same procedure as that adopted by~\citet{william2020hierarchical}.
At every 20th batch, the two codes that are most and least used in the recent 20 batches are found, which are denoted as $\bb_\mathrm{most}$ and $\bb_\mathrm{least}$, respectively.
If the usage of $\bb_\mathrm{least}$ is less than 3\% of that of $\bb_\mathrm{most}$, the position of $\bb_\mathrm{least}$ is reset to $\bb_\mathrm{least}^\mathrm{reset}\sim\mathcal{N}(\bb_\mathrm{most},s_\mathrm{reset}^2\bI)$ with $s_\mathrm{reset}^2=0.01$.

We set the dimension of the latent space for the VAE on CelebA to 72 such that the number of bits used to represent the latent space is the same as that of the other models, i.e., $32~\text{bits}\times72=9~\text{bits}\times16\times16$.
We use the same architecture for VAE as that of the other models except appending a linear layer at the end of the encoder and doing the same at the beginning of the decoder. This is to adjust the dimension of latent variables.

%- subsubsection
\subsubsection{Reconstructed and Generated Samples on CelebA 64$\times$64}

\begin{figure*}[t]
\vskip 0.1in
  \centering
   \includegraphics[width=.98\textwidth]{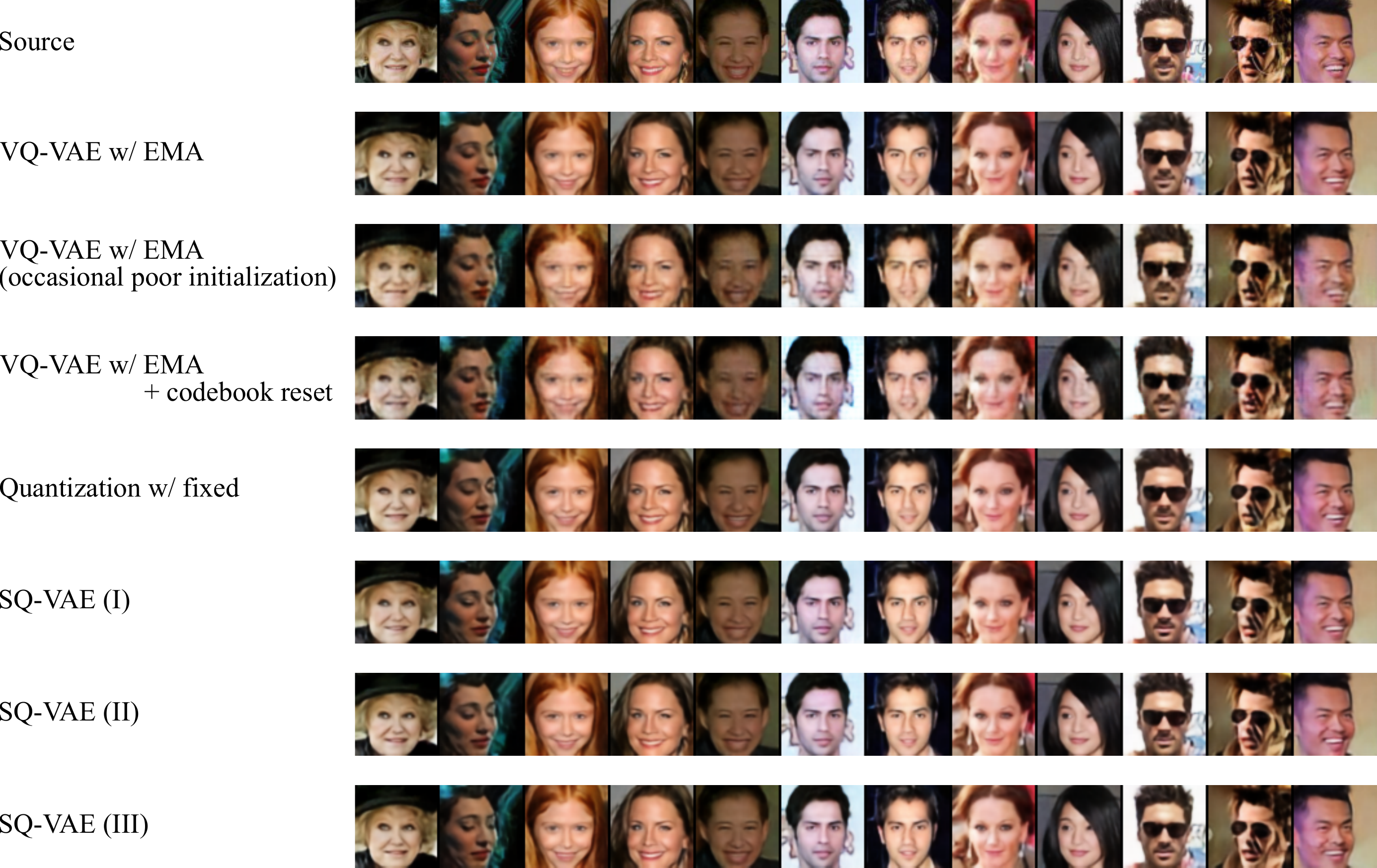}
   \caption{
   Reconstructed samples of CelebA 64$\times$64.
   }
   \label{fig:vision_reconstruction}
\vskip -0.1in
\end{figure*}

\begin{figure*}[t]
\vskip 0.1in
  \centering
  \subfigure[VQ-VAE]{\includegraphics[width=0.36\textwidth]{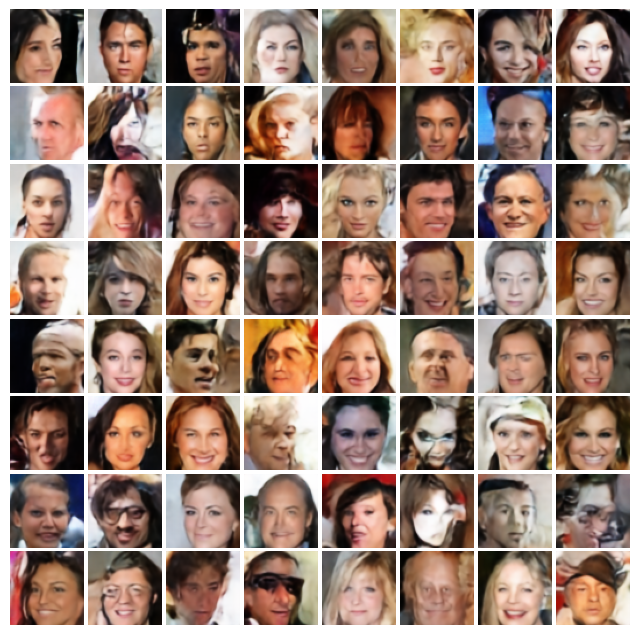}}
  \hspace{10pt}
  \subfigure[VQ-VAE (with occasional poor initialization)]{\includegraphics[width=0.36\textwidth]{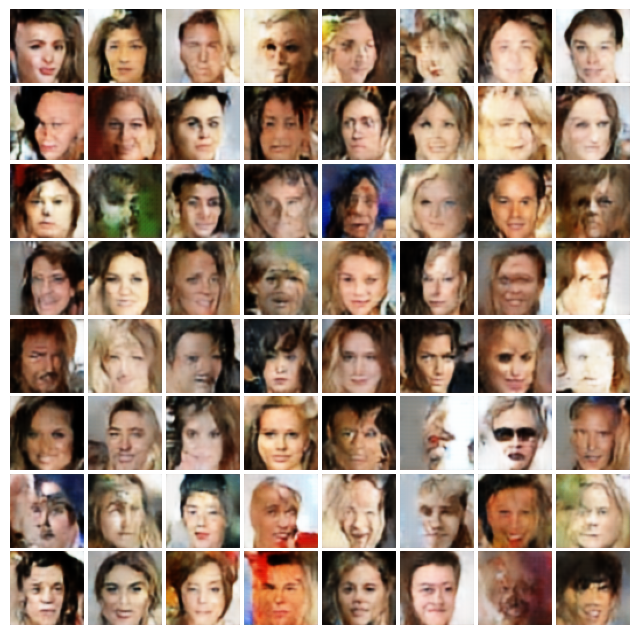}}
  \\
  \subfigure[Quantization w/ fixed $\sigma_\mathrm{q}^2$]{\includegraphics[width=0.36\textwidth]{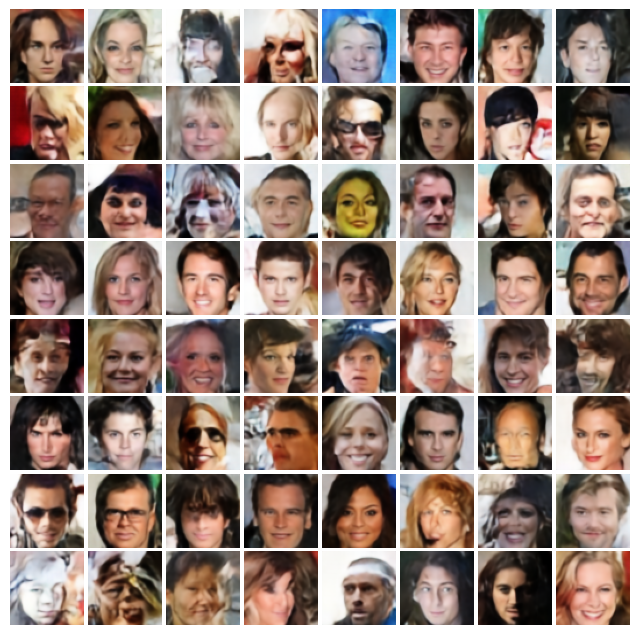}}
  \hspace{10pt}
  \subfigure[SQ-VAE (\uppercase\expandafter{\romannumeral 1})]{\includegraphics[width=0.36\textwidth]{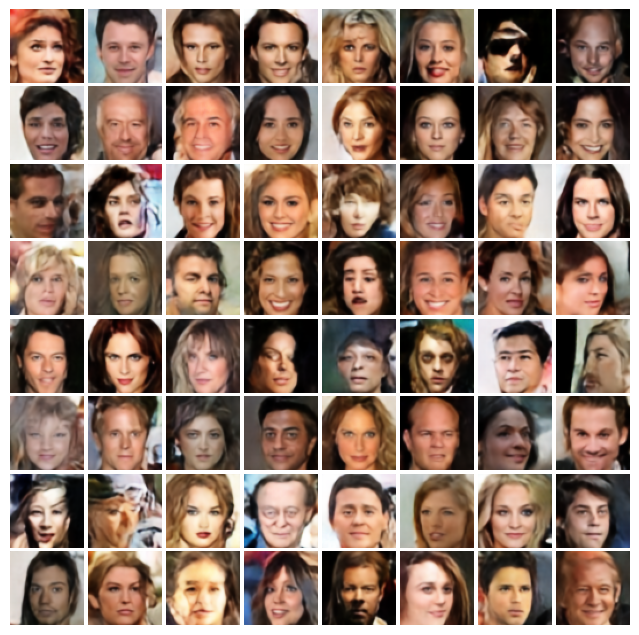}}
  \\
  \subfigure[SQ-VAE (\uppercase\expandafter{\romannumeral 2})]{\includegraphics[width=0.36\textwidth]{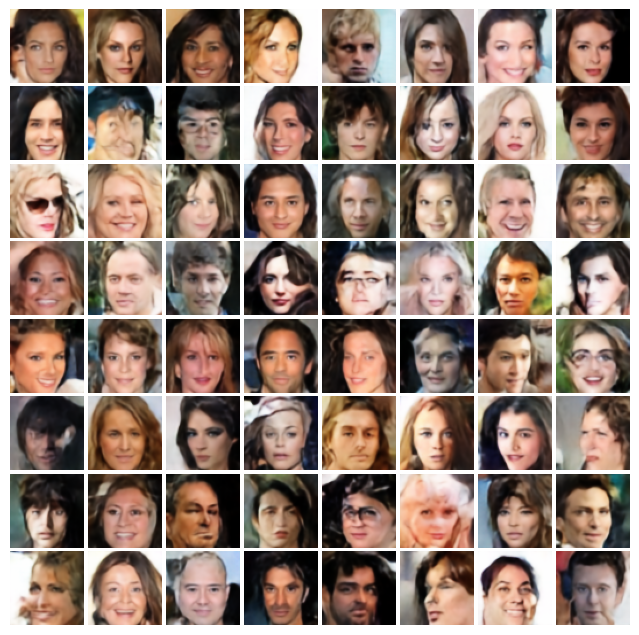}}
  \hspace{10pt}
  \subfigure[SQ-VAE (\uppercase\expandafter{\romannumeral 3})]{\includegraphics[width=0.36\textwidth]{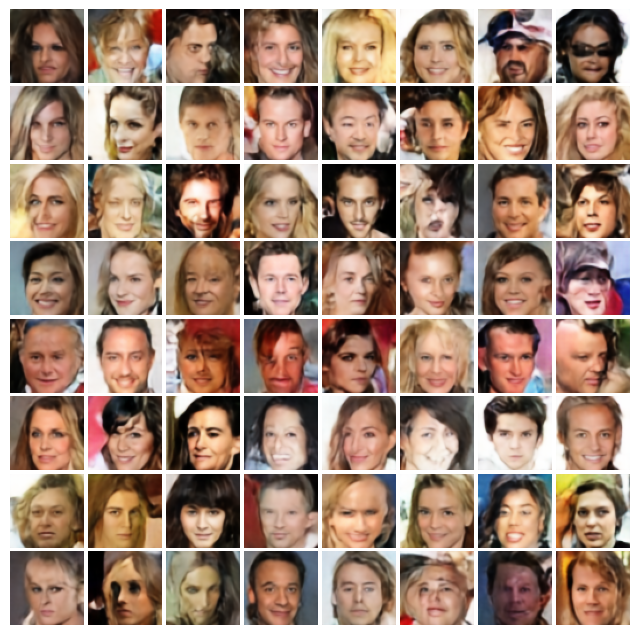}}
  \caption{Generated samples of CelebA 64$\times$64.}
  \label{fig:vision_prior}
\vskip -0.1in
\end{figure*}

We show examples of reconstructed images and images generated with the learned approximated prior in Figures~\ref{fig:vision_reconstruction} and \ref{fig:vision_prior}.
We adopt PixelCNN~\citep{van2016pixel} as the estimator of the prior while we believe the quality of synthetic samples of SQ-VAE can be improved by using other stronger autoregressive estimators~\citep{child2019generating}.
We observe that the quality of the samples from VQ-VAE varies depending on random seeds, as shown in Figure~\ref{fig:vision_reconstruction}.

% \begin{figure*}[t]
% \vskip 0.1in
%   \centering
%   \subfigure[VQ-VAE]{\includegraphics[width=0.9\textwidth]{figures/vision/vqvae_prior.png}}
%   \\
%   \subfigure[Quantization w/ fixed $\sigma_\mathrm{q}^2$]{\includegraphics[width=0.9\textwidth]{figures/vision/fixsqvae_prior.png}}
%   \\ %\hspace{10pt}
%   \subfigure[SQ-VAE (\uppercase\expandafter{\romannumeral 1})]{\includegraphics[width=0.9\textwidth]{figures/vision/sqvae1_prior.png}}
%   \\
%   \subfigure[SQ-VAE (\uppercase\expandafter{\romannumeral 2})]{\includegraphics[width=0.9\textwidth]{figures/vision/sqvae2_prior.png}}
%   \\ %\hspace{10pt}
%   \subfigure[SQ-VAE (\uppercase\expandafter{\romannumeral 3})]{\includegraphics[width=0.9\textwidth]{figures/vision/sqvae3_prior.png}}
%   \caption{Reconstructed samples of CelebA 64$\times$64.}
%   \label{fig:vision_prior}
% \vskip -0.1in
% \end{figure*}

%-- subsection
\subsection{Gaussian SQ-VAE on Speech Dataset}
\label{sec:app_sub_details_gaussian_speech}

%- subsubsection
\subsubsection{Datasets and Preprocessing}

\paragraph{VCTK}
VCTK version 0.80~\citep{veaux2017cstr} is a speech dataset of $109$ English speakers.
Each speaker reads out about 400 sentences, which were mostly selected from a newspaper.
We use $90\%$ of the samples as the training set, and the remaining $10\%$ as the test set.

We preprocess waveforms following the approach of the GitHub repository \texttt{kan-bayashi/ParallelWaveGAN}\footnote{\url{https://github.com/kan-bayashi/ParallelWaveGAN}}, which is one of the most popular repositories related to speech generation and provides implementations of several types of vocoder such as Parallel WaveGAN~\citep{yamamoto2020parallel} and MelGAN~\citep{Kumar2019melgan}.
The details of the preprocessing are as follows.
\begin{enumerate}
\item We resample $48000$ Hz signals to $24000$ Hz.
\item We extract $80$-dimensional Mel spectrogram features with a $50.0$ ms Hann window, $12.5$ ms frame shift, $2048$-point FFT, and $80$ Hz and $7600$ Hz frequency cutoffs.
\item We convert the amplitude spectrograms to dB-scaled spectrograms and clip the bins below $-200.0$ dB.
\item We perform dimension-wise standardization.
\end{enumerate}

\paragraph{ZeroSpeech 2019}
ZeroSpeech 2019 English~\citep{dunbar19zero} is a multi-speaker corpus sampled at $16000$ Hz.
This corpus was originally made for a speech task called acoustic unit discovery.
It consists of four subsets: Train Voice Dataset (directory \texttt{train/voice}), Train Unit Dataset (directory \texttt{train/unit}), Train Parallel Dataset (directories \texttt{train/parallel/source} and \texttt{train/parallel/voice}), and Test Dataset (directory \texttt{test}).
We use the \texttt{.wav} files under the directories \texttt{train/voice} and \texttt{train/unit} for training data, following \citet{niekerk2020vector}, and the \texttt{.wav} files under the directory \texttt{train/parallel/voice} for testing.

We preprocess waveforms following the approach of \citet{niekerk2020vector}.
The details of the preprocessing are as follows:
\begin{enumerate}
\item We scale the maximum amplitude of each audio signal to $0.999$.
\item We pre-emphasize the scaled audio signals with a first-order autoregressive filter: $y[n] - 0.97 * y[n-1]$.
\item We extract $80$-dimensional Mel-spectrogram features with a $25.0$ ms Hann window, $10.0$ ms frame shift, $2048$-point FFT, and $50$ Hz and $8000$ Hz frequency cutoffs.
\item We convert the amplitudes to decibels and clip those bins that are $80.0$ dB lower than the maximum.
\item We rescale the dB-scaled Mel-spectrogram by dividing it by $80.0$.
\end{enumerate}

%- subsubsection
\subsubsection{Model Description and Training}
We adopt the model proposed by \citet{niekerk2020vector}\footnote{\url{https://github.com/bshall/ZeroSpeech}} as a baseline and replace its RNN-based vocoder with a projection layer.
The encoder consists of a stack of five convolutional layers, which downsamples the input by $2$, and the latent representation is quantized with 512 codes ($K=512$).
The decoder aims to reconstruct the normalized log-Mel spectrogram, conditioned on both the quantized latent representation and a speaker embedding.
Time-jitter regularization~\citep{chorowski2019unsupervised} is applied with a replacement probability of $0.5$.
\begin{flalign*}
  x\in\mathbb{R}^{80\times N}
  &\to\mathrm{Conv}_{768}^{(3)}\to\mathrm{BatchNorm}\to\mathrm{ReLU}&\text{size: }(768,N)\\
  &\to\mathrm{Conv}_{768}^{(3)}\to\mathrm{BatchNorm}\to\mathrm{ReLU}&\text{size: }(768,N)\\
  &\to\mathrm{Conv}_{768}^{(4)}\to\mathrm{BatchNorm}\to\mathrm{ReLU}&\text{size: }(768,N/2)\\
  &\to\mathrm{Conv}_{768}^{(3)}\to\mathrm{BatchNorm}\to\mathrm{ReLU}&\text{size: }(768,N/2)\\
  &\to\mathrm{Conv}_{768}^{(3)}\to\mathrm{BatchNorm}\to\mathrm{ReLU}&\text{size: }(768,N/2)\\
  &\to\mathrm{FC}_{64}&\text{size: }(64,N/2),\\
  z\in\mathbb{R}^{64\times N/2}
  &\to\mathrm{Jitter}&\text{size: }(64,N/2)\\
  &\to\mathrm{Concat(Embedding}_{64}\mathrm{)}&\text{size: }(128,N/2)\\
  &\to\mathrm{Interpolate}&\text{size: }(128,N)\\
  &\to\mathrm{BiGRU}_{128}&\text{size: }(256,N)\\
  &\to\mathrm{BiGRU}_{128}&\text{size: }(256,N)\\
  &\to\mathrm{FC}_{80}&\text{size: }(80,N).
\end{flalign*}
The notations of the layers are listed in Table~\ref{tb:block_list_speech}, and $N$ denotes the number of frames of an input data.
\begin{table}
\centering
    \caption{Notations of network layers on speech datasets.} 
    \vskip 0.1in
    \renewcommand{\arraystretch}{1.1}
    % \resizebox{\columnwidth}{!}{
    \small
    \begin{tabular}{l|l}
        \bhline{0.8pt}
        Notation & Description\\
        \bhline{0.8pt}
        $\mathrm{Conv}_{n}^{(3)}$
        & 1D Convolutional layer (channel$=n$, kernel$=3$, stride$=1$) \\
        $\mathrm{Conv}_{n}^{(4)}$
        & 1D Convolutional layer (channel$=n$, kernel$=4$, stride$=2$) \\
        $\mathrm{Jitter}$
        & Time-jitter regularization~\citep{chorowski2019unsupervised} operation \\
        $\mathrm{Concat(}\dots\mathrm{)}$
        & Concatenation operation \\
        $\mathrm{Embedding}_{n}$
        & Lookup table (dimension$=n$) \\
        $\mathrm{Interpolate}$
        & Nearest-neighbor interpolation operation \\
        $\mathrm{BiGRU}_{n}$
        & Bidirectional gated recurrent unit~\citep{cho2014learning} layer (dimension$=n$) \\
        \bhline{0.8pt}
    \end{tabular}
    % }
    \label{tb:block_list_speech}
    \vskip -0.1in
\end{table}

Following \citet{niekerk2020vector}, the model is trained on a minibatch of $52$ segments, each $32$ frames long.
We use the Adam optimizer with an initial learning rate of $4\cdot10^{-4}$, and the learning rate is halved after $300$k and $400$k steps. The network is trained for a total of $500$k steps.

% For VQ-VAE, we set the hyperparameter $\beta$ in \beqref{eq:objective_vq} to $0.25$.
% For SQ-VAE, we annealed the temperature parameter of Gumbel-softmax, $\tau$, using the schedule $\tau = \exp(10^{-5}\cdot t)$ of the global training step $t$.

We do not apply SQ-VAE (\Rnum{2}) in this evaluation because of the variable length property of speech data and the different manipulations of speech signals between training and inference.
SQ-VAE (\Rnum{2}) has one parameter $\sigma_\bvarphi^2(\mathbf{x})$ for one trimmed speech signal (32 frames long) in training but has to deal with longer signals in inference.
$\sigma_\bvarphi^2(\mathbf{x})$ is calculated based on the content of a trimmed signal during training regardless of the content of the whole signal.
That means there is a discrepancy between training and inference with SQ-VAE (\Rnum{2}).
This discrepancy does not exist when it is applied to image datasets, where the size of images is fixed.
On the other hand, SQ-VAE (\Rnum{3}) has a parameter $\sigma_{\bvarphi,i}^2(\mathbf{x})$ for every 2 consecutive frames regardless of the signal length.
This parameterization is applicable even when signals are trimmed to 32 frames long during training but are not trimmed during inference.
The same is true for SQ-VAE (\Rnum{4}).

%- subsubsection
\subsubsection{Details of Experimental Results}

\begin{table}
\centering
    \caption{Evaluation on VCTK and ZeroSpeech 2019. The MSE (dB$^2$) of sample reconstruction is evaluated using the test set.} 
    \vskip 0.1in
    \renewcommand{\arraystretch}{1.1}
    \begin{tabular}{l|c|c}
        \bhline{0.8pt}
        \multirow{2}{*}{Model} & \multicolumn{2}{c}{MSE (dB$^2$)} \\
        \cline{2-3}
         & VCTK & ZeroSpeech 2019 \\
        \bhline{0.8pt}
        VQ-VAE w/ EMA ($\sigma^2=10^{-2}$) & 31.25 $\pm$ 0.40 & 35.20 $\pm$ 1.21 \\
        VQ-VAE w/ EMA ($\sigma^2=10^{-1}$) & 30.89 $\pm$ 0.46 & 34.33 $\pm$ 1.57 \\
        VQ-VAE w/ EMA ($\sigma^2=10^{0}$) & 29.59 $\pm$ 0.25 & 40.40 $\pm$ 1.24 \\
        VQ-VAE w/ EMA ($\sigma^2=10^{1}$) & 36.92 $\pm$ 0.95 & 79.30 $\pm$ 42.95 \\
        \hline
        Gaussian SQ-VAE (\Rnum{1}) & 25.52 $\pm$ 0.08 & 33.17 $\pm$ 1.11 \\
        Gaussian SQ-VAE (\Rnum{3}) & 25.94 $\pm$ 0.22 & 34.35 $\pm$ 1.07 \\
        Gaussian SQ-VAE (\Rnum{4}) & \textbf{24.68} $\pm$ 0.21 & \textbf{32.32} $\pm$ 0.88 \\
        \bhline{0.8pt}
    \end{tabular}
    \label{tb:speech_detailed_mse_result}
    \vskip -0.1in
\end{table}

The MSEs of VQ-VAE models with various $\sigma^2$ values are shown in Table~\ref{tb:speech_detailed_mse_result}.

%- subsubsection
\subsubsection{Reconstructed Samples}
\begin{figure*}[t]
\vskip 0.1in
  \centering
  \subfigure[Source]{\includegraphics[width=0.36\textwidth]{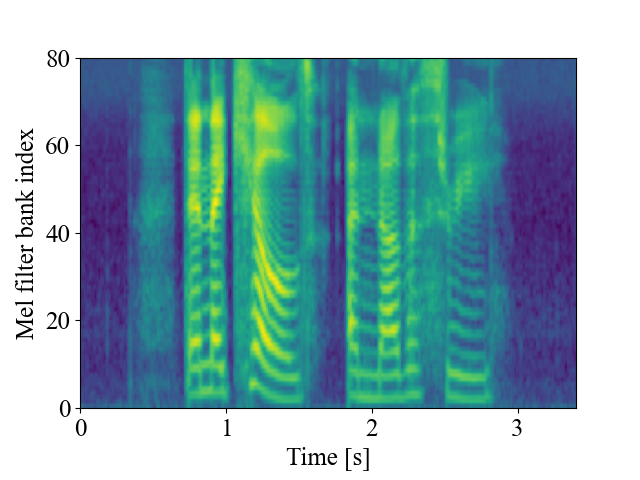}}
  \subfigure[VQ-VAE w/ EMA ($\sigma^2=10^{-2}$)]{\includegraphics[width=0.36\textwidth]{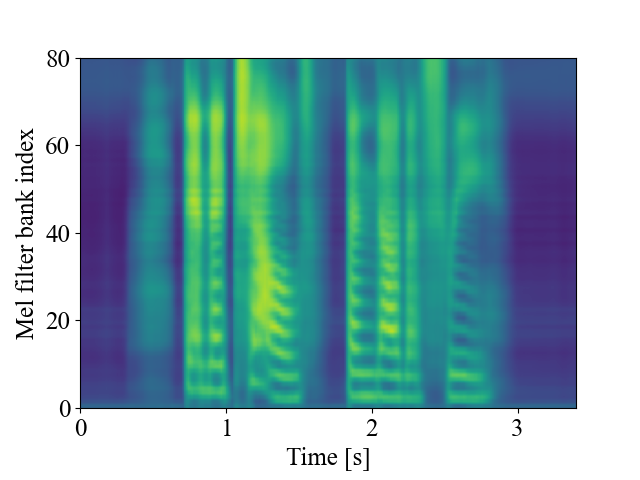}}
  \\
  \subfigure[VQ-VAE w/ EMA ($\sigma^2=10^{-1}$)]{\includegraphics[width=0.36\textwidth]{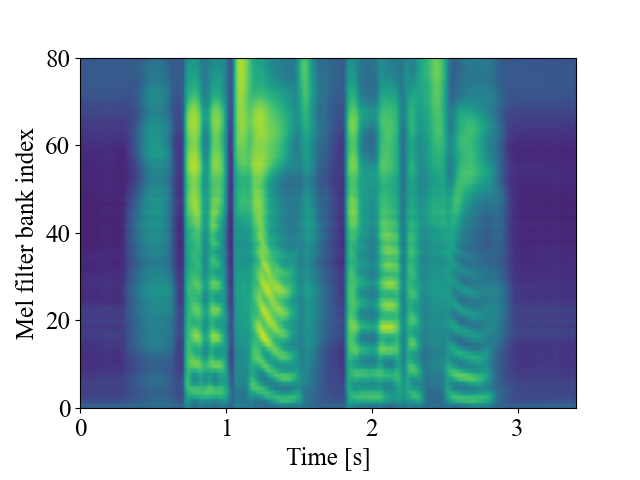}}
  \subfigure[VQ-VAE w/ EMA ($\sigma^2=10^{0}$)]{\includegraphics[width=0.36\textwidth]{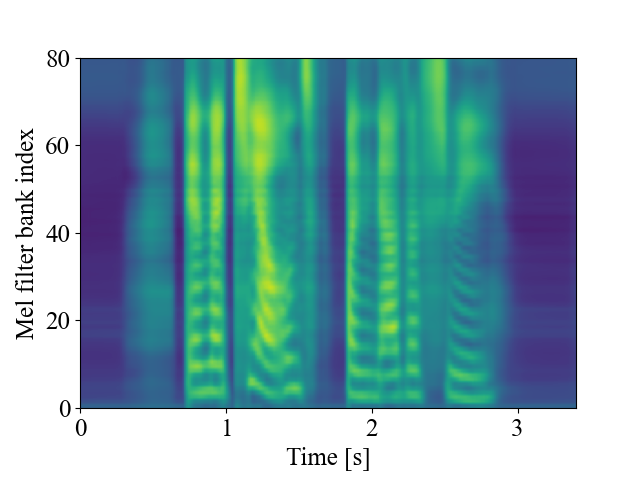}}
  \\
  \subfigure[VQ-VAE w/ EMA ($\sigma^2=10^{1}$)]{\includegraphics[width=0.36\textwidth]{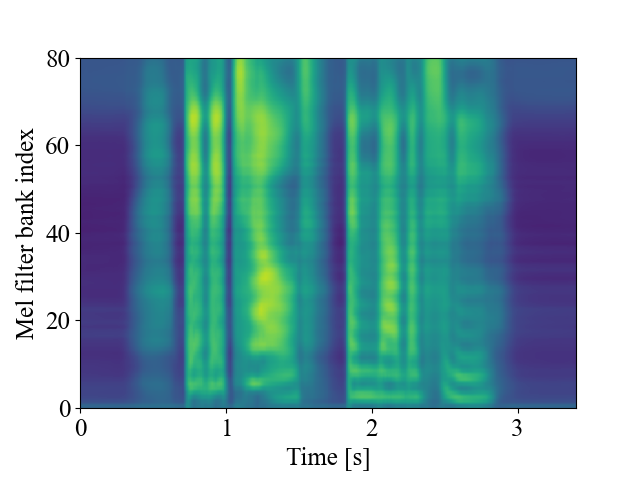}}
  \subfigure[SQ-VAE (\uppercase\expandafter{\romannumeral 1})]{\includegraphics[width=0.36\textwidth]{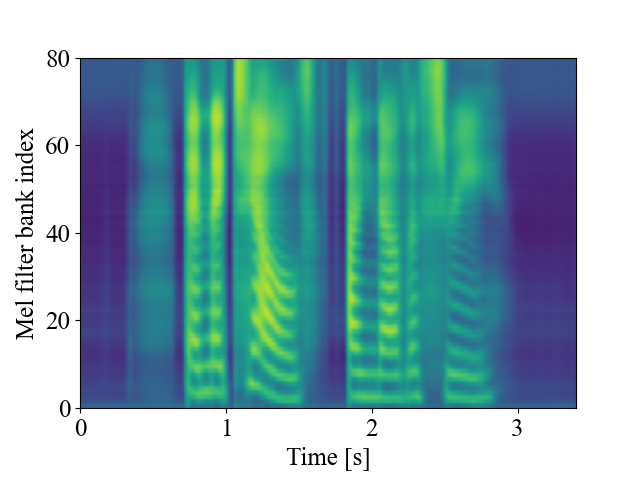}}
  \\
  \subfigure[SQ-VAE (\uppercase\expandafter{\romannumeral 3})]{\includegraphics[width=0.36\textwidth]{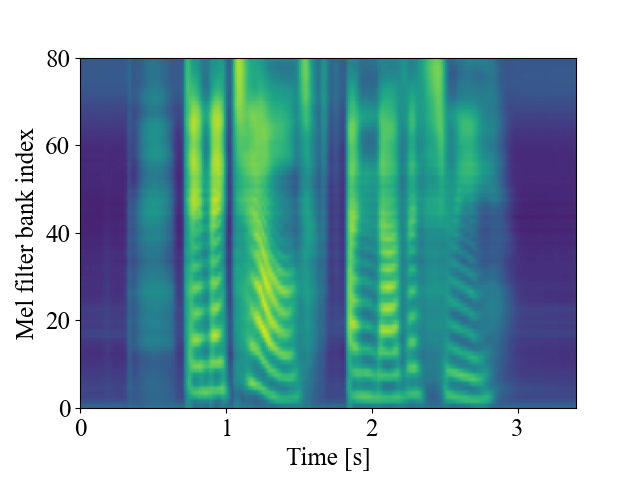}}
  \subfigure[SQ-VAE (\uppercase\expandafter{\romannumeral 4})]{\includegraphics[width=0.36\textwidth]{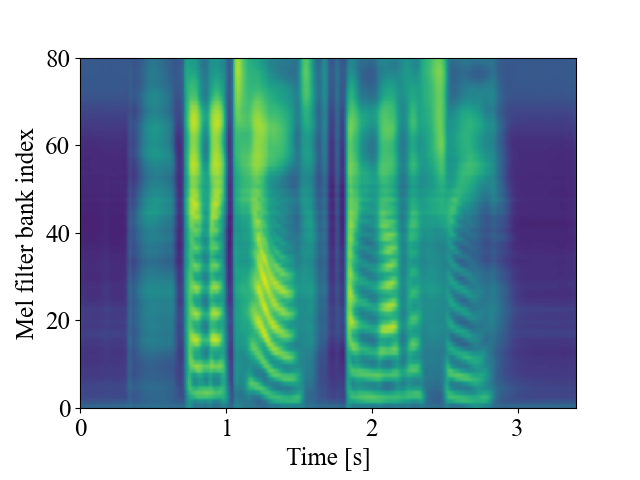}}
  \caption{Reconstructed log-Mel spectrograms of \texttt{p323\_064} in VCTK.}
  \label{fig:speech_reconstruction_1}
\vskip -0.1in
\end{figure*}

\begin{figure*}[t]
\vskip 0.1in
  \centering
  \subfigure[Source]{\includegraphics[width=0.36\textwidth]{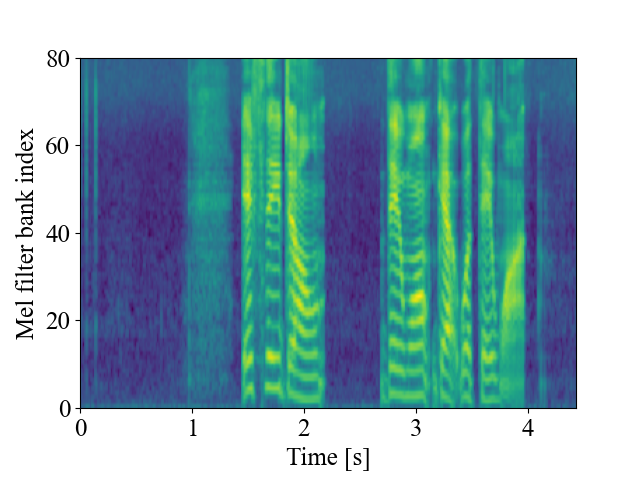}}
  \subfigure[VQ-VAE w/ EMA ($\sigma^2=10^{-2}$)]{\includegraphics[width=0.36\textwidth]{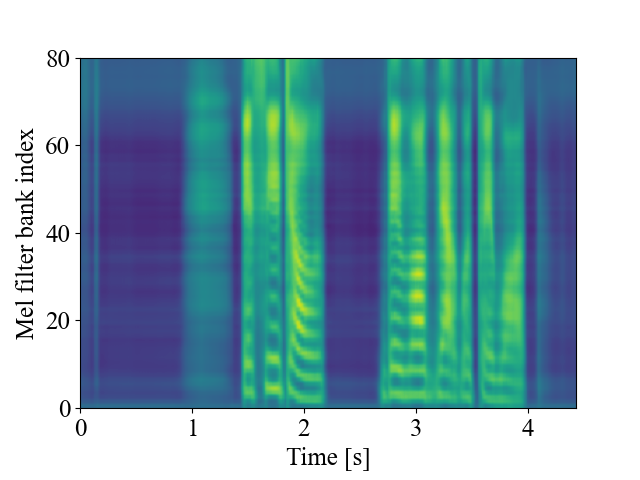}}
  \\
  \subfigure[VQ-VAE w/ EMA ($\sigma^2=10^{-1}$)]{\includegraphics[width=0.36\textwidth]{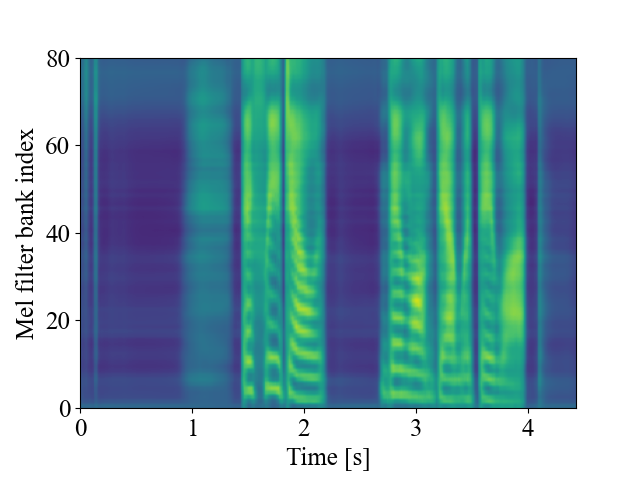}}
  \subfigure[VQ-VAE w/ EMA ($\sigma^2=10^{0}$)]{\includegraphics[width=0.36\textwidth]{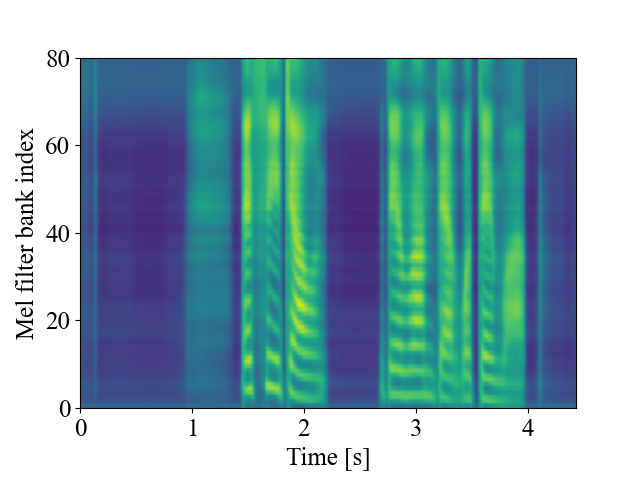}}
  \\
  \subfigure[VQ-VAE w/ EMA ($\sigma^2=10^{1}$)]{\includegraphics[width=0.36\textwidth]{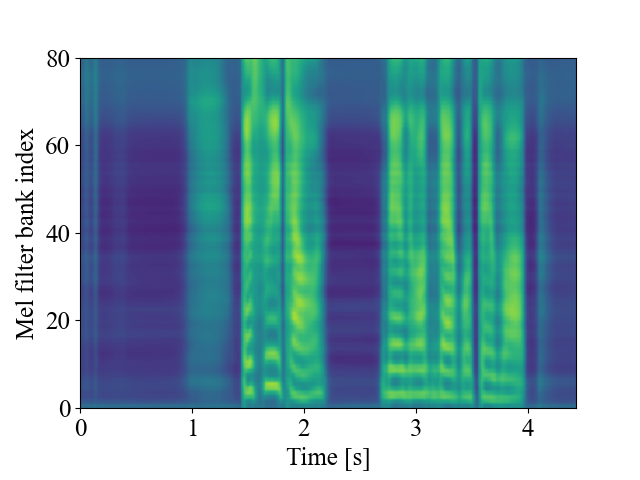}}
  \subfigure[SQ-VAE (\uppercase\expandafter{\romannumeral 1})]{\includegraphics[width=0.36\textwidth]{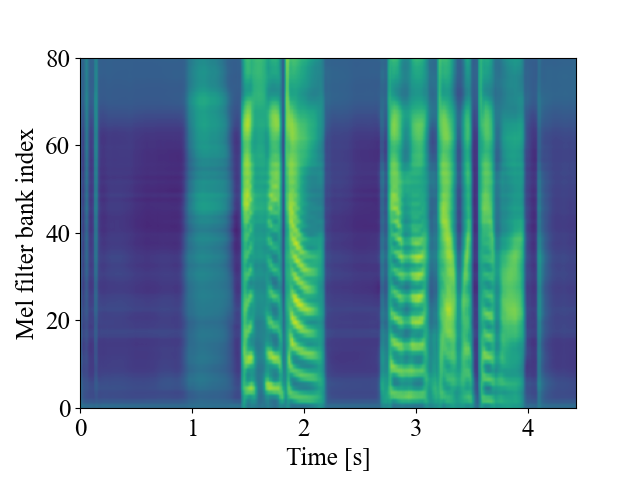}}
  \\
  \subfigure[SQ-VAE (\uppercase\expandafter{\romannumeral 3})]{\includegraphics[width=0.36\textwidth]{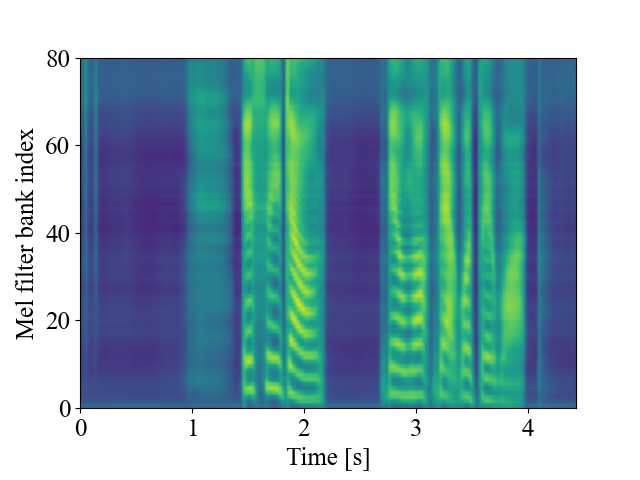}}
  \subfigure[SQ-VAE (\uppercase\expandafter{\romannumeral 4})]{\includegraphics[width=0.36\textwidth]{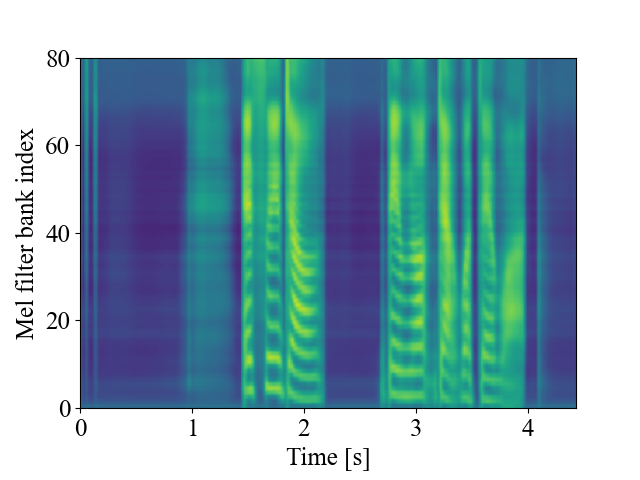}}
  \caption{Reconstructed log-Mel spectrograms of \texttt{p341\_285} in VCTK.}
  \label{fig:speech_reconstruction_2}
\vskip -0.1in
\end{figure*}

As a sample of demonstration, we randomly select two speech signals from our test split of VCTK and show their reconstructed log-Mel spectrograms from VQ-VAE and SQ-VAE in Figures~\ref{fig:speech_reconstruction_1} and ~\ref{fig:speech_reconstruction_2}, respectively.

%- subsubsection
\subsubsection{Acoustic Unit Discovery}
We compare the performance of SQ-VAE with that of VQ-VAE in the acoustic unit discovery task, which evaluates the feasibility of a representation to discriminate phonetic units. VQ-VAE is a popular approach used for this task~\citep{chorowski2019unsupervised,eloff2019unsupervised,niekerk2020vector,tjandra2020transformer}.

We follow the evaluation scheme of ZeroSpeech 2019~\citep{dunbar19zero}  and use the minimal pair ABX discriminability test~\citep{schatz2013evaluating} for comparison.
This test asks whether a triphone X is more similar to triphone A than triphone B.
Here, A and X are instances of the same triphone (e.g., ``beg''), whereas B differs in the middle phone (e.g., ``bag'').
Moreover, A and B are selected from the same speaker, but X is selected from a different speaker.
An ABX score is reported as an aggregated error rate over all pairs of triphones in the test set.
The lower the ABX score, the better the performance.
We compute the ABX scores of the trained models using Test Dataset (the \texttt{.wav} files under the directory \texttt{test}).

\begin{table}
\centering
    \caption{Evaluation on ZeroSpeech 2019. The average ABX score is evaluated on the test set.} 
    \vskip 0.1in
    \renewcommand{\arraystretch}{1.1}
    \begin{tabular}{l|c}
        \bhline{0.8pt}
        Model & ABX score $\downarrow$ \\
        \hline
        VQ-VAE w/ EMA ($\sigma^2=10^{-2}$) & 22.36 $\pm$ 0.14 \\
        VQ-VAE w/ EMA ($\sigma^2=10^{-1}$) & 22.53 $\pm$ 0.33 \\
        VQ-VAE w/ EMA ($\sigma^2=10^{0}$) & 23.78 $\pm$ 0.74 \\
        VQ-VAE w/ EMA ($\sigma^2=10^{1}$) & 35.40 $\pm$ 6.73 \\
        \hline
        Gaussian SQ-VAE (\Rnum{1}) & \textbf{22.11} $\pm$ 0.62 \\
        Gaussian SQ-VAE (\Rnum{3}) & 22.13 $\pm$ 0.29 \\
        Gaussian SQ-VAE (\Rnum{4}) & 24.46 $\pm$ 0.66 \\
        \bhline{0.8pt}
    \end{tabular}
    \label{tb:speech_aud_result}
    \vskip -0.1in
\end{table}

Surprisingly, as shown by the results in Table~\ref{tb:speech_aud_result}, SQ-VAE is on par with VQ-VAE in this experiment, with no statistically significant difference. % at least with this dataset and this model architecture.
%There is no significant difference between the performance of VQ-VAE and that of SQ-VAE.
In addition, SQ-VAE (\Rnum{4}) performs slightly worse than SQ-VAE (\Rnum{1}) or SQ-VAE (\Rnum{3}) here, although SQ-VAE (\Rnum{4}) is better in terms of MSE as shown in Table~\ref{tb:speech_mse_result}.
We plan to perform further analysis on the learned representations in our future work.
%Properties of the latent representations need to be analyzed.
%We leave an analysis of them for future work.

%-- subsection
\subsection{vMF SQ-VAE on Vision Dataset}
\label{sec:app_sub_details_vmf_vision}

%- subsubsection
\subsubsection{Datasets and Preprocessing}

\paragraph{CelebAHQ-Mask}
CelebAHQ-Mask is a dataset of colored face images with segmentation maps, which have 19 categories including all facial components and accessories such as skin, nose, eyes, eyebrows, ears, mouth, lip, hair, hat, eyeglass, earring, necklace, neck, and cloth.

\paragraph{Gray-CelebA}
We obtain Gray-CelebA by converting CelebA to grayscale.
As in CelebA, we preprocess the grayscaled images with center cropping of 140$\times$140 and resized to 64$\times$64 using the bilinear interpolation.

%- subsubsection
\subsubsection{Setup}
\paragraph{CelebAHQ-Mask}
We set the codebook capacity to $(d_b,K)=(64,64)$ for all the discrete models.
In SQ-VAE, we introduce a hypersphere $\mathcal{S}^{19-1}$ as a space for  projecting data categories.
We set the projections of the categories $(\bw_l)_{l=1}^L$ to one-hot vectors, where all $\bw_l$ reside on $\mathcal{S}^{18}$ and are orthogonal to each others.

For CIFAR10 and CelebA, we use the following networks for the encoder and decoder, respectively.
\begin{flalign*}
  \bx\in\mathbb{R}^{64\times64\times19}
  &\to\mathrm{Conv}_{32}^{(4\times4)}\to\mathrm{BatchNorm}\to\mathrm{ReLU}&\text{size: }(32,32,32)\\
  &\to\mathrm{Conv}_{64}^{(4\times4)}\to\mathrm{BatchNorm}\to\mathrm{ReLU}&\text{size: }(64,16,16)\\
  &\to\mathrm{Conv}_{64}^{(4\times4)}&\text{size of }(64,8,8)\\
  &\to[\mathrm{ResBlock}_{64}]_{\times2}&\text{size of }(64,8,8)\\
  \bZq\in\mathbf{B}^{8\times8}\subset\mathbb{R}^{64\times8\times8}
  &\to[\mathrm{ResBlock}_{64}]_{\times2}&\text{size of }(64,8,8)\\
  &\to\mathrm{ConvT}_{64}^{(4\times4)}\to\mathrm{BatchNorm}\to\mathrm{ReLU}&\text{size: }(64,16,16)\\
  &\to\mathrm{ConvT}_{32}^{(4\times4)}\to\mathrm{BatchNorm}\to\mathrm{ReLU}&\text{size: }(32,32,32)\\
  &\to\mathrm{ConvT}_{19}^{(4\times4)}&\text{size: }(19,64,64).
\end{flalign*}
In vMF SQ-VAE, the outputs of the encoder and decoder are normalized along the channel axis so that the normalized vectors are on the hypersphere $\mathcal{S}^{64-1}$ and $\mathcal{S}^{19-1}$, respectively.

In this experiment, the optimizer, the initial learning rate and the learning rate scheduling are the same as those in Appendix~\ref{sec:app_sub_details_gaussian_vision}. We run 70 epochs with a minibatch size of 32.

In VAE, we set the dimension of the latent space to 12 such that the number of bits representing the latent space is the same as that of the other models, i.e., $32~\text{bit}\times12=6~\text{bit}\times8\times8$.
We use the same architecture for VAE as those of the other models except adding a linear layer to the end of the encoder.

\paragraph{MNIST and Gray-CelebA}
We set the codebook capacity $(d_b,K)$ to $(64,128)$ and $(64,512)$ on MNIST and Gray-CelebA, respectively.
We introduce a unit circle $\mathcal{S}^{1}$ as a space for projecting data categories and set $(\bw_l)_{l=1}^L$ as $\bw_l=[\cos(\alpha_l),\sin(\alpha_l)]^\top$ with $\alpha_l=\frac{\pi}{L}l$, where all $\bw_l$ reside on $\mathcal{S}^{1}$.
The common network architecture for MNIST and Gray-CelebA is as follows:
\begin{flalign*}
  \bx\in\mathbb{R}^{28\times28\times256}
  &\to\mathrm{Conv}_{32}^{(4\times4)}\to\mathrm{BatchNorm}\to\mathrm{ReLU}&\text{size: }(32,14,14)\\
  &\to\mathrm{Conv}_{64}^{(4\times4)}&\text{size of }(64,7,7)\\
  &\to[\mathrm{ResBlock}_{64}]_{\times{}2}&\text{size of }(64,7,7)\\
  \bZq\in\mathbf{B}^{7\times7}\subset\mathbb{R}^{64\times7\times7}
  &\to[\mathrm{ResBlock}_{64}]_{\times{}2}&\text{size of }(64,7,7)\\
  &\to\mathrm{ConvT}_{32}^{(4\times4)}\to\mathrm{BatchNorm}\to\mathrm{ReLU}&\text{size: }(32,14,14)\\
  &\to\mathrm{ConvT}_{c}^{(4\times4)}&\text{size: }(c,28,28),
\end{flalign*}
and
\begin{flalign*}
  \bx\in\mathbb{R}^{64\times64\times256}
  &\to\mathrm{Conv}_{32}^{(4\times4)}\to\mathrm{BatchNorm}\to\mathrm{ReLU}&\text{size: }(32,32,32)\\
  &\to\mathrm{Conv}_{64}^{(4\times4)}\to\mathrm{BatchNorm}\to\mathrm{ReLU}&\text{size: }(64,16,16)\\
  &\to\mathrm{Conv}_{64}^{(4\times4)}&\text{size of }(64,8,8)\\
  &\to[\mathrm{ResBlock}_{64}]_{\times{}6}&\text{size of }(64,8,8)\\
  \bZq\in\mathbf{B}^{8\times8}\subset\mathbb{R}^{64\times8\times8}
  &\to[\mathrm{ResBlock}_{64}]_{\times{}6}&\text{size of }(64,8,8)\\
  &\to\mathrm{ConvT}_{64}^{(4\times4)}\to\mathrm{BatchNorm}\to\mathrm{ReLU}&\text{size: }(64,16,16)\\
  &\to\mathrm{ConvT}_{32}^{(4\times4)}\to\mathrm{BatchNorm}\to\mathrm{ReLU}&\text{size: }(32,32,32)\\
  &\to\mathrm{ConvT}_{c}^{(4\times4)}&\text{size: }(c,64,64),
\end{flalign*}
where $c$ is the the number of output channel, which is 2 for vMF SQ-VAE and 256 for other models. In vMF SQ-VAE, the outputs of the encoder and decoder are normalized along the channel axis so that the normalized vectors are on hyperspheres $\mathcal{S}^{64-1}$ and $\mathcal{S}^{2-1}$, respectively.

In this experiment, the optimizer, the initial learning rate and the learning rate scheduling are the same as those in Appendix~\ref{sec:app_sub_details_gaussian_vision}.
%In this experiment, we use the same settings for initial learning rate and learning rate scheduler as those in Appendix~\ref{sec:app_sub_details_gaussian_vision}. 
We run 100 and 70 epochs on MNIST and Gray-CelebA, respectively, with a minibatch size of 32.

We set the latent space dimension of VAE to 11 on MNIST and 18 on Gray-CelebA. In this way, the number of bits representing the latent space becomes almost the same as that of the other models, i.e., $32~\text{bit}\times11\approx7~\text{bit}\times7\times7$ and $32~\text{bit}\times18=9~\text{bit}\times8\times8$.
% In VAE, we use the similar architectures as those of the other models but add a linear layer to the end of the encoder.

\begin{figure*}[t]
\vskip 0.1in
  \centering
  \subfigure[Source]{\includegraphics[width=0.3\textwidth]{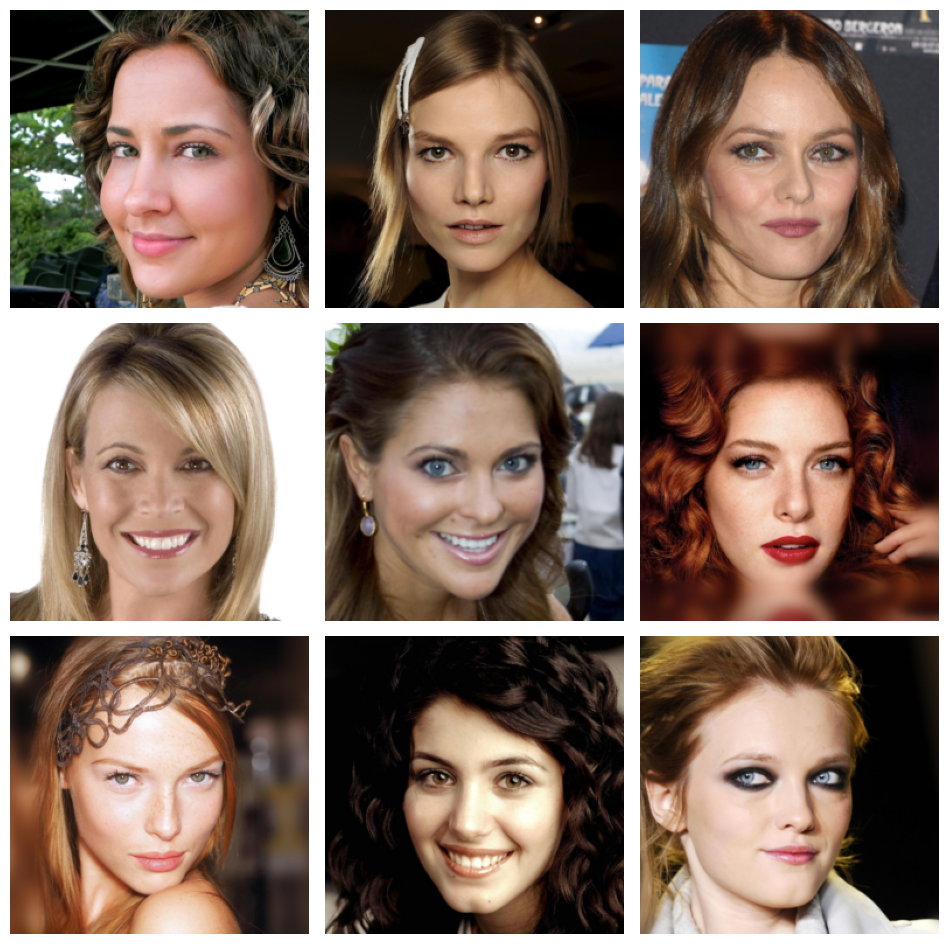}}
%   \subfigure[VQ-VAE w/ EMA]{\includegraphics[width=0.3\textwidth]{figures/vision/vqvae_celebahq_reconst.png}}
  \hspace{20pt}
  \subfigure[Reconstructed samples ]{\includegraphics[width=0.3\textwidth]{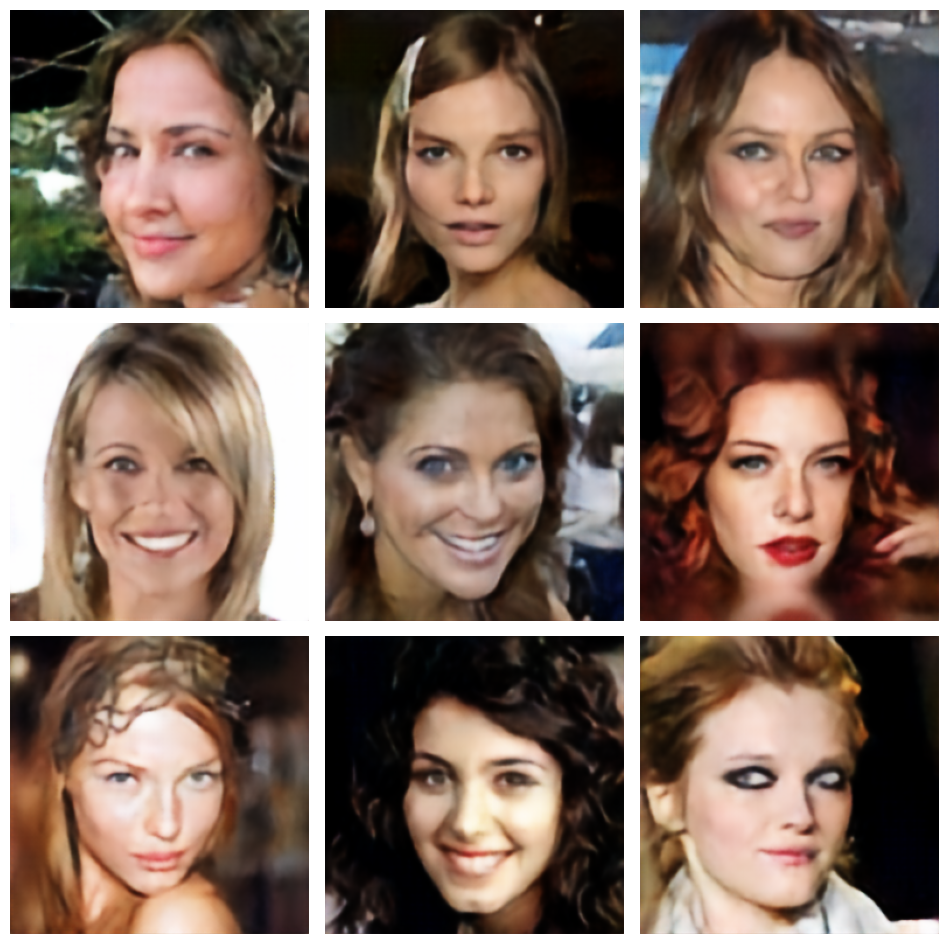}}\\
%   \caption{Reconstructed samples of CelebAHQ 256$\times$256.}
%   \label{fig:celeba_hq_reconst}
% \vskip -0.1in
% \end{figure*}
% \begin{figure}[t]
% \vskip 0.1in
   \centering
   \subfigure[Generated samples]{\includegraphics[width=.3\textwidth]{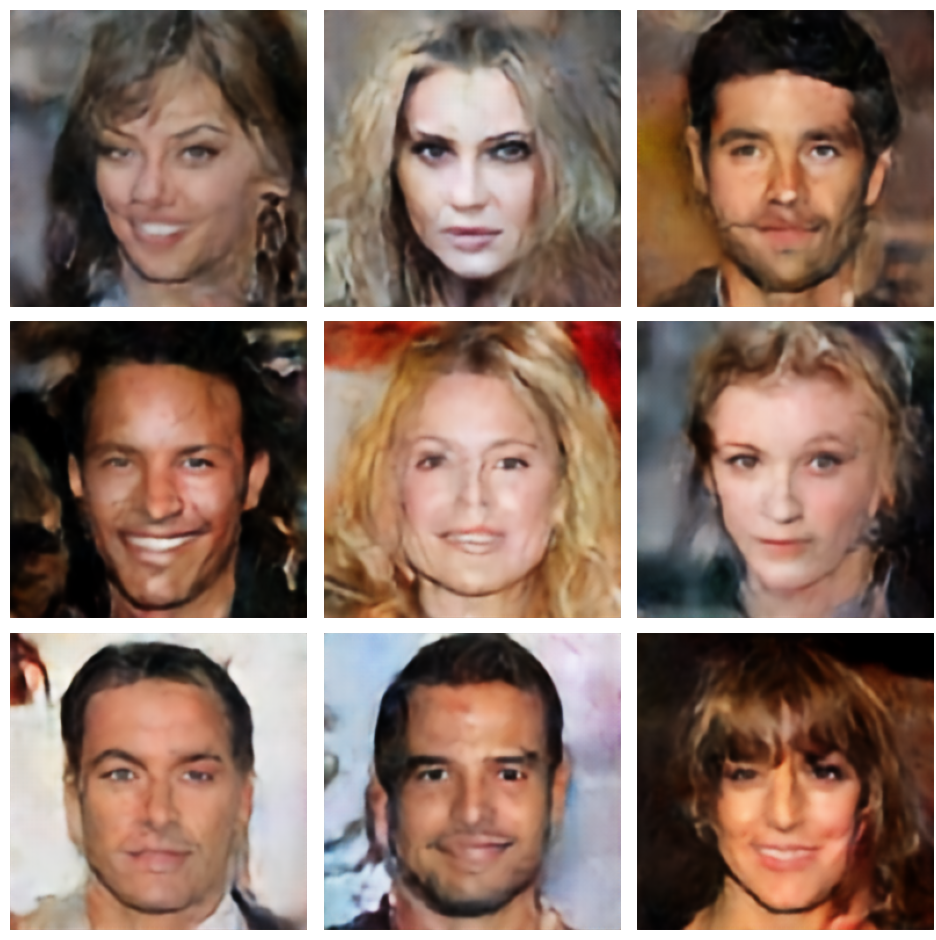}}
   \caption{
   Samples of CelebAHQ 256$\times$256 from SQ-VAE (\uppercase\expandafter{\romannumeral 1}).
   }
   \label{fig:celeba_hq_images}
\vskip -0.1in
\end{figure*}
%%----- Section
\section{Experiments on CelebA HQ 256$\times$256}
\label{sec:app_experiment_celebahq}

To demonstrate that our SQ-VAE is applicable to larger scale datasets, we apply Gaussian SQ-VAE (\Rnum{1}) to CelebA HQ 256$\times$256.
We train Gaussian SQ-VAE (\Rnum{1}) using
the following network architecture:
\begin{flalign*}
  \bx\in\mathbb{R}^{256\times256\times3}
  &\to\mathrm{Conv}_{16}^{(4\times4)}\to\mathrm{BatchNorm}\to\mathrm{ReLU}&\text{size: }(16,128,128)\\
  &\to\mathrm{Conv}_{32}^{(4\times4)}\to\mathrm{BatchNorm}\to\mathrm{ReLU}&\text{size: }(32,64,64)\\
  &\to\mathrm{Conv}_{64}^{(4\times4)}\to\mathrm{BatchNorm}\to\mathrm{ReLU}&\text{size: }(64,32,32)\\
  &\to\mathrm{Conv}_{64}^{(3\times3)}&\text{size of }(64,32,32)\\
  &\to[\mathrm{ResBlock}_{64}]_{\times{}6}&\text{size of }(64,32,32)\\
  \bZq\in\mathbf{B}^{32\times{}32}\subset\mathbb{R}^{64\times{}32\times{}32}
  &\to[\mathrm{ResBlock}_{64}]_{\times{}6}&\text{size of }(64,32,32)\\
  &\to\mathrm{ConvT}_{64}^{(3\times3)}\to\mathrm{BatchNorm}\to\mathrm{ReLU}&\text{size: }(64,32,32)\\
  &\to\mathrm{ConvT}_{32}^{(4\times4)}\to\mathrm{BatchNorm}\to\mathrm{ReLU}&\text{size: }(32,64,64)\\
  &\to\mathrm{ConvT}_{16}^{(4\times4)}\to\mathrm{BatchNorm}\to\mathrm{ReLU}&\text{size: }(16,128,128)\\
  &\to\mathrm{ConvT}_{3}^{(4\times4)}\to\mathrm{Sigmoid}&\text{size: }(3,256,256),
\end{flalign*}

We follow the same experimental setup as that on CelebA 64$\times$64 except for the network architecture.
We show examples of reconstructed images of the models in Figure~\ref{fig:celeba_hq_images}(b).

To ease the difficulty of training a feasible prior, we adapt the hierarchical structure which is popular in the recent autoencoder-based works~\citep{razavi2019generating,dhariwal2020jukebox,william2020hierarchical}. We follow a similar procedure to that of~\citet{dhariwal2020jukebox} by stacking two SQ-VAE models. The architecture of the first SQ-VAE is described as above. The second SQ-VAE has two extra $\mathrm{Conv}^{(4\times4)}$ layers at the end of its encoder and the beginning of its decoder. Also, its latent space is $\bB^{8\times8}$. We use two PixelSNAIL~\citep{chen2018pixel} models to act as the prior and the upsampler between the latent codes of the two SQ-VAEs. As a result, the second PixelSNAIL is conditioned by the latent code of the first SQ-VAE. 
%To randomly generate samples from the SQ-VAE models, we train the prior with conditional PixelSNAIL~\citep{chen2018pixel}. For the conditioning, we train an additional SQ-VAE model, separately, by following the similar procedure in~\citet{dhariwal2020jukebox}.
% The network architecture of the additional model has two extra $\mathrm{Conv}^{(4\times4)}$ layers in both the encoder and decoder and the latent space is $\bB^{8\times8}$.
%The additional model has two extra $\mathrm{Conv}^{(4\times4)}$ layers in both the encoder and decoder, and its latent space is $\bB^{8\times8}$.
We show images generated with the learned approximated prior in Figure~\ref{fig:celeba_hq_images}(c).
% Again, we bilieve the quality of synthetic samples of SQ-VAE can be improved by using other stronger autoregressive estimators. 
As we have demonstrated in Section~\ref{sec:experiments} that SQ-VAE enables to learn good discrete latent features and produce superior reconstructed samples, we believe the quality of synthetic samples of SQ-VAE can be improved via other stronger autoregressive estimators.

\end{document}